\documentclass{article} 
\usepackage{iclr2024_conference,times}

\usepackage{amsmath,amsfonts,bm}

\def\eqref#1{equation~\ref{#1}}

\def\1{\bm{1}}

\DeclareMathAlphabet{\mathsfit}{\encodingdefault}{\sfdefault}{m}{sl}
\SetMathAlphabet{\mathsfit}{bold}{\encodingdefault}{\sfdefault}{bx}{n}

\usepackage{hyperref}
\usepackage{url}
\usepackage{wrapfig}
\usepackage{graphicx}
\usepackage{amsmath}
\usepackage{subcaption}
\usepackage{algorithm}
\usepackage{algorithmicx}
\usepackage{algpseudocode}
\usepackage[utf8]{inputenc} 
\usepackage[T1]{fontenc}    
\usepackage{hyperref}       
\usepackage{url}            
\usepackage{booktabs}       
\usepackage{amsfonts}       
\usepackage{nicefrac}       
\usepackage{microtype}      
\usepackage{xcolor}         
\usepackage{lipsum}
\usepackage{booktabs}
\usepackage{multirow}
\usepackage{natbib}
\iclrfinalcopy

\title{A Benchmark Study on Calibration}

\author{%
  Linwei Tao\\
  University of Sydney\\
  \texttt{linwei.tao@sydney.edu.au} \\
  \And
  Younan Zhu, Haolan Guo \\
  University of Sydney\\
  \texttt{\{yzhu0986, hguo4658\}@uni.sydney.edu.au} \\
  \And
  Minjing Dong \\
  City University of Hong Kong\\
  \texttt{minjdong@cityu.edu.hk} \\
  \And
  Chang Xu \\
  University of Sydney\\
  \texttt{c.xu@sydney.edu.au} \\
}

\begin{document}

\maketitle

\begin{abstract}
Deep neural networks are increasingly utilized in various machine learning tasks. However, as these models grow in complexity, they often face calibration issues, despite enhanced prediction accuracy. Many studies have endeavored to improve calibration performance through the use of specific loss functions, data preprocessing and training frameworks. Yet, investigations into calibration properties have been somewhat overlooked.
Our study leverages the Neural Architecture Search (NAS) search space, offering an exhaustive model architecture space for thorough calibration properties exploration. We specifically create a model calibration dataset. This dataset evaluates 90 bin-based and 12 additional calibration measurements across 117,702 unique neural networks within the widely employed NATS-Bench search space.
Our analysis aims to answer several longstanding questions in the field, using our proposed dataset:
(i) Can model calibration be generalized across different datasets?
(ii) Can robustness be used as a calibration measurement?
(iii) How reliable are calibration metrics?
(iv) Does a post-hoc calibration method affect all models uniformly?
(v) How does calibration interact with accuracy?
(vi) What is the impact of bin size on calibration measurement?
(vii) Which architectural designs are beneficial for calibration?
Additionally, our study bridges an existing gap by exploring calibration within NAS. By providing this dataset, we enable further research into NAS calibration. As far as we are aware, our research represents the first large-scale investigation into calibration properties and the premier study of calibration issues within NAS. The project page can be found at \url{https://www.taolinwei.com/calibration-study}.
\end{abstract}
\section{Introduction}
        \vspace{-0.1in}

Despite their widespread success across various domains, deep neural networks (DNNs) are not immune to producing miscalibrated predictions, leading to either overconfidence or underconfidence. This concern becomes particularly salient for safety-critical applications such as autonomous driving~\citep{feng2019can} and medical diagnosis~\citep{thiagarajan2022training}, where reliance on accurate prediction probabilities is paramount. In these contexts, miscalibrated predictions may give rise to potentially catastrophic consequences.

A myriad of attempts~\citep{mukhoti2020calibrating, kumar2018trainable, tao2023calibrating, karandikar2021soft, krishnan2020improving, zhang2022and, hendrycks2019augmix, muller2019does, deng2021learning, kim2021self, tao2023dual} has been made to mitigate the issue of miscalibration, primarily focusing on loss functions or training frameworks. However, the calibration properties of the neural network architectures themselves have received comparatively less attention. \citet{guo2017calibration} were among the first to investigate the relationship between neural architectures and calibration performance, but their work was restricted to the effects of varying depth and width in a ResNet\citep{he2016deep}. More recent studies~\citep{minderer2021revisiting} have extended this exploration to modern neural networks, such as the non-convolutional MLP-Mixer~\citep{tolstikhin2021mlp} and Vision Transformers~\citep{ranftl2021vision}. However, the limited diversity and quantity of model architectures have constrained the depth of calibration property studies.

NAS~\citep{liu2018darts, dong2019searching, dong2019one, xu2019automatic, su2022vitas} has brought about a revolution in the field of deep learning by automating the discovery of neural architectures that outperform traditionally hand-designed architectures like AlexNet\citep{krizhevsky2009learning} and ResNet~\citep{he2016deep}. NAS has achieved advancements in part due to its predefined comprehensive model architecture space. NAS benchmark search spaces~\citep{ying2019nasbench101, dong2020bench, DBLP:journals/corr/abs-2008-09777, dong2021nats} have provided an extensive set of unique convolutional architectures for NAS research. While most prior works on search space focus on the topological architecture of models, the NATS-Bench~\citep{dong2021nats} provides a more expansive search space, taking into account models of different sizes. This comprehensive search space can serve as a potent tool to bridge the gap in previous calibration studies.

In our work, to exam the calibration property and address calibration-related research questions, such as the reliability of calibration metrics, one approach is to assess the consistency of different metrics based on a substantial number of well-trained models. However, collecting such a substantial dataset is often challenging due to the associated training costs. Fortunately, NATS-Bench~\citep{dong2021nats} provides access to 117.9K well-trained models with various architectural designs, enabling us to conduct a comprehensive and generalisable study. Specifically, we evaluate all 117,702 unique CNN architectures concerning topology and model size, and benchmark them on multiple calibration metrics of different types. We also include the results on 11 different Vision Transformers to generalize our findings. The specifics about architectures and metrics are discussed in section~\ref{sec:dataset_generation}. This comprehensive dataset has served as a benchmark for our subsequent studies and in-depth analysis on the calibration properties. In this study, we try to answer the following questions that are longstanding in this field:

\begin{enumerate}
\item Can model calibration be generalized across different datasets?~\ref{sec:Can model calibration be generalized across different datasets?}
\item Can robustness be used as a calibration measurement?~\ref{Can robustness be used as a calibration measurement?}
\item How reliable are calibration metrics?~\ref{How reliable are calibration metrics?}
\item Does a post-hoc calibration method affect all models uniformly?~\ref{Does a post-hoc calibration method affect all models uniformly?}
\item How does calibration interact with accuracy?~\ref{How does calibration interact with accuracy?}
\item What is the impact of bin size on calibration measurement?~\ref{What is the impact of bin size on calibration measurement?}
\item Which architectural designs are beneficial for calibration?~\ref{Which architectural designs are beneficial for calibration?}

\end{enumerate}

This exploration aims to shed light on the often-overlooked aspect of calibration, thereby contributing to a more holistic understanding of deep learning model performance. 
\vspace{-0.1in}

\section{Related Works}
\label{sec: related works}
        \vspace{-0.1in}

\noindent\textbf{Calibration Metrics}
Extensive research has been done on calibration metrics, which is crucial for measure the reliability of predictions. The Expected Calibration Error (ECE) is a widely used metric introduced by \citet{naeini2015obtaining}, which quantifies the absolute difference between predicted confidence and empirical accuracy. However, ECE is susceptible to estimator bias, complicating its estimation \citep{nixon2019measuring, vaicenavicius2019evaluating, gupta2021distribution}. To address the bias-variance trade-off, adaptive binning techniques such as Equal Mass Binning calibration error ($\text{ECE}{em}$) have been proposed by \citet{kumar2019verified} and \citet{nixon2019measuring}. In multi-class settings, class-wise calibration errors, including Classwise Equal-Width Binning calibration error (cwCE) and Classwise Equal-Mass Binning calibration error ($\text{cwCE}{em}$), offer potential solutions \citep{kumar2019verified}. Alternatives to ECE encompass likelihood-based measures, the Brier score \citep{brier1950verification}, and Bayesian methods \citep{gelman2013philosophy}. \citet{zhang2020mixnmatch} proposed the Top-Label calibration error using Kernel Density Estimation (KDECE) to avoid binning schemes. \citet{gupta2020calibration} introduced the Kolmogorov-Smirnov calibration error (KSCE), which uses the Kolmogorov-Smirnov test to compare empirical cumulative distribution functions. Additionally, \citet{kumar2018trainable} suggested the Maximum Mean calibration error (MMCE), a differentiable estimator that compares the Top-Label confidence with the conditional accuracy.

\noindent\textbf{Empirical Study on Calibration}~There have been limited studies on evaluating the calibration of uncertainty estimates in deep learning models. \citet{guo2017calibration} discovered that many models with deeper or wider architectures are poorly calibrated and suggested a simple post-processing method called temperature scaling that can substantially improve calibration. In a recent study, \citet{minderer2021revisiting} revisited the calibration of modern neural networks, such as MLP-mixer \citep{tolstikhin2021mlp} and ViT \citep{dosovitskiy2020image}, and found that these new, larger models outperform CNNs. They also examined the impact of training frameworks like SimCLR~\citep{chen2020simple} and CLIP~\citep{radford2021learning}. However, these studies are based on a limited set of model architectures, and more research is needed to generalize their findings.

\noindent\textbf{NAS}~The search space in NAS serves as a critical component for exploring calibration properties.  Various search spaces, such as fully connected, convolutional, recurrent, and attention-based architectures, have been proposed in existing literature.
NAS-Bench-101~\citep{ying2019nasbench101}, the inaugural public architecture dataset for NAS research, comprises 423k unique convolutional architectures. NAS-Bench-201~\citep{dong2020bench} expands on NAS-Bench-101 by offering a different search space. NAS-Bench-301~\citep{DBLP:journals/corr/abs-2008-09777} addresses the issues found in tabular NAS benchmarks. Most prior studies on search spaces concentrate on the topological architecture of models.
Recently, NATS-Bench~\citep{dong2021nats} was introduced as an extension to NAS-Bench-201~\citep{dong2020bench}, providing an expanded search space for models of varying sizes. This expansion marks a significant advancement in the field, creating new possibilities for the exploration and improvement of deep learning models.

        \vspace{-0.1in}

\section{Dataset Generation}
        \vspace{-0.1in}

\label{sec:dataset_generation}
In this section, we detail the metrics and model architectures from NATS-Bench~\citep{dong2021nats} that are involved in our study. To ensure the generality of our findings, we also include 11 vision transformers of different architecture design. Each unique architecture is pretrained for 200 epochs on three benchmark datasets: CIFAR-10~\citep{krizhevsky2009learning}, CIFAR-100~\citep{krizhevsky2009learning}, and ImageNet16-120~\citep{chrabaszcz2017downsampled}. Note that ImageNet16-120 is a down-sampled variant of ImageNet, widely used in NAS literature~\citep{ying2019nasbench101, dong2020bench, dong2021nats, patel2020multi}. We choose this dataset because it can reduce computational costs while maintaining comparable results as ImageNet~\citep{chrabaszcz2017downsampled}. For simplicity, we refer to ImageNet16-120 as ImageNet in the following discussion. To evaluate post temperature scaling, we create a validation set by splitting the original test set into 20\%/80\% for validation/test.

We use extensive bin-based metrics and a range of other calibration measures in this study, which provides a thorough assessment of model calibration performance across different scenarios. We evaluate all calibration metrics discussed in~\ref{sec: related works} and more details about metrics can be found in Appendix~\ref{sec:Calibration Metrics}. Since prediction performance on Out-of-Distribution (OoD) datasets is also a strong indicator of calibration~\citep{mukhoti2020calibrating}, we include the Area Under the Curve (AUC) for the TSS models, evaluating on two OoD datasets, CIFAR-10-C (corrupted with Gaussian noise)~\citep{hendrycks2019benchmarking} and SVHN (Street View House Numbers)~\citep{netzer2011reading}. We evaluate these metrics across a wide range of bin sizes, including 5, 10, 15, 20, 25, 50, 100, 200, and 500 bins. These metrics are assessed both before and after temperature scaling. Overall, this results in the assessment of 102 different measurements, providing a comprehensive evaluation of model calibration.

        \vspace{-0.1in}

\subsection{Architectures Evaluated}
        \vspace{-0.1in}

\begin{wrapfigure}{R}{0.3\textwidth}
    \centering
    \includegraphics[width=0.3\textwidth]{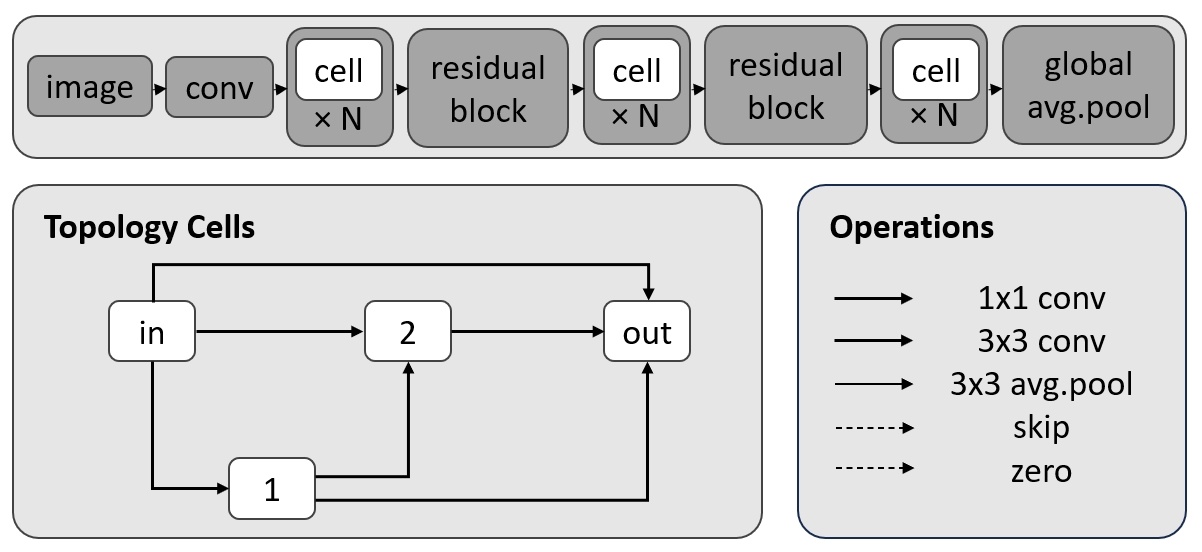}
    \caption{the macro skeleton of each candidate architecture at the top, cell representations at the bottom-left, and operation candidates at the bottom-right. The candidate channels for SSS are 8, 16, 24, 32, 40, 48, 56, and 64.}
    \label{fig:archi}
    \vspace{-0.1in}
\end{wrapfigure}

We evaluate calibration properties on 117,702 unique architectures in NATS-Bench~\citep{dong2021nats}, which is a cell-based neural architecture search space consisting of two search spaces: Topology Search Space (TSS) and Size Search Space (SSS). Referring to Figure~\ref{fig:archi}, each cell is represented by a densely-connected directed acyclic graph with four nodes and six edges. Within this structure, nodes denote feature maps, while edges correspond to selected operations from a predefined operation set \( O \). The operation set \( O \) encompasses the following operations: \(O = \{1 \times 1 \textrm{ convolution}, 3 \times 3 \textrm{ convolution}, 3 \times 3 \textrm{ average pooling}, \textrm{skip}, \textrm{zero}\}\). The search space contains 15,625 architectures, but only 6,466 are unique due to isomorphic cells resulting from the skip and zero operations.

In SSS, each architecture has a different configuration for the number of channels in each layer. In this space, ``stacks” refer to the aggregation of multiple cells. Each stack is constructed using the cell structure that has demonstrated the best performance in the TSS on the CIFAR-100 dataset. A total of 32,768 architecture candidates are evaluated on the same three datasets for 90 epochs. For our evaluations, we assess all architectures within both search spaces and evaluate them on different test splits of their corresponding datasets. In total, we evaluate a total of $3 \times 6466 = 19398$ networks on TSS and $3 \times 32768 = 98304$ on SSS. 

To ensure the generality of our findings, we also include 11 vision transformers of different architecture design including 
T2T-ViT-7, T2T-ViT-10, T2T-ViT-12, T2T-ViT-19, T2T-ViT-24~\citep{yuan2021tokens}, ViT-b-16~\citep{dosovitskiy2020transformers}, Swin-T~\citep{liu2021swin}, Deit-T~\citep{touvron2021training}, Cait-XXS24~\citep{touvron2021going}, PvTv2-T~\citep{wang2022pvt} and PoolFormer-S12~\citep{yu2022metaformer}
. Each transformer is fine tuned 60 epochs on CIFAR-10, CIFAR-100 and ImageNet-16-120 based on the pretrained weights on ImageNet-1k.

        \vspace{-0.1in}

\section{Experiments and Discussion}
        \vspace{-0.1in}

Our research endeavors to investigate the influence of several factors, including the dataset, calibration metrics, post-hoc calibration methods, accuracy and robustness on the calibration performance of neural networks. Additionally, we seek to analyze the impact of the architecture designs on calibration. With the comprehensive evaluation results in hand, we have made several empirical observations and derived insightful conclusions, which could be summarized to answer the following questions.

\begin{wrapfigure}{R}{0.3\textwidth}
    \centering
    \includegraphics[width=0.3\textwidth]{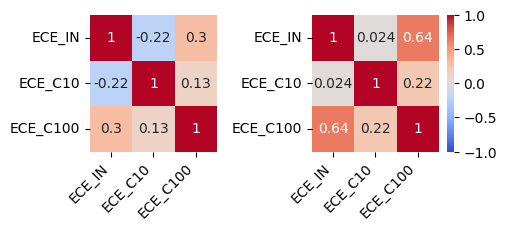}
    \caption{Kendall ranking correlation matrix of ECE for different TSS architecture subsets. The left subplot corresponds to the top 1000 architectures based on accuracy, while the right subplot represents the entire set of models.}
    \label{fig:1000_all_kendall_all_datasets}
    \vspace{-0.1in}
\end{wrapfigure}
        \vspace{-0.05in}

\subsection{Can model calibration be generalized across different datasets?}
\label{sec:Can model calibration be generalized across different datasets?}
\vspace{-0.05in}
We're particularly interested in understanding if there are noticeable variations in model calibration when different datasets are used for training. Additionally, we aim to investigate how the complexity and diversity of the dataset might impact calibration performance.

\noindent\textbf{Experimental setup:}
We examine two subsets of ECE measurement from our TSS dataset: the first subset consists of architectures with good prediction accuracy in each dataset (ranked top 1000 by accuracy), while the second subset includes all architectures available in the dataset. The aim of our investigation is to probe the model calibration performance correlation between datasets. To achieve this, we resort to the Kendall ranking correlation coefficient\citep{kendall1938new}. This non-parametric statistic is a potent tool for measuring the degree of correspondence between two rankings, with its values ranging from -1 (indicating perfect disagreement) to +1 (indicating perfect agreement). A Kendall coefficient of zero would signify a lack of correlation. With the help of Kendall ranking correlation coefficient, we build a correlation matrix for ECE on CIFAR-10 (ECE\_C10), CIFAR-100 (ECE\_C100), ImageNet (ECE\_IN) for both models with high accuracy (left) and all models (right) as displayed in Figure~\ref{fig:1000_all_kendall_all_datasets}. 

\noindent\textbf{Discussion:}~Upon examining the correlation across datasets, we observed a substantial variation in the ranking of calibration metrics. Notably, the calibration performance on CIFAR-10 exhibited little correlation with the calibration performance on CIFAR-100, despite their images being the same. Furthermore, the calibration performance on CIFAR-10 exhibited no correlation with the performance on ImageNet and even negatively correlated for top 1000 models. In contrast, the correlation of calibration performance between the more complex datasets CIFAR-100 and ImageNet was relatively higher. This observation suggests that the relationship between calibration performance of a certain model and the evaluation dataset may be weak or non-existent. Thus, \textbf{the calibration property of a certain architecture can not generalize well to different datasets}, researchers cannot rely on evaluation results on different datasets when selecting a calibrated architecture for downstream datasets. We draw the similar conclusion among Transformer architectures as shown in Appendix~\ref{sec:Experiments on Transformers}. Our observation is also consistent among other calibration metrics, as detailed in the Appendix~\ref{sec:Calibration Kendall Ranking Confusion Matrix for TSS}.
\vspace{-0.05in}
\subsection{Can robustness be used as a calibration measurement?}
\label{Can robustness be used as a calibration measurement?}
\vspace{-0.05in}
Previous research~\citep{thulasidasan2019mixup} suggests that a well-calibrated model should perform well on OoD datasets. \begin{wrapfigure}{R}{0.6\textwidth}
    \centering
    \vspace{-0.1in}
    \includegraphics[width=0.6\textwidth]{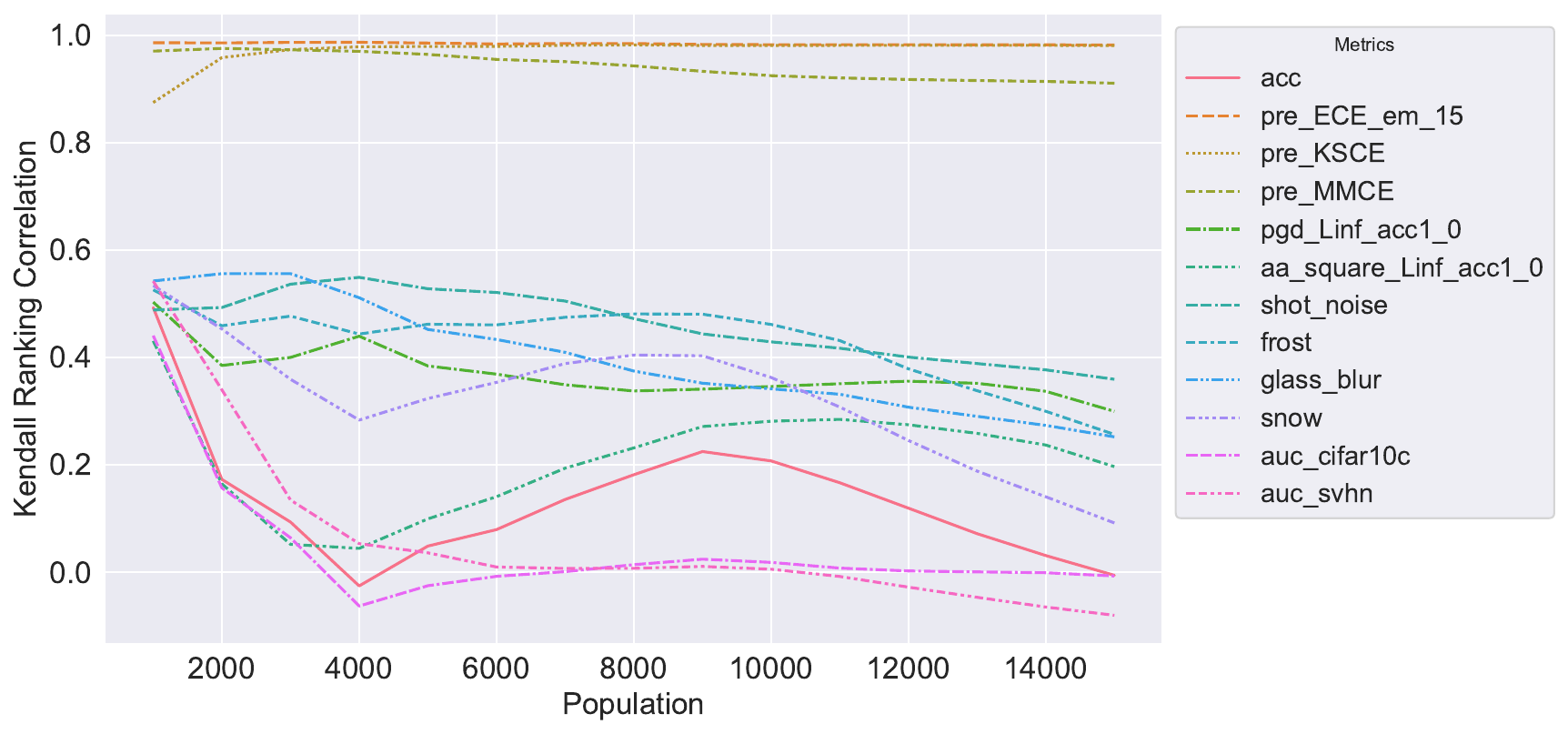}
    \caption{Kendall ranking correlation of various metrics against ECE different top-ranked model population.}
    \label{fig:kendall_ece_top_ranked_models}
    \vspace{-0.1in}
\end{wrapfigure}
Several studies, including~\citep{mukhoti2020calibrating}, have utilized this notion to evaluate their calibration methods. Given that accuracy on corruption datasets is a robustness indicator and have a strong correlation with adversarial robustness, a question arises: can other robustness metrics serve as a measure of calibration?

\noindent \textbf{Experiment Setup:}~To address this question, we focus on the Kendall ranking correlation matrix between ECE and various metrics for models ranked in the top 1000, 2000, and so on, based on their accuracy. Recently, a dataset based on NAS-Bench-201~\citep{jung2023neural} was proposed, which evaluates the robustness performance on all TSS models available in NATS-Bench and enables us to evaluate the relationship between robustness and calibration. Thus, the considered metrics include accuracy, calibration metrics such as bin-based $\text{ECE}_{em}$, statistics-based KSCE, and kernel-based MMCE, as well as robustness metrics like adversarial robustness (PGD attack and square attack) and accuracy on corruption datasets (forest, shot noise, snow, and glass blur), and Area under the Curve (AuC) on CIFAR-10-C and SVHN. We depicted the correlation coefficients for different populations using a line plot, as shown in Figure~\ref{fig:kendall_ece_top_ranked_models}. The experiment was conducted on CIFAR-10, and experiments for other datasets are provided in Appendix~\ref{sec:Relationship Accuracy, Robustness and Calibration Measurements}.

\noindent \textbf{Discussion:}~Upon analyzing the top 1000 ranked models, we found that all robustness metrics, including adversarial robustness and accuracy on corruption datasets, exhibited a strong correlation with ECE~\citep{jung2023neural}. Additionally, even prediction accuracy had a high correlation with ECE. However, as we included worse-performing models, such as top 4000 accuracy models, the correlation was not observed for most metrics such as accuracy and snow corruption accuracy. Nonetheless, some robustness metrics, such as PGD attack robustness accuracy, still displayed a high correlation with ECE. On the other hand, bin-based $\text{ECE}_{em}$, statistics-based KSCE, and kernel-based MMCE had a high correlation with ECE regardless of the model's performance on prediction accuracy.

The study reveals that \textbf{calibration performance can be measured not only by the robustness accuracy on the corruption dataset, but also by other robustness metrics only among models with high prediction accuracy}. However, when considering models with varying prediction performance, there seems to be no correlation between the AuC on OoD datasets and ECE, as with most other robustness metrics. This suggests that \textbf{including AuC on OoD datasets in robustness metrics may not reliably measure calibration performance for models of varying prediction performance}. It is worth more caution when using AuC as a metric for evaluating the calibration of neural networks.

\vspace{-0.05in}
\subsection{How reliable are calibration metrics?}
\label{How reliable are calibration metrics?}
\vspace{-0.05in}
\label{metric_reliability}
We aim to explore whether certain metrics are particularly effective in specific scenarios, or if their assessments generally align. A critical aspect of this investigation involves identifying if there is a substantial disparity in the results when employing different calibration metrics.

\noindent\textbf{Experimental Setup.}~
We provide a Kendall ranking correlation matrix between 7 calibration metrics of different types within the CIFAR-10 and ImageNet datasets on TSS to demonstrate ranking correlations between these metrics.

\begin{figure}[!ht]
    \centering
    \begin{subfigure}{0.64\textwidth}
        \centering
        \vspace{-0.1in}
        \includegraphics[width=\textwidth]{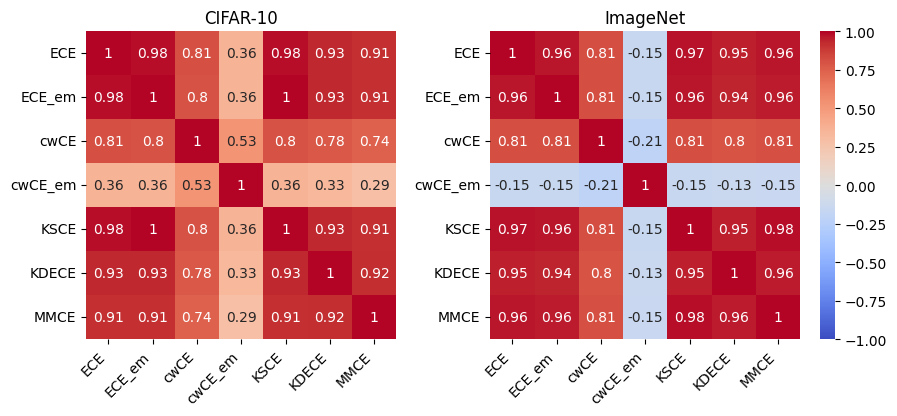}
        \caption{\label{fig:metric_correlation}Kendall ranking correlation between diverse calibration metrics. The metrics are evaluated across the entire set of TSS models. The analysis spans each of the CIFAR-10(left) and ImageNet(right).}
    \end{subfigure}
        \hfill
    \begin{subfigure}{0.33\textwidth}
        \centering
        \includegraphics[width=\textwidth]{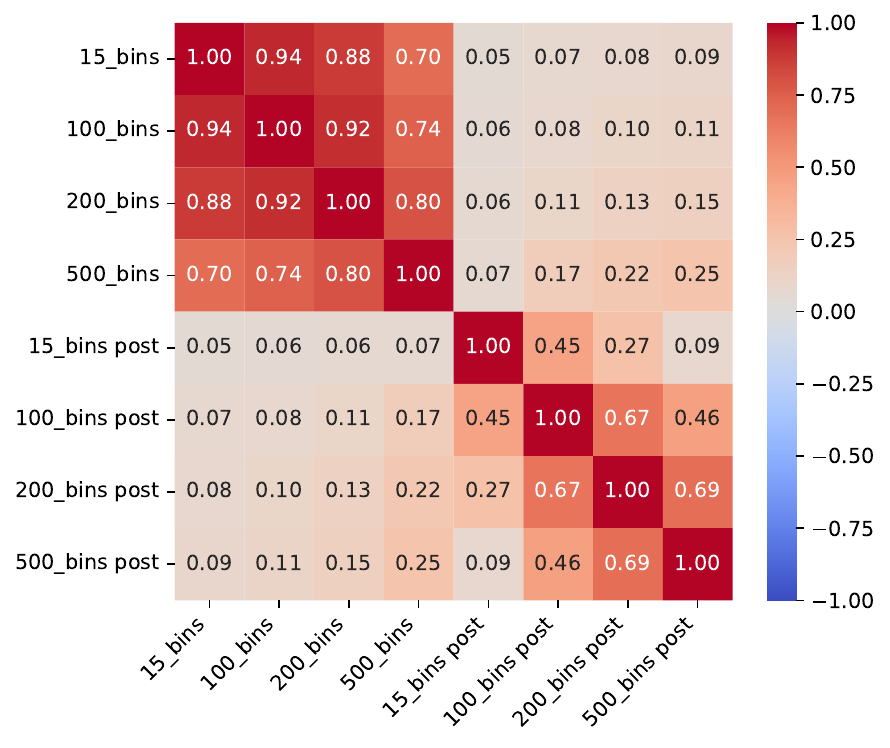}
        \caption{\label{fig:Kendall_cifar10_temp_8x8}Kendall Ranking Correlation Matrix of ECE before and after temperature scaling on CIFAR-10.}
        \label{fig:imagenet_modelsize_sss}
    \end{subfigure}

    \caption{Explore the properties of calibration metrics.}
    \vspace{-0.2in}
\end{figure}

\noindent\textbf{Discussion:}~Despite the availability of numerous calibration metrics, a theoretical gap exists between bin-based, kernel-based, and statistics-based metrics. Our evaluation reveals the correlation between metrics, as illustrated in Figure~\ref{fig:metric_correlation}, \textbf{a consistent trend in the ranking of most calibration performance regardless of metric type}. Although multiple works~\citep{nixon2019measuring, kumar2019verified, roelofs2022mitigating, gupta2020calibration} point out the demerits of ECE, the ECE shows consistent results with most other metrics. Our analysis of the correlation coefficients of NLL and Brier score showed consistency across various metrics. We extended our correlation analysis to all models in SSS and observed similar trends. It is worth noting that the classwise-based calibration error metrics, such as cwCE and $\text{cwCE}_{em}$, exhibited a lower correlation with other metrics, where $\text{cwCE}_{em}$ is even negatively correlated with other widely accepted metrics on ImageNet. This indicates that \textbf{$\text{cwCE}_{em}$ may not be a reliable metric for calibration measurement.} We postulate that this discrepancy could stem from inherent limitations associated with equal-mass binning. Specifically, equal-mass binning might yield a bin that covers a broad range of high uncertainty. This could prioritize smaller quantization errors for low-ranked logits over focusing on high-ranked logits, which could subsequently undermine calibration performance. We observed that this degradation caused by equal-mass binning tends to diminish when datasets have fewer classes. For instance, in the case of CIFAR-10, the higher class prior appears to mitigate the negative effects of equal-mass binning. We draw the similar conclusion among Transformer architectures as shown in Appendix~\ref{sec:Experiments on Transformers}.

\vspace{-0.05in}
\subsection{Does a post-hoc calibration method affect all models uniformly?}
\label{Does a post-hoc calibration method affect all models uniformly?}
\vspace{-0.05in}
The research conducted by~\citet{wang2021rethinking} 
suggests that regularized models typically generate more accurate and well-calibrated predictions, but may have lower calibratability. 
In other words, regularized models have lower calibration space for post-hoc calibration techniques, such as temperature scaling and histogram binning. However, this observation is based on a limited number of cases. Our dataset allows for a more comprehensive exploration of this issue.

\noindent \textbf{Experiment Setup:}~To investigate this issue thoroughly, we computed the Kendall ranking correlation matrix between pre-temperature-scaling ECE (pre-ECE) and post-temperature-scaling ECE (post-ECE) on CIFAR-10 with different bin sizes for all models in TSS dataset.

\noindent \textbf{Discussion:}~As shown in Figure~\ref{fig:Kendall_cifar10_temp_8x8}, the correlation coefficient between pre-ECE and post-ECE is nearly zero, suggesting that a well-calibrated model may not maintain its ranking of calibration performance after undergoing post-hoc calibration methods. On the other hand, a less well-calibrated model may improve its calibration performance after such methods. Our findings expand on the conclusion for regularized models and indicate that \textbf{well-calibrated models do not necessarily exhibit better calibration performance after post-hoc calibration techniques}. This observation is align with ~\citep{ashukha2020pitfalls}, which indicates the comparison of calibration performance between different methods without post-calibration might not provide a fair ranking. Furthermore, we note that the selection of bin size has a greater influence on post-hoc calibration measurements. Additionally, we observe similar results present on ImageNet and CIFAR-100 for both top 1000 models and all models. We draw the similar conclusion among Transformer architectures as shown in Appendix~\ref{sec:Experiments on Transformers}. Additional results are included in the Appendix~\ref{sec:Kendall Ranking Correlation Matrix before and after Temperature Scaling}.
\vspace{-0.05in}
\subsection{How does calibration interact with accuracy?}
\label{How does calibration interact with accuracy?}
\vspace{-0.05in}
\begin{wrapfigure}{R}{0.6\textwidth}
    \centering
    \vspace{-0.1in}
    \includegraphics[width=\linewidth]{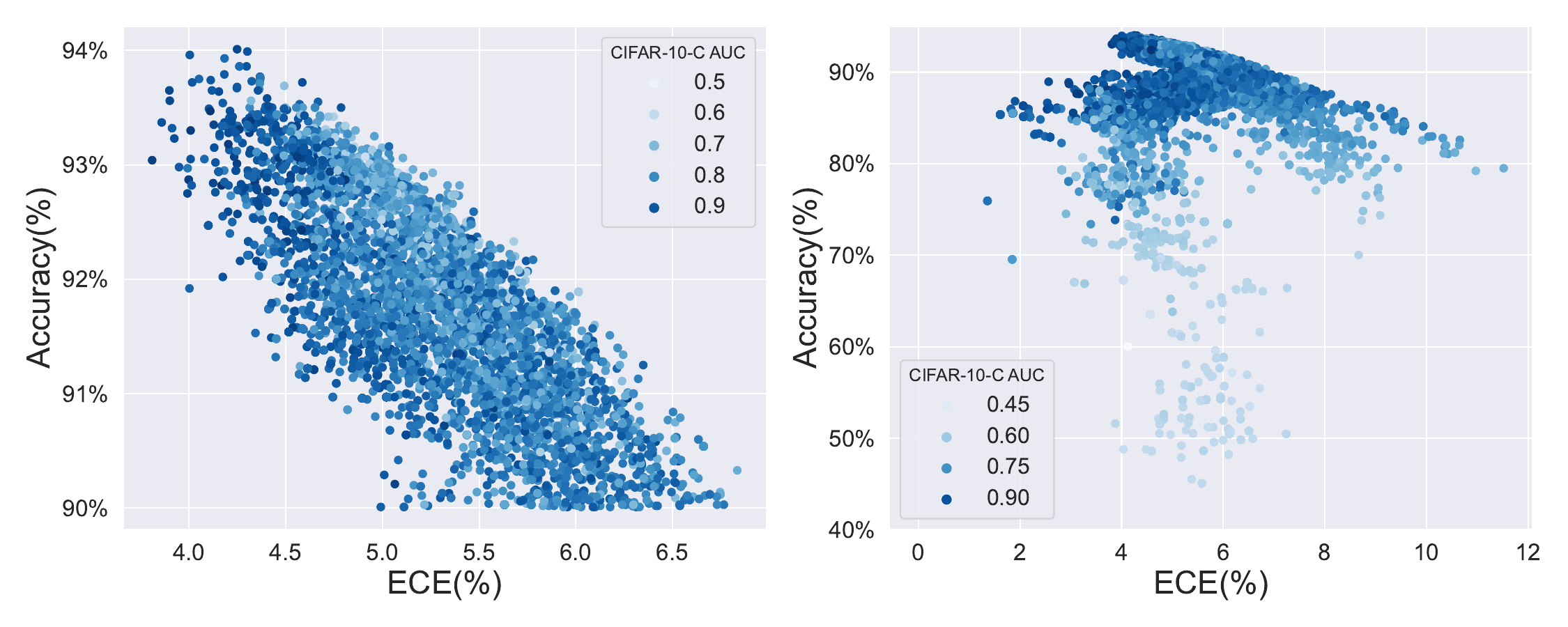}
   \caption{Scatter plots depict the ECE versus Accuracy of model with accuracy larger than 90\% (left) and all TSS models (right) on CIFAR-10. The color-coded markers represent CIFAR-10-C AUC scores.}
\label{fig:auc_ece_acccuracy_cifar10c}
\vspace{-0.1in}
\end{wrapfigure}
Extensive research in the robustness literature has investigated the trade-off between accuracy and robustness such as \citep{zhang2019theoretically}. However, the interplay between accuracy and calibration has not been as thoroughly explored. While some prior studies hinted at a potential trade-off between accuracy and calibration \citep{mukhoti2020calibrating, karandikar2021soft}, these findings seem limited or not universally consistent. Therefore, further investigation into this topic using our dataset could yield valuable insights.

\noindent \textbf{Experiment Setup:}~We created a scatter plot for all models and models with high accuracy on ECE versus Accuracy. The experiment was conducted on CIFAR-10. Experiments for other datasets are included in the Appendix~\ref{sec:Datasets}.

\noindent \textbf{Discussion:}~Figure~\ref{fig:auc_ece_acccuracy_cifar10c} illustrates that when considering only models with accuracy over 90\% (as depicted in the left plot), a distinct pattern emerges: higher accuracy is associated with better calibration performance, whereas lower accuracy results in poorer calibration performance. However, when analyzing all models in TSS, this pattern is not evident. Therefore, it can be concluded that \textbf{the positive correlation between accuracy and calibration exists only among architectures with good prediction performance, challenging the previously hinted trade-off.}
\vspace{-0.05in}
\subsection{What is the impact of bin size on calibration measurement?}
\label{What is the impact of bin size on calibration measurement?}
\vspace{-0.05in}
A pivotal factor in bin-based calibration measurement is the choice of bin size (number of bins). Larger bin sizes encapsulate a wider range of confidence levels, which can potentially mask variations and result in an underestimation of miscalibration. Conversely, very small bin sizes may lead to overfitting or unstable estimates due to insufficient data in each bin. Therefore, the study on bin size choice is essential for accurate calibration measurements.

\noindent \textbf{Experimental Setup:}~To address this issue, we focus on the evaluation of the bin-based calibration metrics with different bin sizes on before and after temperature scaling. We also explore the calibration performance on both 12 and 200 epochs to examine the effect of bin size on under-fitting models.
\begin{figure}[!ht]
\vspace{-0.1in}
    \centering
    \begin{subfigure}{0.32\textwidth}
        \centering
        \vspace{-0.1in}
        \includegraphics[width=\textwidth]{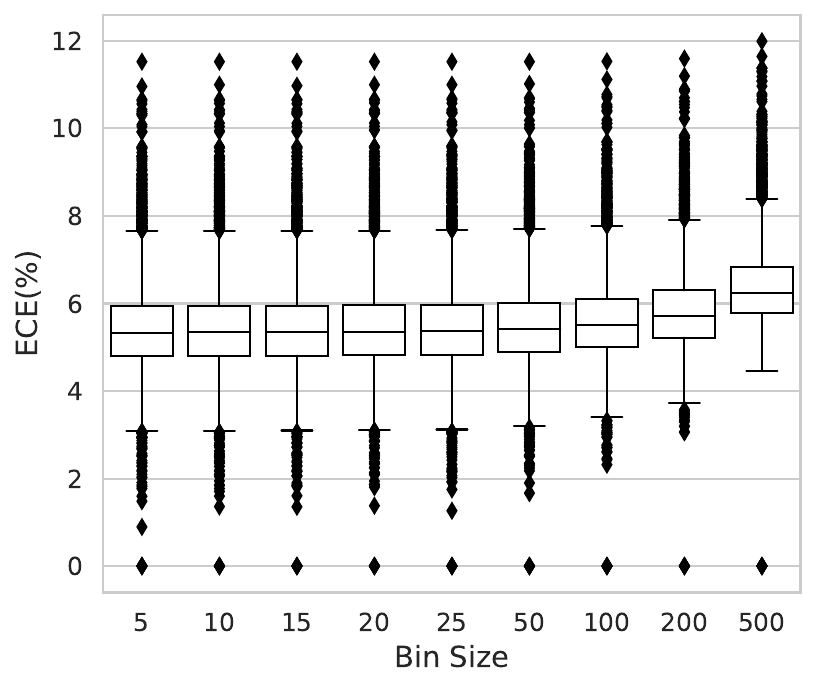}
        \caption{Boxplot of ECE of different bin size before temperature scaling on 200 epochs.}
        \label{fig:boxplot_cifar10_preece}
    \end{subfigure}
    \hfill
    \begin{subfigure}{0.32\textwidth}
        \centering
        \includegraphics[width=\textwidth]{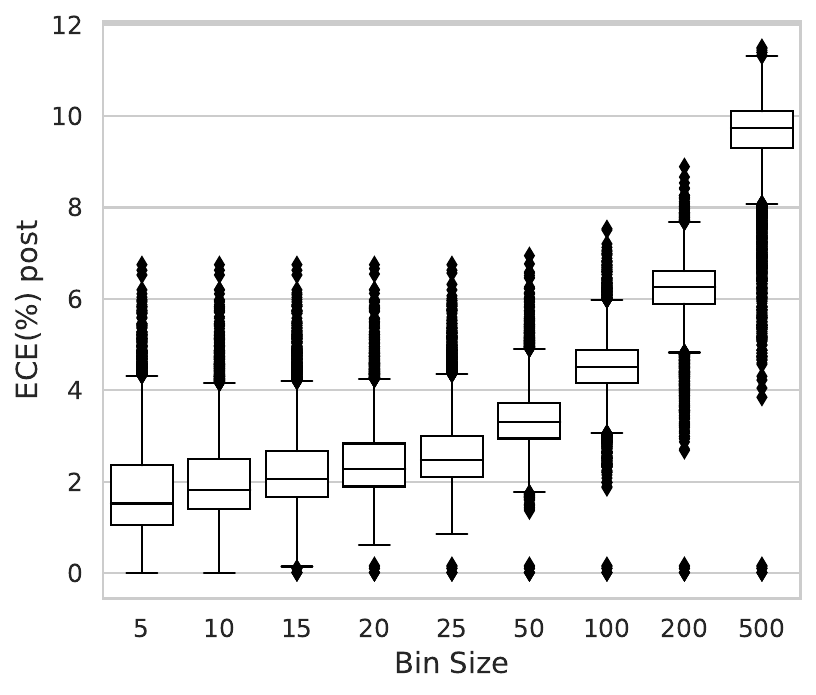}
        \caption{Boxplot of ECE of different bin size after temperature scaling on 200 epochs.}
        \label{fig:boxplot_cifar10_postece}
    \end{subfigure}
    \hfill
    \begin{subfigure}{0.32\textwidth}
        \centering
        \includegraphics[width=\textwidth]{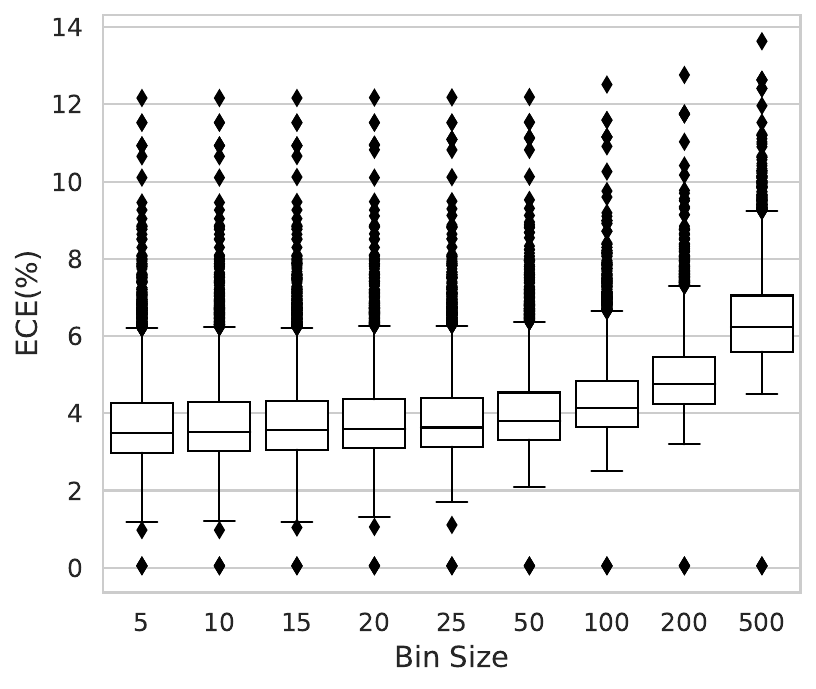}
        \caption{Boxplot of ECE of different bin size before temperature scaling on 12 epochs.}
        \label{fig:boxplot_cifar10_preece_12}
    \end{subfigure}
    \caption{Explore the impact of bin size on calibration before and after temperature scaling.}
\vspace{-0.1in}
\end{figure}

\noindent \textbf{Discussion:}~The results from Figures \ref{fig:boxplot_cifar10_preece} and \ref{fig:boxplot_cifar10_postece} demonstrate that both pre- and post-ECE increase in correlation with bin size. However, the post-ECE showcases a more pronounced rise—doubling at a bin size of 100 and quintupling at a bin size of 500. This observation is further substantiated by Figure \ref{fig:Kendall_cifar10_temp_8x8}, where post-ECE measurements with a bin size of 500 reveal barely any correlation when compared to those at a bin size of 100. Figures \ref{fig:boxplot_cifar10_preece_12} and \ref{fig:boxplot_cifar10_preece} display the variation of ECE across diverse bin sizes at different stages of model training. Remarkably, for models trained for only 12 epochs (shown in Figure \ref{fig:boxplot_cifar10_preece_12}), the choice of bin size significantly influences calibration performance more than fully trained models that have trained 200 epochs.

These findings suggest that \textbf{bin size has a more substantial impact on post-ECE}, with a negligible correlation between post-ECE measurements at varying bin sizes. Conversely, pre-ECE seems more resistant to changes in bin sizes. Therefore, for a holistic comparison, \textbf{it is recommended to assess post-hoc calibration performance across a range of bin sizes.} Moreover, the influence of bin size is more conspicuous for underfitting models, which implies that choosing the correct bin size is particularly crucial for underfitting models. These trends were also observed when evaluating other bin-based calibration metrics and in the SSS, details of which are provided in the Appendix~\ref{sec:Calibration Kendall Ranking Confusion Matrix for SSS}.


        

\vspace{-0.05in}
\subsection{Which architectural designs are beneficial for calibration?}
\label{Which architectural designs are beneficial for calibration?}
\vspace{-0.05in}
The focus now shifts towards comprehending the influence of architecture design on calibration performance. Specifically, the number of kernels and the topological architecture design are investigated. Previous research~\citep{guo2017calibration} has suggested that complex neural networks tend to be poorly calibrated. However, our findings only consistent with this observation in limited cases. 

\begin{figure}[!ht]
    \centering
    \begin{subfigure}{0.32\textwidth}
        \centering
        \includegraphics[width=\textwidth]{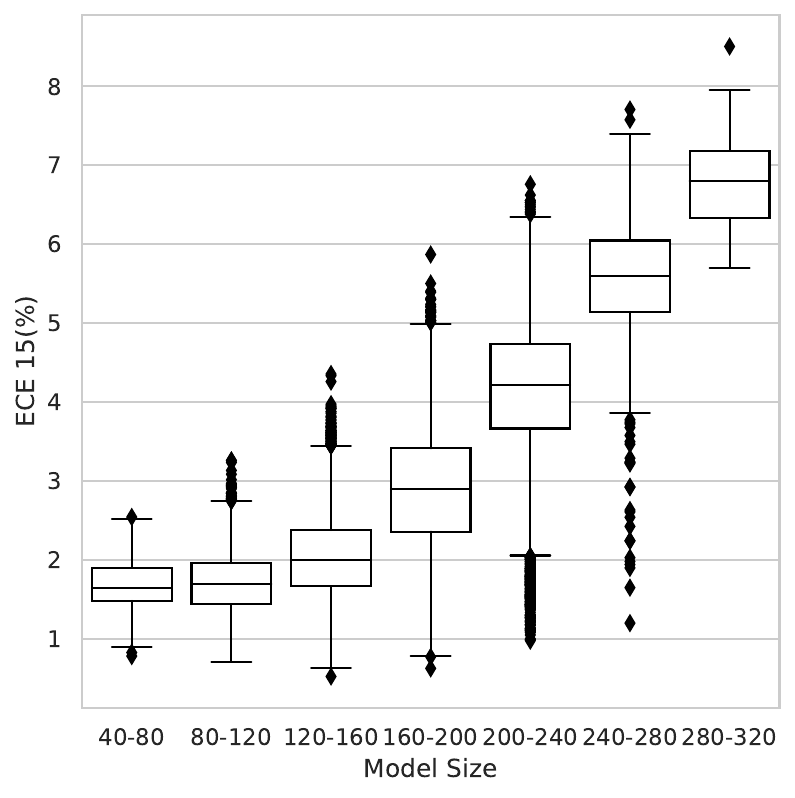}
        \caption{Box-plots of ECE across all architectures within SSS subset on ImagenNet, segregated by different size brackets.}
        \label{fig:imagenet_sss_singleplot}
    \end{subfigure}
    \hfill
    \begin{subfigure}{0.66\textwidth}
        \centering
        \vspace{-0.1in}
        \includegraphics[width=\textwidth]{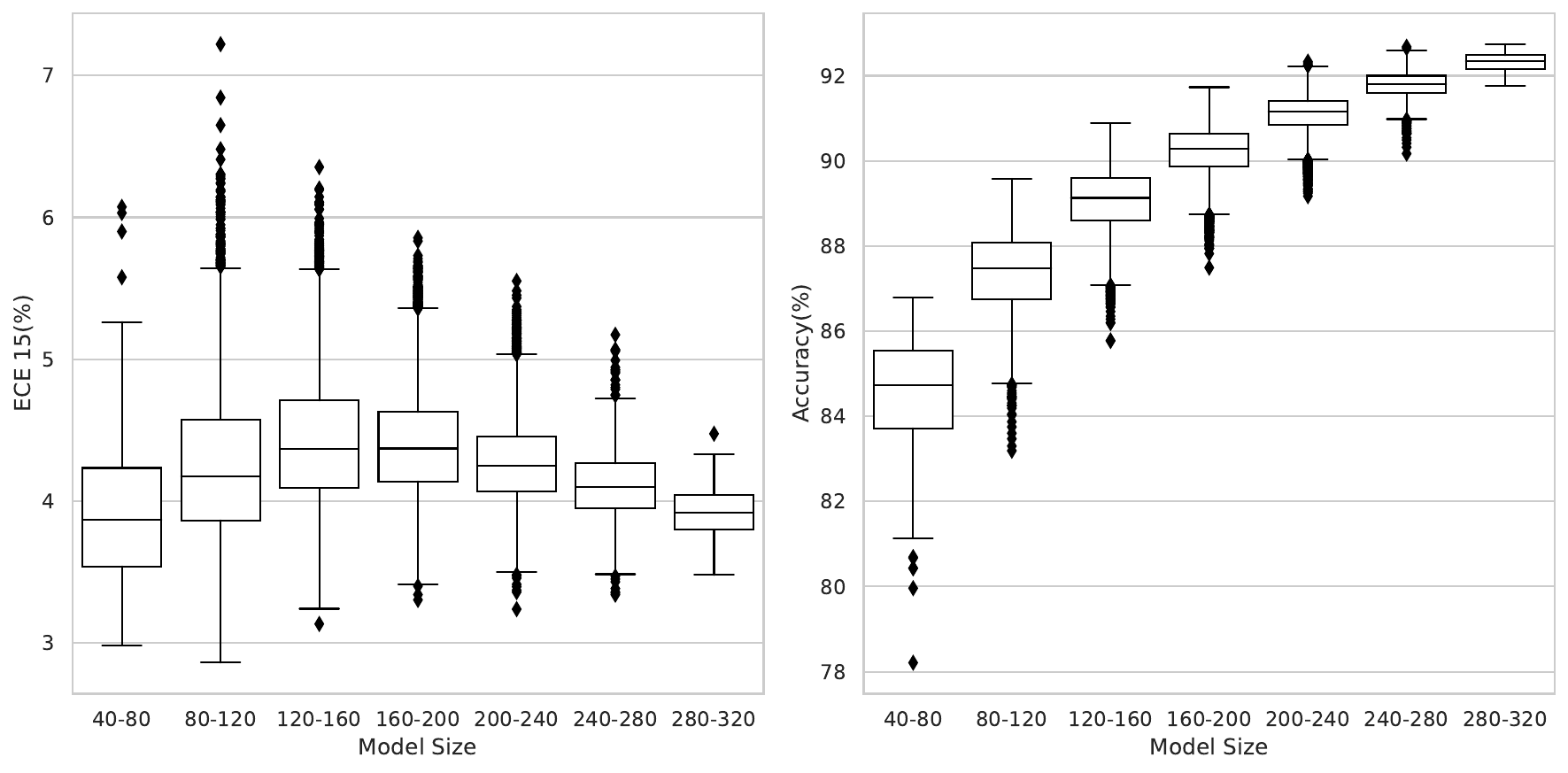}
        \caption{Boxplots illustrating ECE and Accuracy across all architectural within SSS subset, segregated by different size brackets. Each plot represents the performance evaluation on CIFAR-10. The ``model size” is defined as total number of kernels in each layer.}
        \label{fig:cifar10_sss}
    \end{subfigure}
    \caption{Explore the impact of model size on calibration performance.}
    \label{fig:model_size_calibration}
    \vspace{-0.1in}
\end{figure}

\begin{figure}[!ht]
    \centering
    \vspace{-0.1in}
    \includegraphics[width=0.95\textwidth]{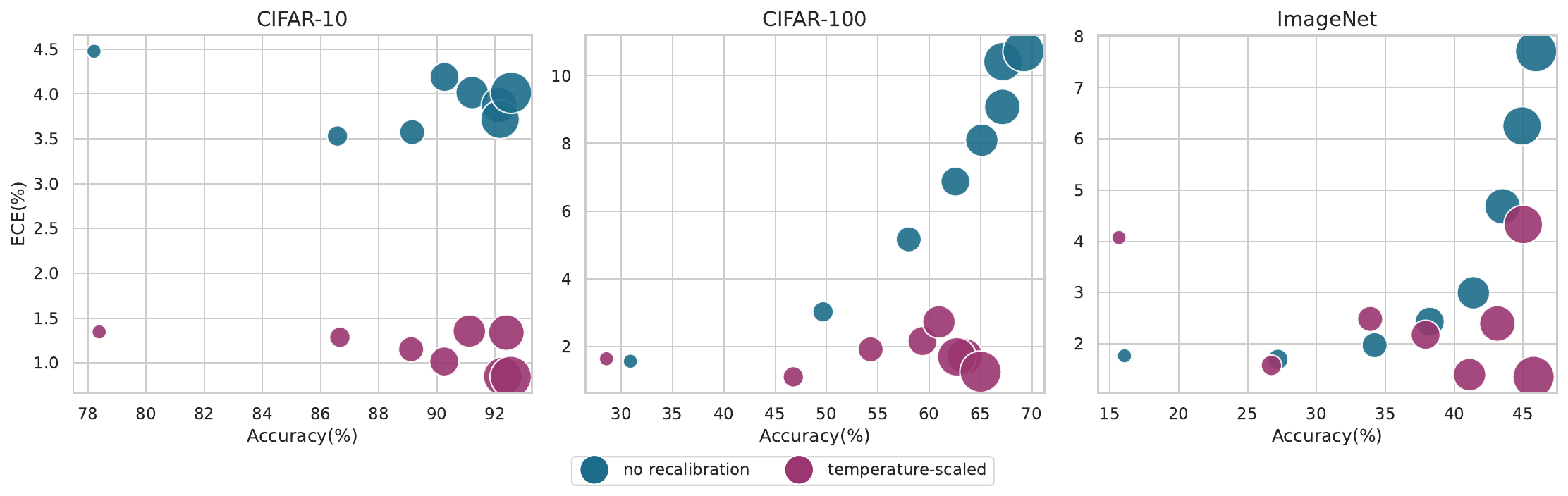}
    \caption{ECE measured on CIFAR-10, CIFAR-100, and ImageNet datasets before and after applying temperature scaling. Marker size represents model size progression from 8:8:8:8:8, 16:16:16:16:16, up to 64:64:64:64:64, where number indicate the number of kernels in a certain layer.}
    \label{fig:sss_width}
    \vspace{-0.1in}
\end{figure}

\begin{figure}[!ht]
    \centering
    \vspace{-0.1in}
    \includegraphics[width=0.95\textwidth]{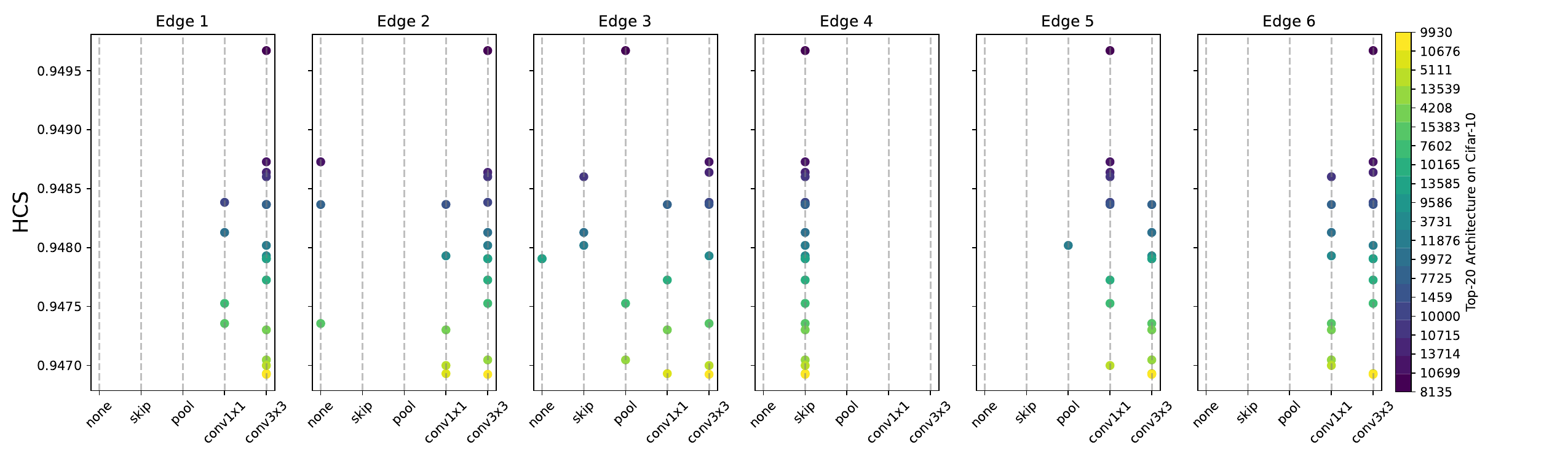}
       \vspace{-0.1in}
   \caption{Top 20 HCS architectures out of the topology search space according to the ECE on CIFAR-10 dataset. $\beta$ is set to 1.}
    \label{fig:arch_analysis}
    \vspace{-0.1in}
\end{figure}

\noindent \textbf{Experiment Setup:}~In this study, we focus on our SSS dataset for ImageNet and CIFAR-10, segmented by different size brackets based on the total number of kernels in each layer, as shown in Figure~\ref{fig:model_size_calibration}. Additionally, the performance of models of different width on CIFAR-10, CIFAR-100, and ImageNet was plotted to compare the impact of model width on different datasets, as shown in Figure~\ref{fig:sss_width}. 
We also choose top-ranked model in terms of our proposed metric to analyze the architecture design preference in Figure~\ref{fig:arch_analysis}.

\begin{figure}[!ht]
\begin{center}
\vspace{-0.1in}
   \includegraphics[width=0.95\linewidth]{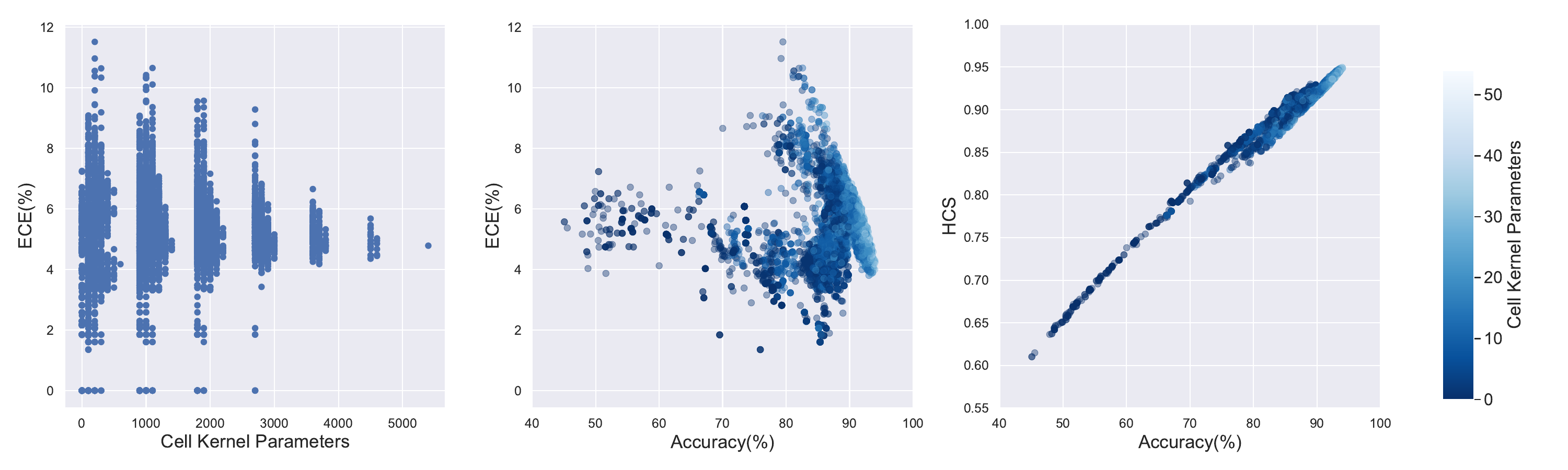}
\end{center}
\vspace{-0.1in}
   \caption{Calibration performance with cell kernel parameters. (Left) ECE distribution across different cell kernel parameters; (Middle) Scatter plot of all models on ECE and Accuracy; (Right): Scatter plot of all models on HCS and Accuracy. $\beta$ is set to 1.}
\label{fig:cell_kernel_and_HCS}
\vspace{-0.1in}
\end{figure}
\noindent \textbf{Discussion:} As shown in Figure~\ref{fig:model_size_calibration}, the results show that larger models do not necessarily have worse calibration performance, as it initially declines for the first few size brackets, but subsequently improves with increasing model size for CIFAR-10, while it worsens for ImageNet. Therefore, \textbf{calibration performance may depend on both model size and dataset complexity.} 
Additionally, the study found that fewer parameters do not necessarily lead to better calibration performance, as a larger kernel size does not necessarily result in worse calibration performance but instead results in a smaller calibration performance variance as depicted in the left plot of Figure~\ref{fig:cell_kernel_and_HCS}. The conclusion of a prior study that calibration decreases with model size may be due to the survivorship bias in calibration, where architectures with worse performance are often ignored, and the literature overlooks the fact that \textbf{less parameters do not necessarily lead to better calibration performance.}
Following our examination of the overall model size, we next direct our attention towards understanding the impact of the model's width, specifically the number of filters per layer, on its calibration properties. As depicted in Figure~\ref{fig:sss_width}, on the CIFAR-10 dataset, the width of the model appears to have a modest impact on calibration. However, as the dataset complexity increases, particularly with ImageNet, we observe a dramatic surge in ECE, rising from 3\% to a substantial 8\%. This suggests that \textbf{wider model can worsen calibration performance, especially when dealing with complex datasets.}

To analyze the best calibrated model, it is not appropriate to directly choose the model with the lowest ECE, since it may have unsatisfactory prediction accuracy performance. To address this issue, we propose a combined metric that takes into account both calibration and accuracy, called the Harmonic Calibration Score (HCS), denoted as $\text{HCS}_{\beta} = (1+\beta) \cdot \frac{(\text{Acc} \cdot (1-\text{ECE}))}{(\beta \text{Acc} + (1-\text{ECE}))}$, where $\beta$ controls the balance between accuracy and calibration with larger $\beta$ prefer more on calibration.
Figure~\ref{fig:cell_kernel_and_HCS} illustrates the distribution of models with respect to ECE and accuracy, where the best ECE is not always obtained at the same time as the highest accuracy. However, our proposed HCS provides a good trade-off between accuracy and calibration, and the model with the highest HCS indicates a better balance between calibration and prediction accuracy. After applying the HCS metric and replotting, a clear trend can be observed where the HCS increases as the cell kernel parameters increase. 
We subsequently selected the top 20 models based on their HCS scores from TSS and performed an architectural analysis of each edge, as shown in Figure~\ref{fig:arch_analysis}. This analysis shows that well-calibrated models tend to prefer either $1 \times 1$ convolution or $3 \times 3$ convolution for edges 1, 5, and 6. Intriguingly, we observed a unanimous preference for the skip connection for edge 4 among these well-calibrated models. These observations offer valuable insights into the design preferences of well-calibrated models, which can guide the development of future neural network architectures.

\section{Conclusions}
While our investigation provides significant insights into calibration studies, it is essential to note its specific scope. Our findings predominantly relate to image classifications, implying that extrapolation to other domains should be approached with caution. Within our analysis, we emphasized the post-hoc Temperature Scaling technique; however, the vast domain of post-hoc calibration houses numerous other techniques that may present divergent perspectives or different effects on model calibration. Importantly, our conclusions are primarily empirical. A theoretical exploration remains outside the purview of this work. Future research could remedy these confines by widening the spectrum of evaluated tasks and architectures and diving deeper into theoretical dissections.

\newpage
\section*{Acknowledgements}
We thank Yang You and Yifeng Gao for valuable discussions and feedbacks. This work was supported in part by the Australian Research Council under Projects DP240101848 and FT230100549.
\small
\bibliography{iclr2024_conference}

\begin{thebibliography}{68}
\providecommand{\natexlab}[1]{#1}
\providecommand{\url}[1]{\texttt{#1}}
\expandafter\ifx\csname urlstyle\endcsname\relax
  \providecommand{\doi}[1]{doi: #1}\else
  \providecommand{\doi}{doi: \begingroup \urlstyle{rm}\Url}\fi

\bibitem[Ashukha et~al.(2020)Ashukha, Lyzhov, Molchanov, and
  Vetrov]{ashukha2020pitfalls}
Arsenii Ashukha, Alexander Lyzhov, Dmitry Molchanov, and Dmitry Vetrov.
\newblock Pitfalls of in-domain uncertainty estimation and ensembling in deep
  learning.
\newblock \emph{arXiv preprint arXiv:2002.06470}, 2020.

\bibitem[Brier(1950)]{brier1950verification}
Glenn~W Brier.
\newblock Verification of forecasts expressed in terms of probability.
\newblock \emph{Monthly Weather Review}, 78\penalty0 (1):\penalty0 1--3, 1950.

\bibitem[Chen et~al.(2020)Chen, Kornblith, Norouzi, and Hinton]{chen2020simple}
Ting Chen, Simon Kornblith, Mohammad Norouzi, and Geoffrey Hinton.
\newblock A simple framework for contrastive learning of visual
  representations.
\newblock In \emph{International conference on machine learning}, pp.\
  1597--1607. PMLR, 2020.

\bibitem[Chrabaszcz et~al.(2017)Chrabaszcz, Loshchilov, and
  Hutter]{chrabaszcz2017downsampled}
Patryk Chrabaszcz, Ilya Loshchilov, and Frank Hutter.
\newblock A downsampled variant of imagenet as an alternative to the cifar
  datasets.
\newblock \emph{CoRR}, abs/1707.08819, 2017.

\bibitem[Deng et~al.(2009)Deng, Dong, Socher, Li, Li, and
  Fei-Fei]{deng2009imagenet}
Jia Deng, Wei Dong, Richard Socher, Li-Jia Li, Kai Li, and Li~Fei-Fei.
\newblock Imagenet: A large-scale hierarchical image database.
\newblock In \emph{2009 IEEE conference on computer vision and pattern
  recognition}, pp.\  248--255. Ieee, 2009.

\bibitem[Deng \& Zhang(2021)Deng and Zhang]{deng2021learning}
Xiang Deng and Zhongfei Zhang.
\newblock Learning with retrospection.
\newblock In \emph{Proceedings of the AAAI Conference on Artificial
  Intelligence}, volume~35, pp.\  7201--7209, 2021.

\bibitem[Dong \& Yang(2019{\natexlab{a}})Dong and Yang]{dong2019one}
Xuanyi Dong and Yi~Yang.
\newblock One-shot neural architecture search via self-evaluated template
  network.
\newblock In \emph{Proceedings of the IEEE/CVF International Conference on
  Computer Vision}, pp.\  3681--3690, 2019{\natexlab{a}}.

\bibitem[Dong \& Yang(2019{\natexlab{b}})Dong and Yang]{dong2019searching}
Xuanyi Dong and Yi~Yang.
\newblock Searching for a robust neural architecture in four gpu hours.
\newblock In \emph{Proceedings of the IEEE/CVF Conference on Computer Vision
  and Pattern Recognition}, pp.\  1761--1770, 2019{\natexlab{b}}.

\bibitem[Dong \& Yang(2020)Dong and Yang]{dong2020bench}
Xuanyi Dong and Yi~Yang.
\newblock Nas-bench-201: Extending the scope of reproducible neural
  architecture search.
\newblock \emph{arXiv preprint arXiv:2001.00326}, 2020.

\bibitem[Dong et~al.(2021)Dong, Liu, Musial, and Gabrys]{dong2021nats}
Xuanyi Dong, Lu~Liu, Katarzyna Musial, and Bogdan Gabrys.
\newblock Nats-bench: Benchmarking nas algorithms for architecture topology and
  size.
\newblock \emph{IEEE transactions on pattern analysis and machine
  intelligence}, 44\penalty0 (7):\penalty0 3634--3646, 2021.

\bibitem[Dosovitskiy et~al.(2020{\natexlab{a}})Dosovitskiy, Beyer, Kolesnikov,
  Weissenborn, Zhai, and Unterthiner]{dosovitskiy2020transformers}
A~Dosovitskiy, L~Beyer, A~Kolesnikov, D~Weissenborn, X~Zhai, and T~Unterthiner.
\newblock Transformers for image recognition at scale.
\newblock \emph{arXiv preprint arXiv:2010.11929}, 2020{\natexlab{a}}.

\bibitem[Dosovitskiy et~al.(2020{\natexlab{b}})Dosovitskiy, Beyer, Kolesnikov,
  Weissenborn, Zhai, Unterthiner, Dehghani, Minderer, Heigold, Gelly,
  et~al.]{dosovitskiy2020image}
Alexey Dosovitskiy, Lucas Beyer, Alexander Kolesnikov, Dirk Weissenborn,
  Xiaohua Zhai, Thomas Unterthiner, Mostafa Dehghani, Matthias Minderer, Georg
  Heigold, Sylvain Gelly, et~al.
\newblock An image is worth 16x16 words: Transformers for image recognition at
  scale.
\newblock \emph{arXiv preprint arXiv:2010.11929}, 2020{\natexlab{b}}.

\bibitem[Feng et~al.(2019)Feng, Rosenbaum, Glaeser, Timm, and
  Dietmayer]{feng2019can}
Di~Feng, Lars Rosenbaum, Claudius Glaeser, Fabian Timm, and Klaus Dietmayer.
\newblock Can we trust you? on calibration of a probabilistic object detector
  for autonomous driving.
\newblock \emph{arXiv preprint arXiv:1909.12358}, 2019.

\bibitem[Gelman \& Shalizi(2013)Gelman and Shalizi]{gelman2013philosophy}
Andrew Gelman and Cosma~Rohilla Shalizi.
\newblock Philosophy and the practice of bayesian statistics.
\newblock \emph{British Journal of Mathematical and Statistical Psychology},
  66\penalty0 (1):\penalty0 8--38, 2013.

\bibitem[Guo et~al.(2017)Guo, Pleiss, Sun, and Weinberger]{guo2017calibration}
Chuan Guo, Geoff Pleiss, Yu~Sun, and Kilian~Q Weinberger.
\newblock On calibration of modern neural networks.
\newblock In \emph{International conference on machine learning}, pp.\
  1321--1330. PMLR, 2017.

\bibitem[Gupta \& Ramdas(2021)Gupta and Ramdas]{gupta2021distribution}
Chirag Gupta and Aaditya Ramdas.
\newblock Distribution-free calibration guarantees for histogram binning
  without sample splitting.
\newblock In \emph{International Conference on Machine Learning}, pp.\
  3942--3952. PMLR, 2021.

\bibitem[Gupta et~al.(2020)Gupta, Rahimi, Ajanthan, Mensink, Sminchisescu, and
  Hartley]{gupta2020calibration}
Kriti Gupta, Amir Rahimi, Thalaiyasingam Ajanthan, Thomas Mensink, Cristian
  Sminchisescu, and Richard Hartley.
\newblock Calibration of neural networks using splines.
\newblock In \emph{International Conference on Learning Representations}, 2020.

\bibitem[He et~al.(2016)He, Zhang, Ren, and Sun]{he2016deep}
Kaiming He, Xiangyu Zhang, Shaoqing Ren, and Jian Sun.
\newblock Deep residual learning for image recognition.
\newblock In \emph{Proceedings of the IEEE conference on computer vision and
  pattern recognition}, pp.\  770--778, 2016.

\bibitem[Hendrycks \& Dietterich(2019)Hendrycks and
  Dietterich]{hendrycks2019benchmarking}
Dan Hendrycks and Thomas Dietterich.
\newblock Benchmarking neural network robustness to common corruptions and
  perturbations.
\newblock \emph{arXiv preprint arXiv:1903.12261}, 2019.

\bibitem[Hendrycks et~al.(2019)Hendrycks, Mu, Cubuk, Zoph, Gilmer, and
  Lakshminarayanan]{hendrycks2019augmix}
Dan Hendrycks, Norman Mu, Ekin~D Cubuk, Barret Zoph, Justin Gilmer, and Balaji
  Lakshminarayanan.
\newblock Augmix: A simple data processing method to improve robustness and
  uncertainty.
\newblock \emph{arXiv preprint arXiv:1912.02781}, 2019.

\bibitem[Huang et~al.(2017)Huang, Liu, Van Der~Maaten, and
  Weinberger]{huang2017densely}
Gao Huang, Zhuang Liu, Laurens Van Der~Maaten, and Kilian~Q Weinberger.
\newblock Densely connected convolutional networks.
\newblock In \emph{Proceedings of the IEEE conference on computer vision and
  pattern recognition}, pp.\  4700--4708, 2017.

\bibitem[Jung et~al.(2023)Jung, Lukasik, and Keuper]{jung2023neural}
Steffen Jung, Jovita Lukasik, and Margret Keuper.
\newblock Neural architecture design and robustness: A dataset.
\newblock In \emph{The Eleventh International Conference on Learning
  Representations}, 2023.

\bibitem[Karandikar et~al.(2021)Karandikar, Cain, Tran, Lakshminarayanan,
  Shlens, Mozer, and Roelofs]{karandikar2021soft}
Archit Karandikar, Nicholas Cain, Dustin Tran, Balaji Lakshminarayanan,
  Jonathon Shlens, Michael~C Mozer, and Becca Roelofs.
\newblock Soft calibration objectives for neural networks.
\newblock \emph{Advances in Neural Information Processing Systems},
  34:\penalty0 29768--29779, 2021.

\bibitem[Kendall(1938)]{kendall1938new}
Maurice~G. Kendall.
\newblock A new measure of rank correlation.
\newblock \emph{Biometrika}, 30\penalty0 (1/2):\penalty0 81--93, 1938.

\bibitem[Kim et~al.(2021)Kim, Ji, Yoon, and Hwang]{kim2021self}
Kyungyul Kim, ByeongMoon Ji, Doyoung Yoon, and Sangheum Hwang.
\newblock Self-knowledge distillation with progressive refinement of targets.
\newblock In \emph{Proceedings of the IEEE/CVF International Conference on
  Computer Vision}, pp.\  6567--6576, 2021.

\bibitem[Krishnan \& Tickoo(2020)Krishnan and Tickoo]{krishnan2020improving}
Ranganath Krishnan and Omesh Tickoo.
\newblock Improving model calibration with accuracy versus uncertainty
  optimization.
\newblock \emph{Advances in Neural Information Processing Systems},
  33:\penalty0 18237--18248, 2020.

\bibitem[Krizhevsky et~al.(2009)Krizhevsky, Hinton,
  et~al.]{krizhevsky2009learning}
Alex Krizhevsky, Geoffrey Hinton, et~al.
\newblock Learning multiple layers of features from tiny images.
\newblock \emph{arXiv preprint arXiv:1806.09055}, 2009.

\bibitem[Kull et~al.(2019)Kull, Perello~Nieto, K{"a}ngsepp, Silva~Filho, Song,
  and Flach]{kull2019beyond}
Meelis Kull, Marc Perello~Nieto, Markus K{"a}ngsepp, Telmo Silva~Filho, Hao
  Song, and Peter Flach.
\newblock Beyond temperature scaling: Obtaining well-calibrated multi-class
  probabilities with dirichlet calibration.
\newblock In \emph{Advances in Neural Information Processing Systems},
  volume~32, 2019.

\bibitem[Kumar et~al.(2019)Kumar, Liang, and Ma]{kumar2019verified}
Ananya Kumar, Percy~S. Liang, and Tengyu Ma.
\newblock Verified uncertainty calibration.
\newblock In \emph{Advances in Neural Information Processing Systems},
  volume~32, 2019.

\bibitem[Kumar et~al.(2018)Kumar, Sarawagi, and Jain]{kumar2018trainable}
Aviral Kumar, Sunita Sarawagi, and Ujjwal Jain.
\newblock Trainable calibration measures for neural networks from kernel mean
  embeddings.
\newblock In \emph{International Conference on Machine Learning}, pp.\
  2805--2814. PMLR, 2018.

\bibitem[Li \& Talwalkar(2020)Li and Talwalkar]{li2020random}
Liam Li and Ameet Talwalkar.
\newblock Random search and reproducibility for neural architecture search.
\newblock In \emph{Uncertainty in artificial intelligence}, pp.\  367--377.
  PMLR, 2020.

\bibitem[Liu et~al.(2018)Liu, Simonyan, and Yang]{liu2018darts}
Hanxiao Liu, Karen Simonyan, and Yiming Yang.
\newblock Darts: Differentiable architecture search.
\newblock \emph{arXiv preprint arXiv:1806.09055}, 2018.

\bibitem[Liu et~al.(2021)Liu, Lin, Cao, Hu, Wei, Zhang, Lin, and
  Guo]{liu2021swin}
Ze~Liu, Yutong Lin, Yue Cao, Han Hu, Yixuan Wei, Zheng Zhang, Stephen Lin, and
  Baining Guo.
\newblock Swin transformer: Hierarchical vision transformer using shifted
  windows.
\newblock In \emph{Proceedings of the IEEE/CVF international conference on
  computer vision}, pp.\  10012--10022, 2021.

\bibitem[Minderer et~al.(2021)Minderer, Djolonga, Romijnders, Hubis, Zhai,
  Houlsby, Tran, and Lucic]{minderer2021revisiting}
Matthias Minderer, Josip Djolonga, Rob Romijnders, Frances Hubis, Xiaohua Zhai,
  Neil Houlsby, Dustin Tran, and Mario Lucic.
\newblock Revisiting the calibration of modern neural networks.
\newblock \emph{Advances in Neural Information Processing Systems},
  34:\penalty0 15682--15694, 2021.

\bibitem[Mukhoti et~al.(2020)Mukhoti, Kulharia, Sanyal, Golodetz, Torr, and
  Dokania]{mukhoti2020calibrating}
Jishnu Mukhoti, Viveka Kulharia, Amartya Sanyal, Stuart Golodetz, Philip Torr,
  and Puneet Dokania.
\newblock Calibrating deep neural networks using focal loss.
\newblock \emph{Advances in Neural Information Processing Systems},
  33:\penalty0 15288--15299, 2020.

\bibitem[M{\"u}ller et~al.(2019)M{\"u}ller, Kornblith, and
  Hinton]{muller2019does}
Rafael M{\"u}ller, Simon Kornblith, and Geoffrey~E Hinton.
\newblock When does label smoothing help?
\newblock \emph{Advances in neural information processing systems}, 32, 2019.

\bibitem[Naeini et~al.(2015)Naeini, Cooper, and
  Hauskrecht]{naeini2015obtaining}
Mahdi~Pakdaman Naeini, Gregory~F. Cooper, and Milos Hauskrecht.
\newblock Obtaining well calibrated probabilities using bayesian binning.
\newblock In \emph{Proceedings of the Twenty-Ninth AAAI Conference on
  Artificial Intelligence}, pp.\  2901--2907. AAAI Press, 2015.
\newblock ISBN 0262511290.

\bibitem[Netzer et~al.(2011)Netzer, Wang, Coates, Bissacco, Wu, and
  Ng]{netzer2011reading}
Yuval Netzer, Tao Wang, Adam Coates, Alessandro Bissacco, Bo~Wu, and Andrew~Y
  Ng.
\newblock Reading digits in natural images with unsupervised feature learning.
\newblock \emph{arXiv preprint arXiv:1903.12261}, 2011.

\bibitem[Nixon et~al.(2019)]{nixon2019measuring}
Jeremy Nixon et~al.
\newblock Measuring calibration in deep learning.
\newblock In \emph{CVPR workshops}, volume~2, 2019.

\bibitem[Patel et~al.(2020)Patel, Beluch, Yang, Pfeiffer, and
  Zhang]{patel2020multi}
Kanil Patel, William Beluch, Bin Yang, Michael Pfeiffer, and Dan Zhang.
\newblock Multi-class uncertainty calibration via mutual information
  maximization-based binning.
\newblock \emph{arXiv preprint arXiv:2006.13092}, 2020.

\bibitem[Radford et~al.(2021)Radford, Kim, Hallacy, Ramesh, Goh, Agarwal,
  Sastry, Askell, Mishkin, Clark, et~al.]{radford2021learning}
Alec Radford, Jong~Wook Kim, Chris Hallacy, Aditya Ramesh, Gabriel Goh,
  Sandhini Agarwal, Girish Sastry, Amanda Askell, Pamela Mishkin, Jack Clark,
  et~al.
\newblock Learning transferable visual models from natural language
  supervision.
\newblock In \emph{International conference on machine learning}, pp.\
  8748--8763. PMLR, 2021.

\bibitem[Ranftl et~al.(2021)Ranftl, Bochkovskiy, and Koltun]{ranftl2021vision}
Ren{\'e} Ranftl, Alexey Bochkovskiy, and Vladlen Koltun.
\newblock Vision transformers for dense prediction.
\newblock In \emph{Proceedings of the IEEE/CVF International Conference on
  Computer Vision}, pp.\  12179--12188, 2021.

\bibitem[Real et~al.(2019)Real, Aggarwal, Huang, and Le]{real2019regularized}
Esteban Real, Alok Aggarwal, Yanping Huang, and Quoc~V Le.
\newblock Regularized evolution for image classifier architecture search.
\newblock In \emph{Proceedings of the aaai conference on artificial
  intelligence}, volume~33, pp.\  4780--4789, 2019.

\bibitem[Roelofs et~al.(2022)Roelofs, Cain, Shlens, and
  Mozer]{roelofs2022mitigating}
Rebecca Roelofs, Nicholas Cain, Jonathon Shlens, and Michael~C Mozer.
\newblock Mitigating bias in calibration error estimation.
\newblock In \emph{International Conference on Artificial Intelligence and
  Statistics}, pp.\  4036--4054. PMLR, 2022.

\bibitem[Schuhmann et~al.(2021)Schuhmann, Vencu, Beaumont, Kaczmarczyk, Mullis,
  Katta, Coombes, Jitsev, and Komatsuzaki]{schuhmann2021laion}
Christoph Schuhmann, Richard Vencu, Romain Beaumont, Robert Kaczmarczyk,
  Clayton Mullis, Aarush Katta, Theo Coombes, Jenia Jitsev, and Aran
  Komatsuzaki.
\newblock Laion-400m: Open dataset of clip-filtered 400 million image-text
  pairs.
\newblock \emph{arXiv preprint arXiv:2111.02114}, 2021.

\bibitem[Schuhmann et~al.(2022)Schuhmann, Beaumont, Vencu, Gordon, Wightman,
  Cherti, Coombes, Katta, Mullis, Wortsman, et~al.]{schuhmann2022laion}
Christoph Schuhmann, Romain Beaumont, Richard Vencu, Cade Gordon, Ross
  Wightman, Mehdi Cherti, Theo Coombes, Aarush Katta, Clayton Mullis, Mitchell
  Wortsman, et~al.
\newblock Laion-5b: An open large-scale dataset for training next generation
  image-text models.
\newblock \emph{Advances in Neural Information Processing Systems},
  35:\penalty0 25278--25294, 2022.

\bibitem[Siems et~al.(2020)Siems, Zimmer, Zela, Lukasik, Keuper, and
  Hutter]{DBLP:journals/corr/abs-2008-09777}
Julien Siems, Lucas Zimmer, Arber Zela, Jovita Lukasik, Margret Keuper, and
  Frank Hutter.
\newblock Nas-bench-301 and the case for surrogate benchmarks for neural
  architecture search.
\newblock \emph{CoRR}, abs/2008.09777, 2020.
\newblock URL \url{https://arxiv.org/abs/2008.09777}.

\bibitem[Su et~al.(2022)Su, You, Xie, Zheng, Wang, Qian, Zhang, Wang, and
  Xu]{su2022vitas}
Xiu Su, Shan You, Jiyang Xie, Mingkai Zheng, Fei Wang, Chen Qian, Changshui
  Zhang, Xiaogang Wang, and Chang Xu.
\newblock Vitas: Vision transformer architecture search.
\newblock In \emph{European Conference on Computer Vision}, pp.\  139--157.
  Springer Nature Switzerland Cham, 2022.

\bibitem[Tao et~al.(2023{\natexlab{a}})Tao, Dong, Liu, Sun, and
  Xu]{tao2023calibrating}
Linwei Tao, Minjing Dong, Daochang Liu, Changming Sun, and Chang Xu.
\newblock Calibrating a deep neural network with its predecessors.
\newblock \emph{arXiv preprint arXiv:2302.06245}, 2023{\natexlab{a}}.

\bibitem[Tao et~al.(2023{\natexlab{b}})Tao, Dong, and Xu]{tao2023dual}
Linwei Tao, Minjing Dong, and Chang Xu.
\newblock Dual focal loss for calibration.
\newblock \emph{arXiv preprint arXiv:2305.13665}, 2023{\natexlab{b}}.

\bibitem[Thiagarajan et~al.(2022)Thiagarajan, Thopalli, Rajan, and
  Turaga]{thiagarajan2022training}
Jayaraman~J Thiagarajan, Kowshik Thopalli, Deepta Rajan, and Pavan Turaga.
\newblock Training calibration-based counterfactual explainers for deep
  learning models in medical image analysis.
\newblock \emph{Scientific reports}, 12\penalty0 (1):\penalty0 597, 2022.

\bibitem[Thomee et~al.(2016)Thomee, Shamma, Friedland, Elizalde, Ni, Poland,
  Borth, and Li]{thomee2016yfcc100m}
Bart Thomee, David~A Shamma, Gerald Friedland, Benjamin Elizalde, Karl Ni,
  Douglas Poland, Damian Borth, and Li-Jia Li.
\newblock Yfcc100m: The new data in multimedia research.
\newblock \emph{Communications of the ACM}, 59\penalty0 (2):\penalty0 64--73,
  2016.

\bibitem[Thulasidasan et~al.(2019)Thulasidasan, Chennupati, Bilmes,
  Bhattacharya, and Michalak]{thulasidasan2019mixup}
Sunil Thulasidasan, Gopinath Chennupati, Jeff~A Bilmes, Tanmoy Bhattacharya,
  and Sarah Michalak.
\newblock On mixup training: Improved calibration and predictive uncertainty
  for deep neural networks.
\newblock \emph{Advances in Neural Information Processing Systems}, 32, 2019.

\bibitem[Tolstikhin et~al.(2021)Tolstikhin, Houlsby, Kolesnikov, Beyer, Zhai,
  Unterthiner, Yung, Steiner, Keysers, Uszkoreit, et~al.]{tolstikhin2021mlp}
Ilya~O Tolstikhin, Neil Houlsby, Alexander Kolesnikov, Lucas Beyer, Xiaohua
  Zhai, Thomas Unterthiner, Jessica Yung, Andreas Steiner, Daniel Keysers,
  Jakob Uszkoreit, et~al.
\newblock Mlp-mixer: An all-mlp architecture for vision.
\newblock \emph{Advances in neural information processing systems},
  34:\penalty0 24261--24272, 2021.

\bibitem[Touvron et~al.(2021{\natexlab{a}})Touvron, Cord, Douze, Massa,
  Sablayrolles, and J{\'e}gou]{touvron2021training}
Hugo Touvron, Matthieu Cord, Matthijs Douze, Francisco Massa, Alexandre
  Sablayrolles, and Herv{\'e} J{\'e}gou.
\newblock Training data-efficient image transformers \& distillation through
  attention.
\newblock In \emph{International conference on machine learning}, pp.\
  10347--10357. PMLR, 2021{\natexlab{a}}.

\bibitem[Touvron et~al.(2021{\natexlab{b}})Touvron, Cord, Sablayrolles,
  Synnaeve, and J{\'e}gou]{touvron2021going}
Hugo Touvron, Matthieu Cord, Alexandre Sablayrolles, Gabriel Synnaeve, and
  Herv{\'e} J{\'e}gou.
\newblock Going deeper with image transformers.
\newblock In \emph{Proceedings of the IEEE/CVF international conference on
  computer vision}, pp.\  32--42, 2021{\natexlab{b}}.

\bibitem[Vaicenavicius et~al.(2019)Vaicenavicius, Widmann, Andersson, Lindsten,
  Roll, and Schön]{vaicenavicius2019evaluating}
Julius Vaicenavicius, David Widmann, Carl Andersson, Fredrik Lindsten, Jacob
  Roll, and Thomas Schön.
\newblock Evaluating model calibration in classification.
\newblock In \emph{The 22nd International Conference on Artificial Intelligence
  and Statistics}, pp.\  3459--3467. PMLR, 2019.

\bibitem[Wang et~al.(2021)Wang, Feng, and Zhang]{wang2021rethinking}
Deng-Bao Wang, Lei Feng, and Min-Ling Zhang.
\newblock Rethinking calibration of deep neural networks: Do not be afraid of
  overconfidence.
\newblock \emph{Advances in Neural Information Processing Systems},
  34:\penalty0 11809--11820, 2021.

\bibitem[Wang et~al.(2022)Wang, Xie, Li, Fan, Song, Liang, Lu, Luo, and
  Shao]{wang2022pvt}
Wenhai Wang, Enze Xie, Xiang Li, Deng-Ping Fan, Kaitao Song, Ding Liang, Tong
  Lu, Ping Luo, and Ling Shao.
\newblock Pvt v2: Improved baselines with pyramid vision transformer.
\newblock \emph{Computational Visual Media}, 8\penalty0 (3):\penalty0 415--424,
  2022.

\bibitem[White et~al.(2021)White, Nolen, and Savani]{white2021exploring}
Colin White, Sam Nolen, and Yash Savani.
\newblock Exploring the loss landscape in neural architecture search.
\newblock In \emph{Uncertainty in Artificial Intelligence}, pp.\  654--664.
  PMLR, 2021.

\bibitem[Xu et~al.(2019)Xu, Su, Wang, Cai, Cui, and Chen]{xu2019automatic}
Hongyan Xu, Xiu Su, Yi~Wang, Huaiyu Cai, Kerang Cui, and Xiaodong Chen.
\newblock Automatic bridge crack detection using a convolutional neural
  network.
\newblock \emph{Applied Sciences}, 9\penalty0 (14):\penalty0 2867, 2019.

\bibitem[Ying et~al.(2019)Ying, Klein, Real, Christiansen, Murphy, and
  Hutter]{ying2019nasbench101}
Chris Ying, Aaron Klein, Esteban Real, Eric Christiansen, Kevin Murphy, and
  Frank Hutter.
\newblock Nas-bench-101: Towards reproducible neural architecture search.
\newblock \emph{arXiv preprint arXiv:1902.09635}, 2019.

\bibitem[Yu et~al.(2022)Yu, Luo, Zhou, Si, Zhou, Wang, Feng, and
  Yan]{yu2022metaformer}
Weihao Yu, Mi~Luo, Pan Zhou, Chenyang Si, Yichen Zhou, Xinchao Wang, Jiashi
  Feng, and Shuicheng Yan.
\newblock Metaformer is actually what you need for vision.
\newblock In \emph{Proceedings of the IEEE/CVF conference on computer vision
  and pattern recognition}, pp.\  10819--10829, 2022.

\bibitem[Yuan et~al.(2021)Yuan, Chen, Wang, Yu, Shi, Jiang, Tay, Feng, and
  Yan]{yuan2021tokens}
Li~Yuan, Yunpeng Chen, Tao Wang, Weihao Yu, Yujun Shi, Zi-Hang Jiang,
  Francis~EH Tay, Jiashi Feng, and Shuicheng Yan.
\newblock Tokens-to-token vit: Training vision transformers from scratch on
  imagenet.
\newblock In \emph{Proceedings of the IEEE/CVF international conference on
  computer vision}, pp.\  558--567, 2021.

\bibitem[Zagoruyko \& Komodakis(2016)Zagoruyko and
  Komodakis]{zagoruyko2016wide}
Sergey Zagoruyko and Nikos Komodakis.
\newblock Wide residual networks.
\newblock \emph{arXiv preprint arXiv:1605.07146}, 2016.

\bibitem[Zhang et~al.(2019)Zhang, Yu, Jiao, Xing, El~Ghaoui, and
  Jordan]{zhang2019theoretically}
Hongyang Zhang, Yaodong Yu, Jiantao Jiao, Eric Xing, Laurent El~Ghaoui, and
  Michael Jordan.
\newblock Theoretically principled trade-off between robustness and accuracy.
\newblock In \emph{International conference on machine learning}, pp.\
  7472--7482. PMLR, 2019.

\bibitem[Zhang et~al.(2020)Zhang, Kailkhura, and Han]{zhang2020mixnmatch}
Jiayi Zhang, Bhavya Kailkhura, and Tony Y.-J. Han.
\newblock Mix-n-match: Ensemble and compositional methods for uncertainty
  calibration in deep learning.
\newblock In \emph{International Conference on Machine Learning}, pp.\
  11117--11128. PMLR, 2020.

\bibitem[Zhang et~al.(2022)Zhang, Deng, Kawaguchi, and Zou]{zhang2022and}
Linjun Zhang, Zhun Deng, Kenji Kawaguchi, and James Zou.
\newblock When and how mixup improves calibration.
\newblock In \emph{International Conference on Machine Learning}, pp.\
  26135--26160. PMLR, 2022.

\end{thebibliography}
\bibliographystyle{iclr2024_conference}

\newpage
\appendix
\section{Experiments on Transformers}
\label{sec:Experiments on Transformers}
To ensure the generality of our findings, we also include 11 vision transformers of different architecture design including 
T2T-ViT-7, T2T-ViT-10, T2T-ViT-12, T2T-ViT-19, T2T-ViT-24~\citep{yuan2021tokens}, ViT-b-16~\citep{dosovitskiy2020transformers}, Swin-T~\citep{liu2021swin}, Deit-T~\citep{touvron2021training}, Cait-XXS24~\citep{touvron2021going}, PvTv2-T~\citep{wang2022pvt} and PoolFormer-S12~\citep{yu2022metaformer}
. Each transformer is fine tuned 60 epochs on CIFAR-10, CIFAR-100 and ImageNet-16-120 based on the pretrained weights on ImageNet-1k. 
\begin{figure}[!ht]
\begin{center}
   \includegraphics[width=0.2\linewidth]{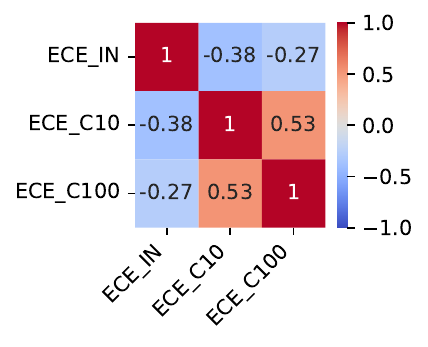}
\end{center}
    \caption{Kendall ranking correlation matrix of ECE for different Vision Transformer architectures.}
\label{fig:vit_ECE15_pre}
\end{figure}

As shown in Figure~\ref{fig:vit_ECE15_pre}, transformers show similar results as CNNs, where ECE on ImageNet shows little correlation with that on CIFAR-10 and CIFAR-100. The observation indicates that calibration property of a certain architecture can not generalize well to different datasets.

\begin{figure}[!ht]
\vspace{-0.1in}
    \centering
    \begin{subfigure}{0.32\textwidth}
        \centering
        \includegraphics[width=\textwidth]{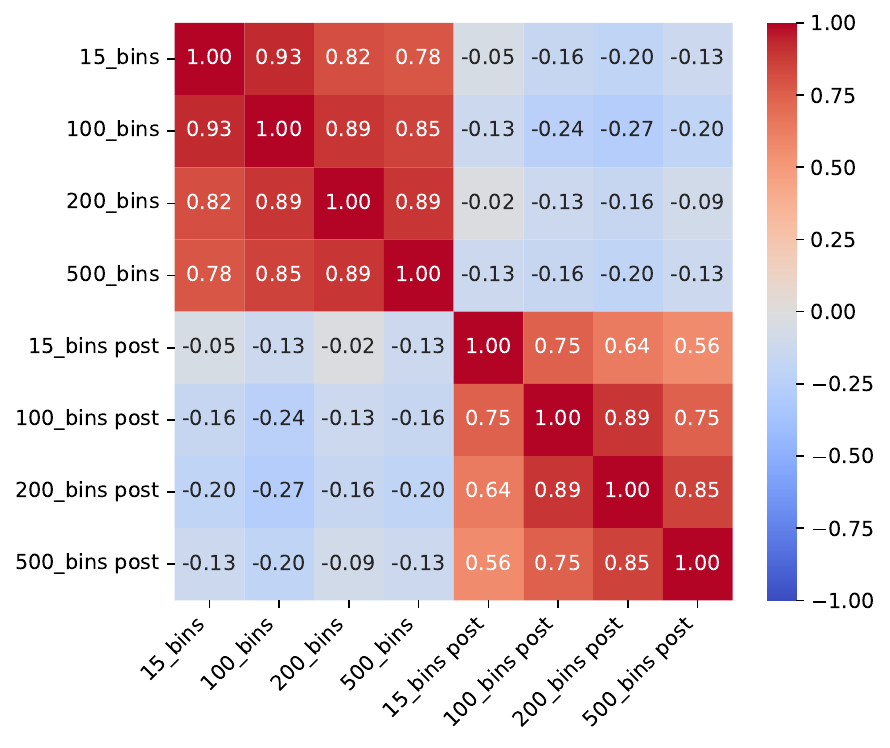}
        \label{fig:vit_temp_cifar10}
    \end{subfigure}
    \hfill    
    \begin{subfigure}{0.32\textwidth}
        \centering
        \includegraphics[width=\textwidth]{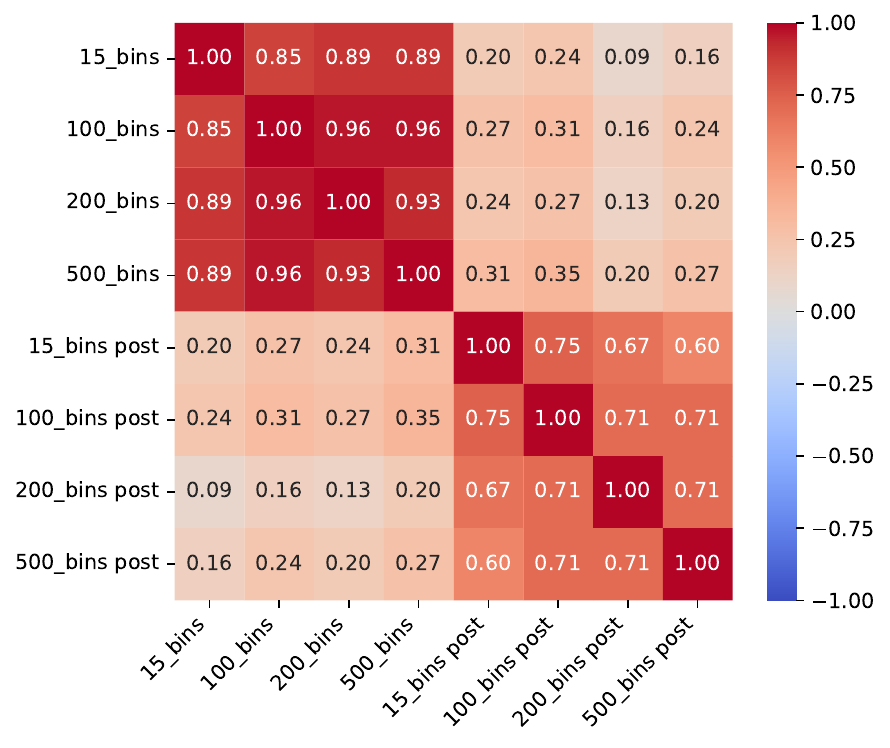}
        \label{fig:vit_temp_cifar100}
    \end{subfigure}
    \hfill
    \begin{subfigure}{0.32\textwidth}
        \centering
        \includegraphics[width=\textwidth]{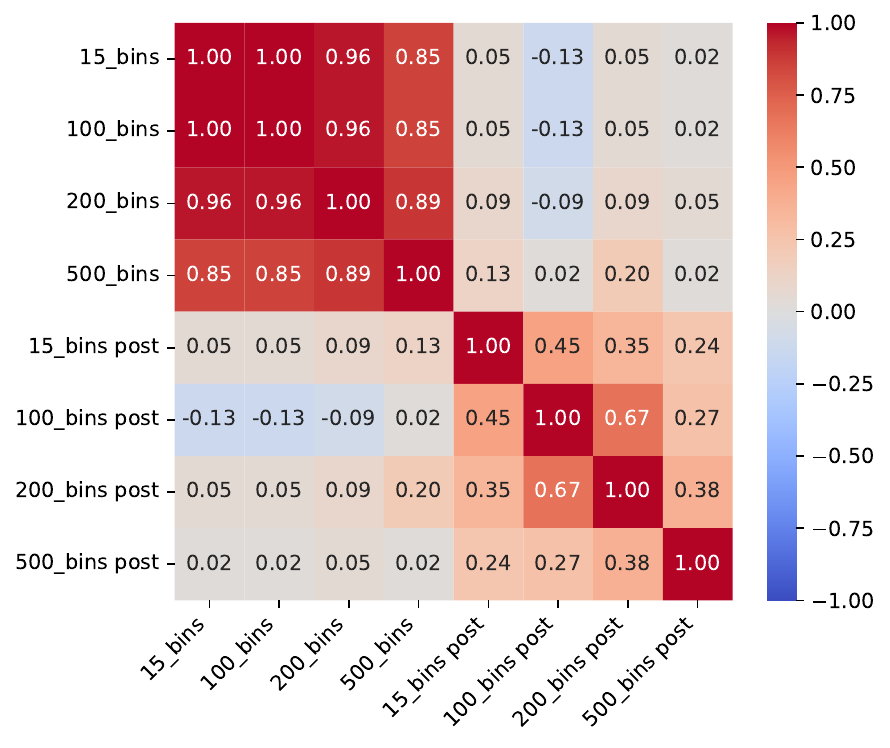}
        \label{fig:vit_temp_IN}
    \end{subfigure}
    \hfill
    \caption{Kendall Ranking Correlation Matrix of ECE before and after temperature scaling on CIFAR-10 (left), CIFAR-100 (middle) and ImageNet (right).}
    \label{fig:corr_beforeandafter_Temp}
\end{figure}

As shown in Figure~\ref{fig:corr_beforeandafter_Temp}, ECE evaluated on different bin size shows little correlation between pre and post temperature scaling. It indicate that well-calibrated models do not necessarily exhibit better calibration performance after post-hoc calibration techniques.

\begin{figure}[!ht]
\vspace{-0.1in}
    \centering
    \begin{subfigure}{0.32\textwidth}
        \centering
        \includegraphics[width=\textwidth]{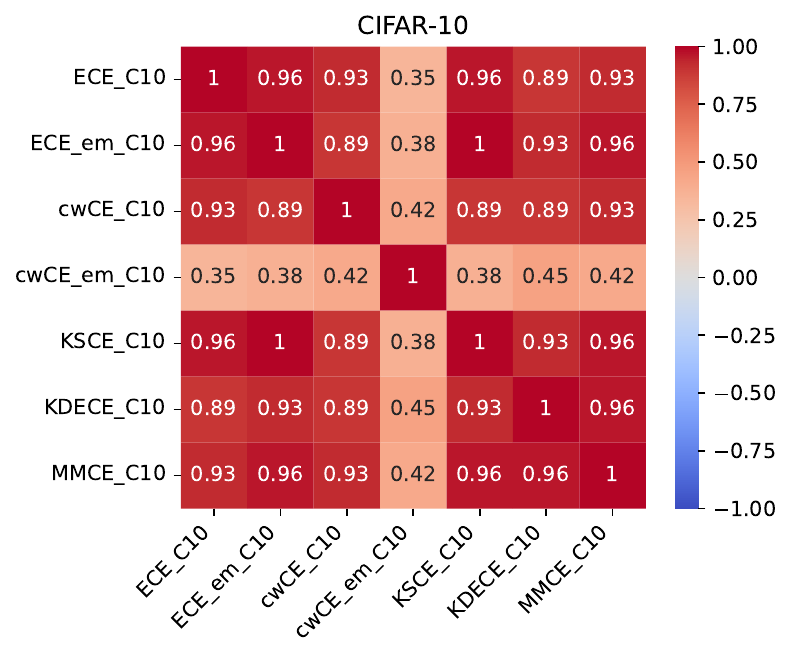}
        \label{fig:vit_cifar10_pre}
    \end{subfigure}
    \begin{subfigure}{0.32\textwidth}
        \centering
        \includegraphics[width=\textwidth]{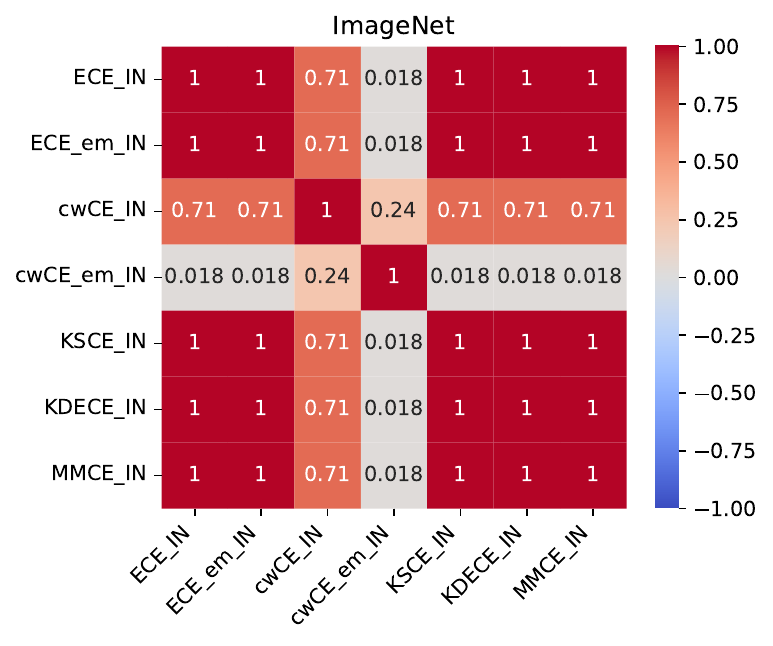}
        \label{fig:vit_imagenet_pre}
    \end{subfigure}
    \caption{Kendall ranking correlation between diverse calibration metrics. The metrics are evaluated across all transformer models on CIFAR-10(left) and ImageNet(right).}
    \label{fig:corr_different_metric_vit}
\end{figure}
\begin{figure}[!ht]
\begin{center}
   \includegraphics[width=\linewidth]{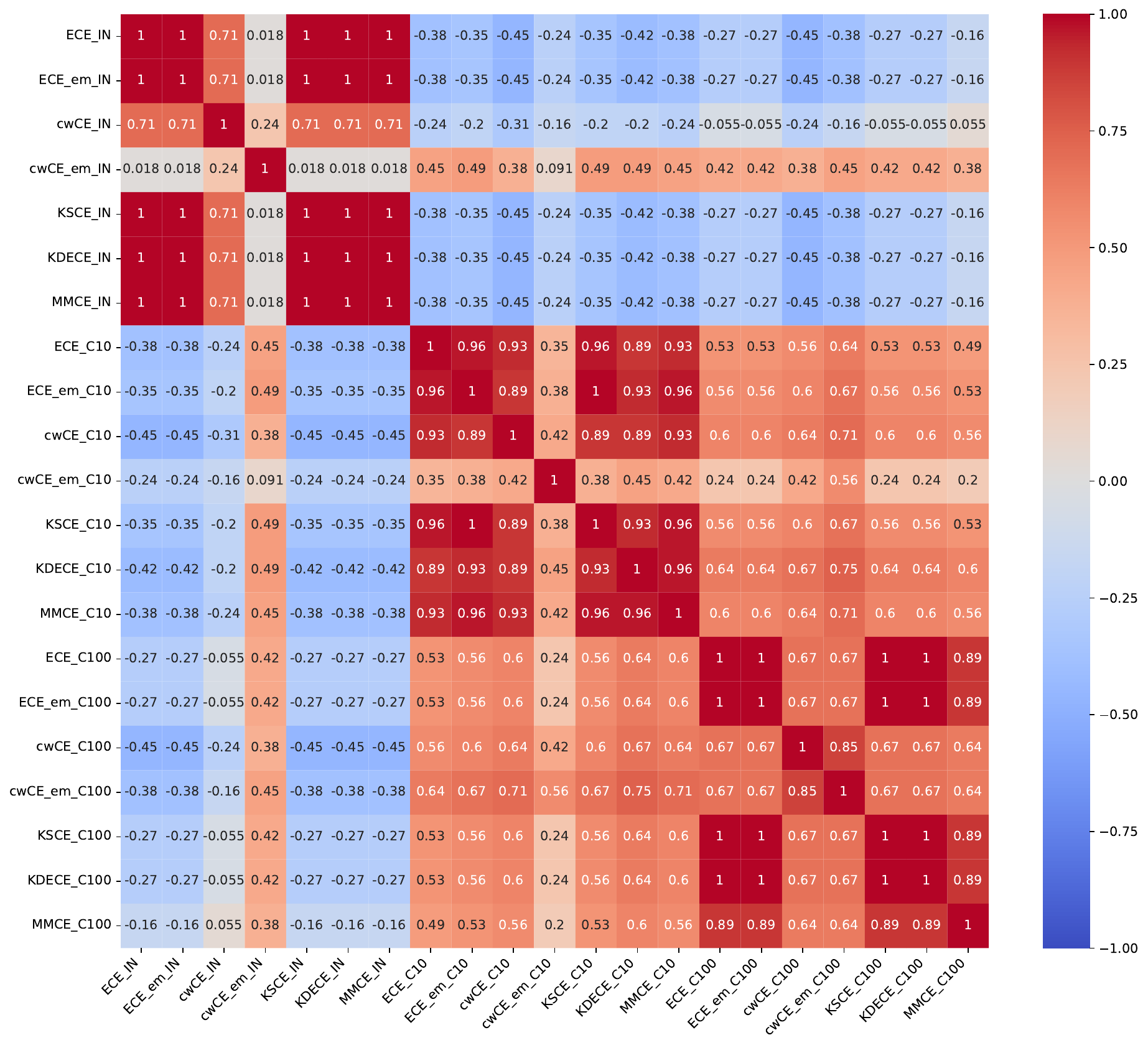}
\end{center}
    \caption{Kendall ranking correlation between all calibration metrics. The metrics are evaluated across all transformer models on CIFAR-10, CIFAR-100 and ImageNet before temperature scaling}
\label{fig:vit_all_pre}
\end{figure}
\begin{figure}[!ht]
\begin{center}
   \includegraphics[width=\linewidth]{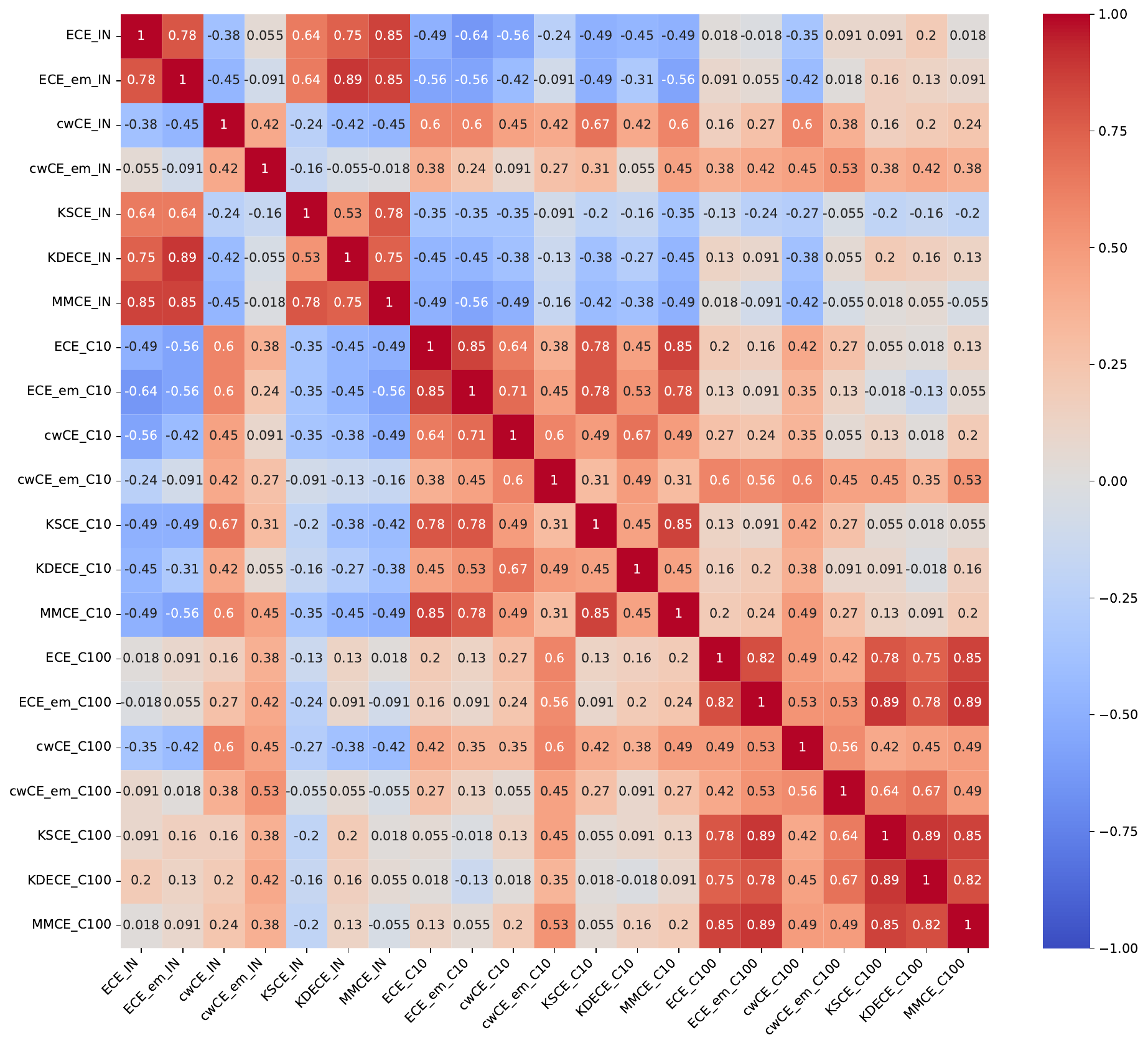}
\end{center}
    \caption{Kendall ranking correlation between all calibration metrics. The metrics are evaluated across all transformer models on CIFAR-10, CIFAR-100 and ImageNet after temperature scaling}
\label{fig:vit_all_post}
\end{figure}

Figure~\ref{fig:corr_different_metric_vit} shows the metric correlation among transformer architectures, which shows similar pattern as that on CNNs. The equal mass classwise-CE shows inconsistent results as other metrics. A full correlation map of all metrics before and after temperature scaling are shown in Figure~\ref{fig:vit_all_pre} and~\ref{fig:vit_all_post}.

\clearpage

\section{Experiments on Other Calibration Methods}
\label{sec:Experiments on Other Calibration Methods}
To ensure the broad applicability of our findings, we conducted evaluations using 6 human-designed Convolutional Neural Networks (CNNs) with diverse architectures, including ResNet18, ResNet34, ResNet50, ResNet110~\citep{he2016deep}, Wide-ResNet~\citep{zagoruyko2016wide}, and DenseNet121~\citep{huang2017densely}. Each CNN underwent 200 epochs of training on CIFAR-10 and CIFAR-100 datasets, employing different loss functions, namely Cross Entropy, Focal Loss~\citep{mukhoti2020calibrating}, and MMCE Loss~\citep{kumar2018trainable}. The selected loss functions, Focal Loss and MMCE Loss, are established train-time calibration methods. Our primary objective is to ascertain whether the patterns observed in our study are consistent across various calibration techniques, extending beyond the only calibration approach used in this work, temperature scaling~\citep{guo2017calibration}.
\begin{figure}[!ht]
\begin{center}
   \includegraphics[width=0.4\linewidth]{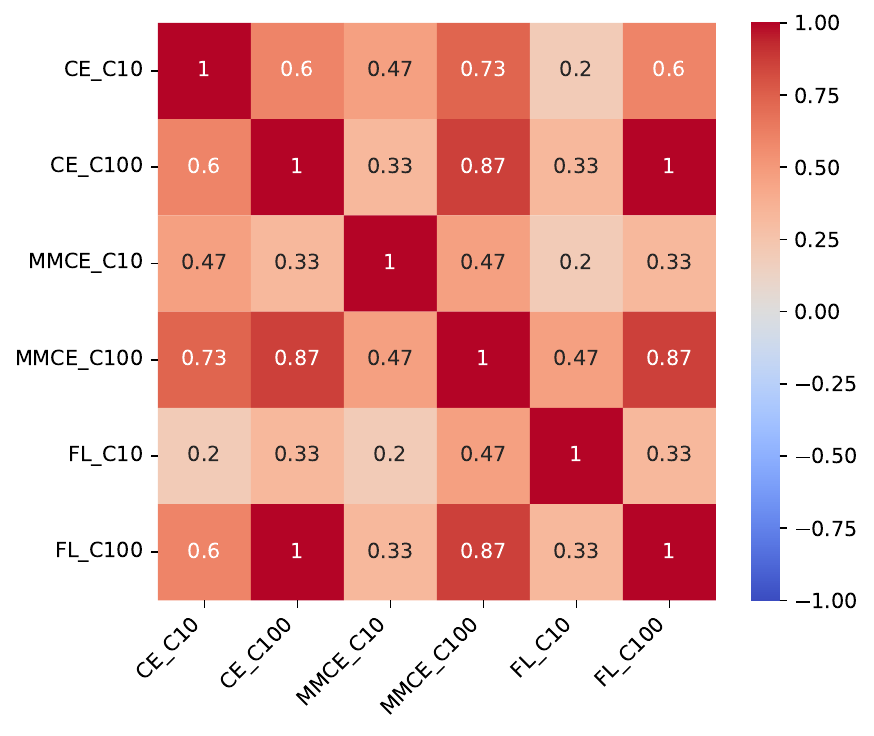}
\end{center}
    \caption{Kendall ranking correlation matrix of ECE for different CNNs.}
\label{fig:cnn_ece15_all_lossess}
\end{figure}

As depicted in Figure~\ref{fig:cnn_ece15_all_lossess}, analogous to the observation in Figure~\ref{fig:1000_all_kendall_all_datasets}, models trained using various train-time calibration methods exhibit minimal correlation between their performance on CIFAR-10 and CIFAR-100. This suggests that the calibration characteristics of a specific architecture may not generalize effectively across different datasets.

\begin{figure}[!ht]
\vspace{-0.1in}
    \centering
    \begin{subfigure}{0.32\textwidth}
        \centering
        \includegraphics[width=\textwidth]{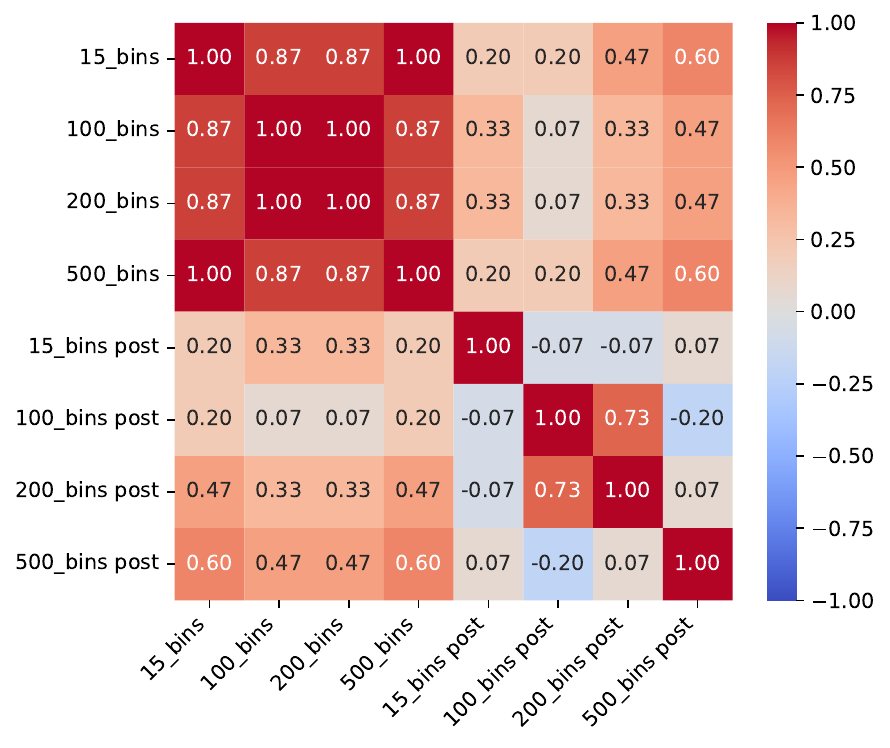}
        \label{fig:c10_ce}
    \end{subfigure}
    \hfill    
    \begin{subfigure}{0.32\textwidth}
        \centering
        \includegraphics[width=\textwidth]{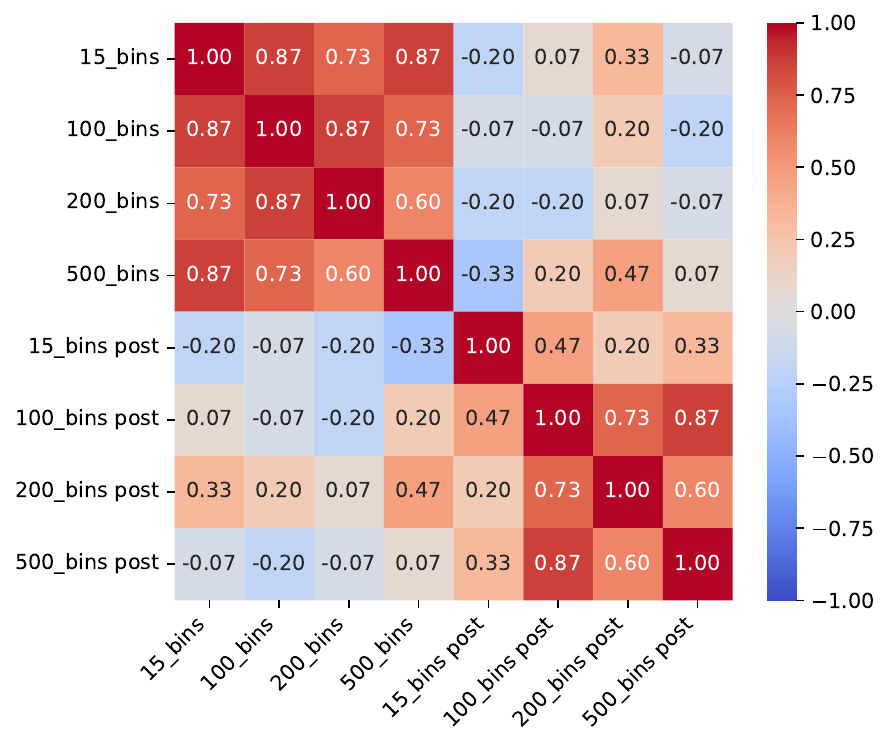}
        \label{fig:c10_fl}
    \end{subfigure}
    \hfill
    \begin{subfigure}{0.32\textwidth}
        \centering
        \includegraphics[width=\textwidth]{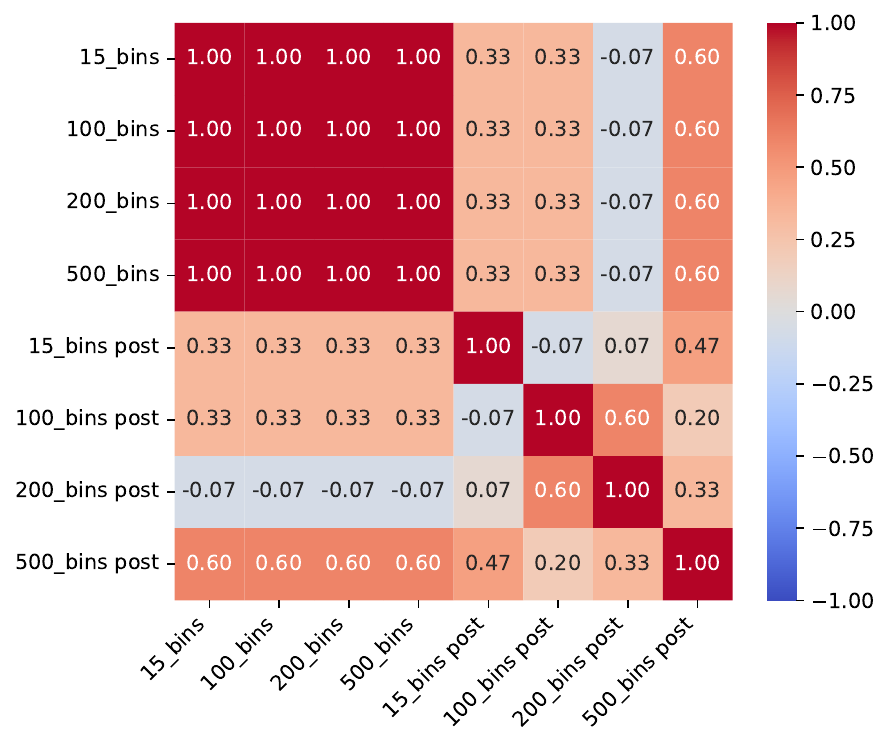}
        \label{fig:c10_mmce}
    \end{subfigure}
    \hfill
    \begin{subfigure}{0.32\textwidth}
        \centering
        \includegraphics[width=\textwidth]{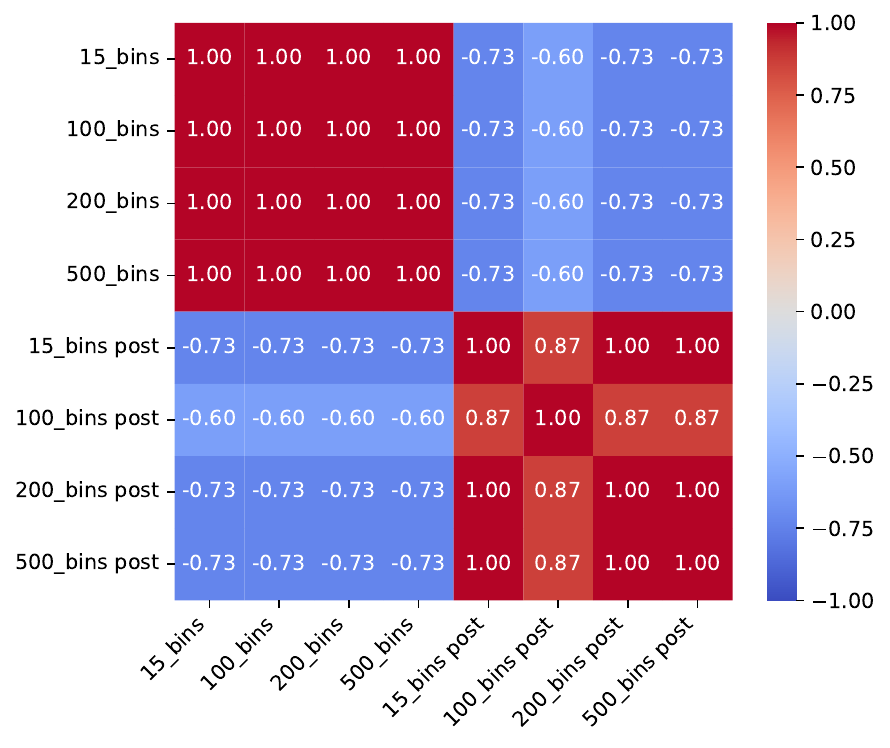}
        \label{fig:c100_ce}
    \end{subfigure}
    \hfill    
    \begin{subfigure}{0.32\textwidth}
        \centering
        \includegraphics[width=\textwidth]{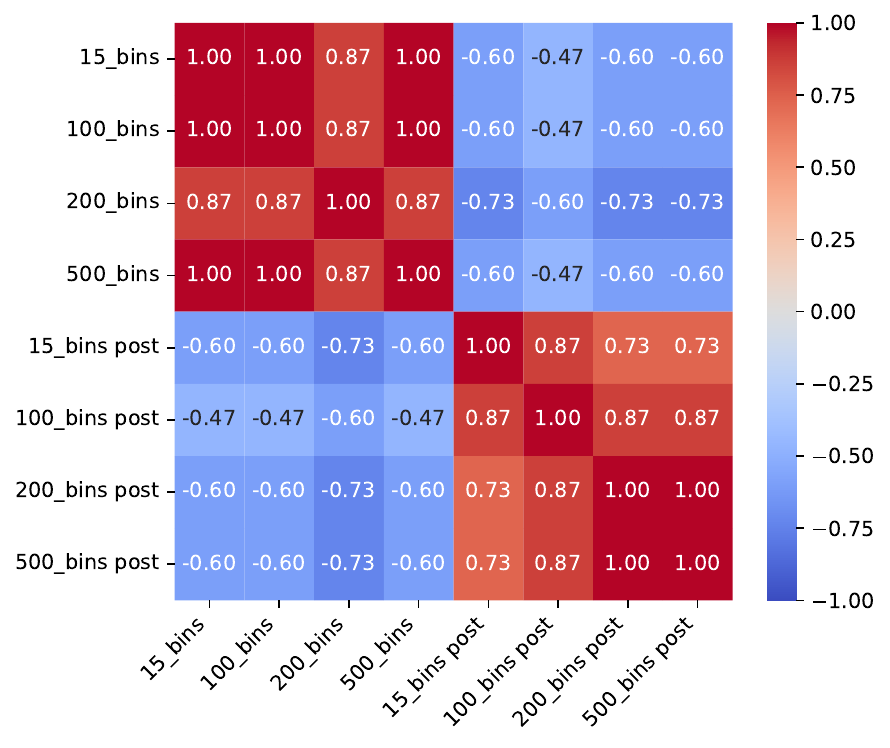}
        \label{fig:c100_fl}
    \end{subfigure}
    \hfill
    \begin{subfigure}{0.32\textwidth}
        \centering
        \includegraphics[width=\textwidth]{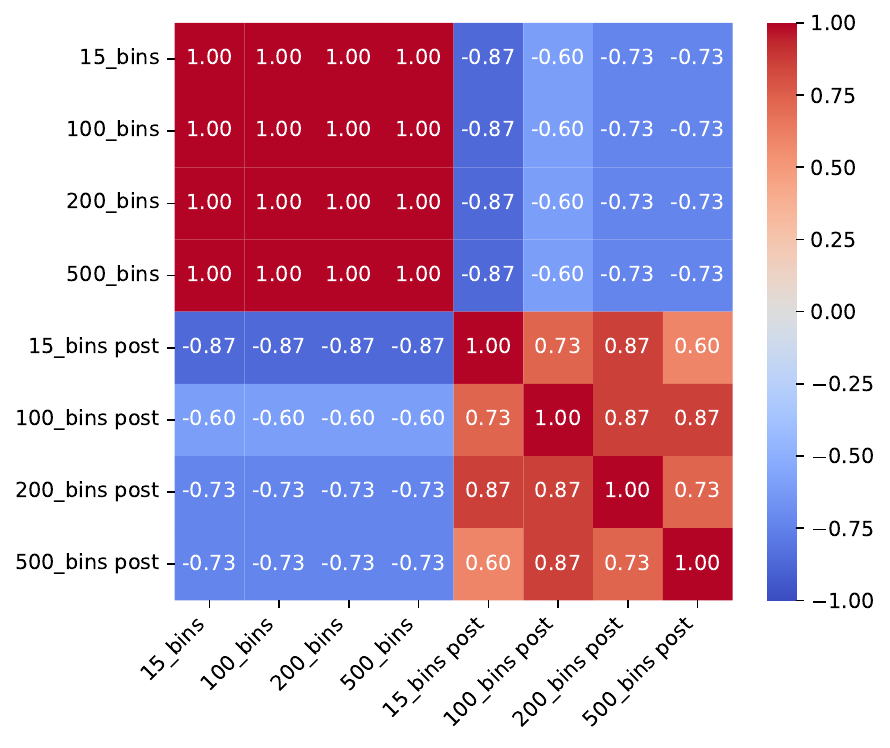}
        \label{fig:c100_mmce}
    \end{subfigure}
    \hfill
    \caption{Kendall Ranking Correlation Matrix of ECE before and after temperature scaling on Cross Entropy (left), Focal Loss (middle) and MMCE loss (right). The first row shows results on CIFAR10 and the second row shows results on CIFAR100.}
    \label{fig:corr_beforeandafter_Temp_ce_fl_mmce}
\end{figure}

As illustrated in Figure~\ref{fig:corr_beforeandafter_Temp_ce_fl_mmce}, the ECE across different bin sizes demonstrates minimal correlation between pre and post temperature scaling. This suggests that well-calibrated models do not necessarily exhibit enhanced calibration performance following post-hoc calibration techniques. This trend is particularly pronounced in the case of CIFAR-100, where post-hoc calibration performance is negatively correlated with pre-calibration performance. Notably, the choice of bin size appears to have a more substantial impact on post-hoc calibration performance.

\begin{figure}[!ht]
\begin{center}
   \includegraphics[width=\linewidth]{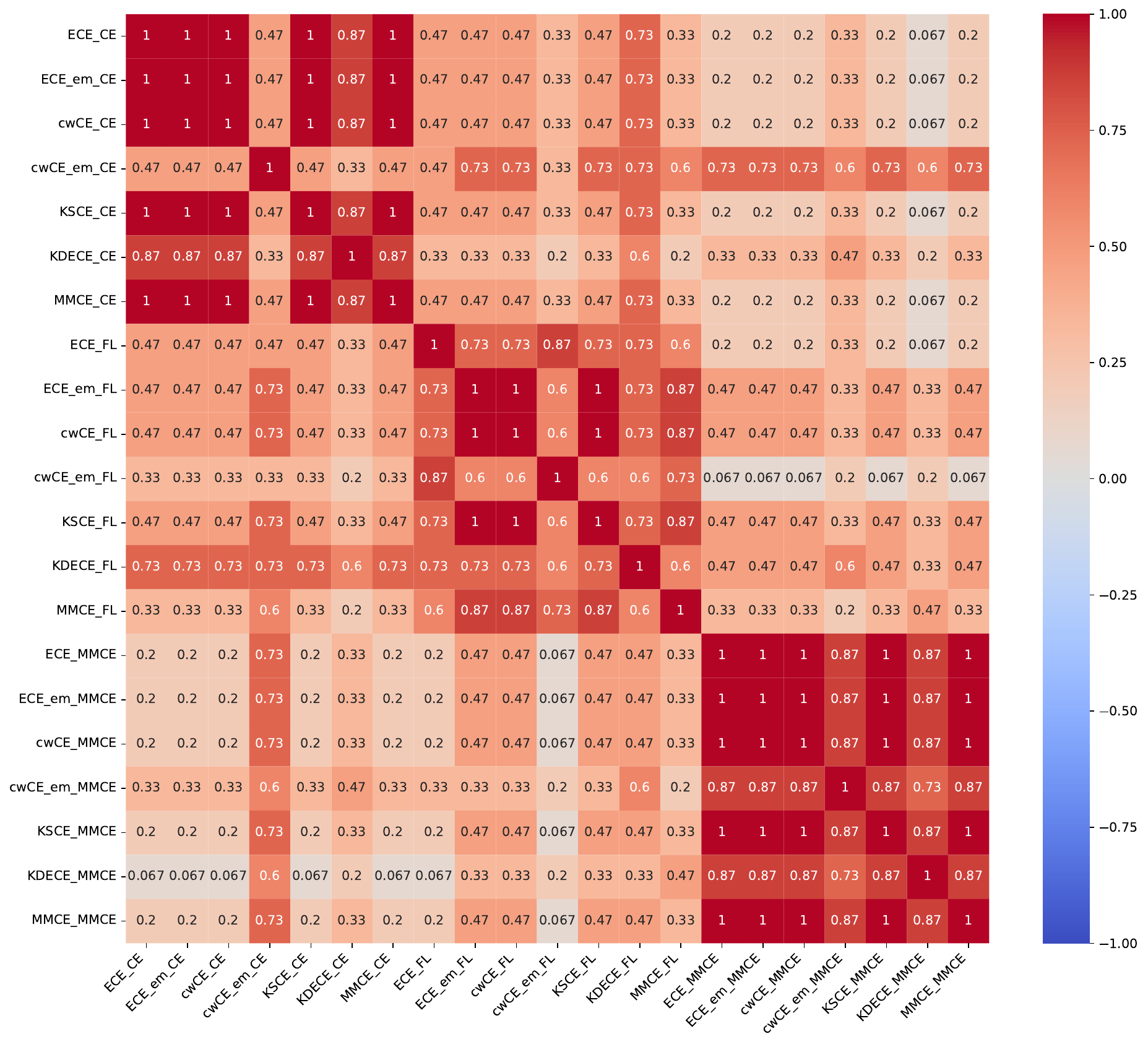}
\end{center}
    \caption{Kendall ranking correlation between all calibration metrics on CIFAR10 for different train time calibration methods.}
\label{fig:c10_ce_fl_mmce}
\end{figure}

\begin{figure}[!ht]
\begin{center}
   \includegraphics[width=\linewidth]{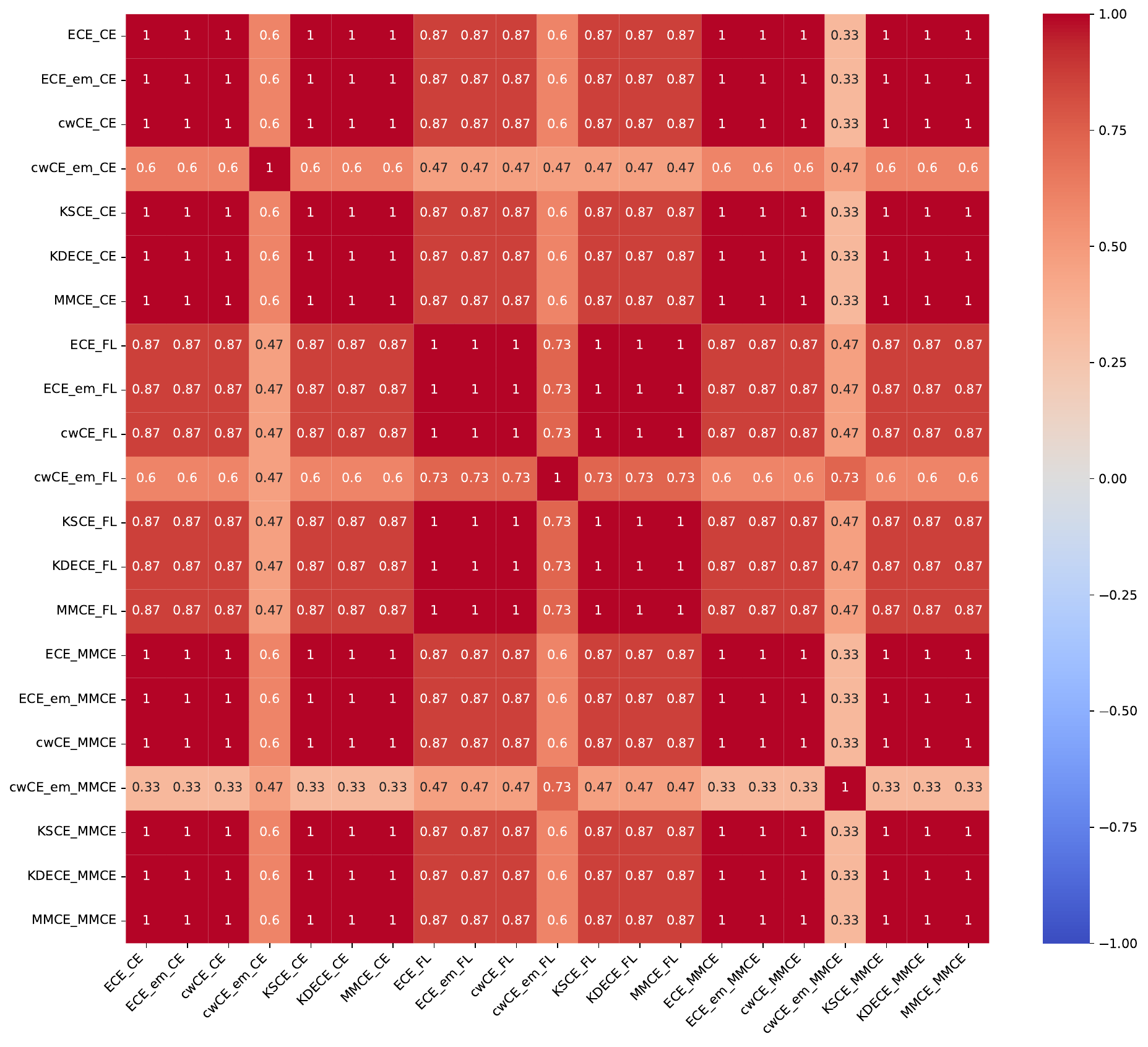}
\end{center}
    \caption{Kendall ranking correlation between all calibration metrics on CIFAR100 for different train time calibration methods.}
\label{fig:c100_ce_fl_mmce}
\end{figure}

To assess the reliability of calibration metrics, we conducted an analysis of the correlation between all calibration metrics, as presented in Figure~\ref{fig:c10_ce_fl_mmce} and Figure~\ref{fig:c100_ce_fl_mmce}. Notably, equal-mass classwise ECE displays a distinct pattern compared to other metrics, particularly on CIFAR-100, reinforcing the observations outlined in Section~\ref{metric_reliability}.

\clearpage
\section{Experiments on Large Datasets Pretrained Models}
\label{sec:Experiments on Large Datasets}
To substantiate our research using modern larger datasets, we assess the performance of seven LAION-2b~\citep{schuhmann2022laion} pretrained models on ImageNet-1K~\citep{deng2009imagenet}. Additionally, we exam the zero-shot calibration capabilities of large models by evaluating seven pretrained CLIP models on diverse large datasets, including YFCC100M~\citep{thomee2016yfcc100m} and LAION-400m~\citep{schuhmann2021laion}, evaluated on CIFAR10 and CIFAR100, respectively. 

All pretrained models are downloaded from HuggingFace, seven LAION-5b pretrained models are:
\begin{enumerate}
    \item vit\_large\_patch14\_clip\_224.laion2b\_ft\_in1k
    \item vit\_base\_patch32\_clip\_224.laion2b\_ft\_in1k
    \item vit\_huge\_patch14\_clip\_224.laion2b\_ft\_in1k
    \item vit\_base\_patch16\_clip\_224.laion2b\_ft\_in1k
    \item convnext\_large\_mlp.clip\_laion2b\_augreg\_ft\_in1k
    \item convnext\_base.clip\_laion2b\_augreg\_ft\_in1k
    \item vit\_base\_patch16\_clip\_224.laion2b\_ft\_in12k\_in1k
\end{enumerate}
Seven zero-shot models are:
\begin{enumerate}
    \item openai/clip-vit-base-patch16
    \item openai/clip-vit-large-patch14
    \item openai/clip-vit-base-patch32
    \item laion/CLIP-ViT-H-14-laion2B-s32B-b79K
    \item patrickjohncyh/fashion-clip
    \item flax-community/clip-rsicd-v2
    \item flax-community/clip-rsicd
\end{enumerate}

\subsection{Pretrained Models on ImageNet}
\begin{figure}[!ht]
\vspace{-0.1in}
    \centering
    \begin{subfigure}{0.32\textwidth}
        \centering
        \includegraphics[width=\textwidth]{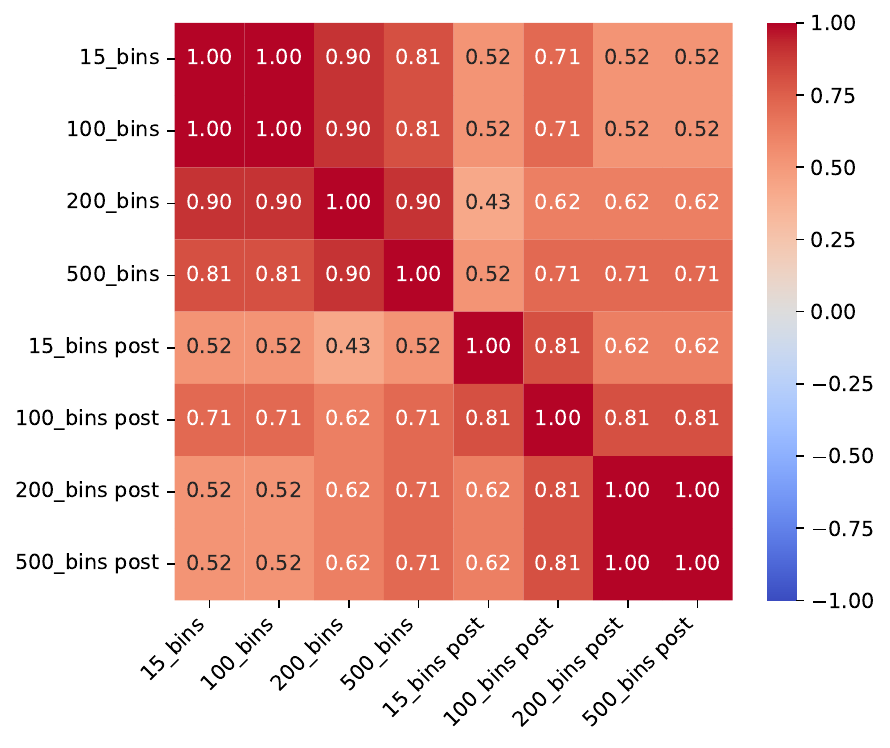}
        \label{fig:pretrained_temp_cifar10}
    \end{subfigure}
    \hfill    
    \begin{subfigure}{0.32\textwidth}
        \centering
        \includegraphics[width=\textwidth]{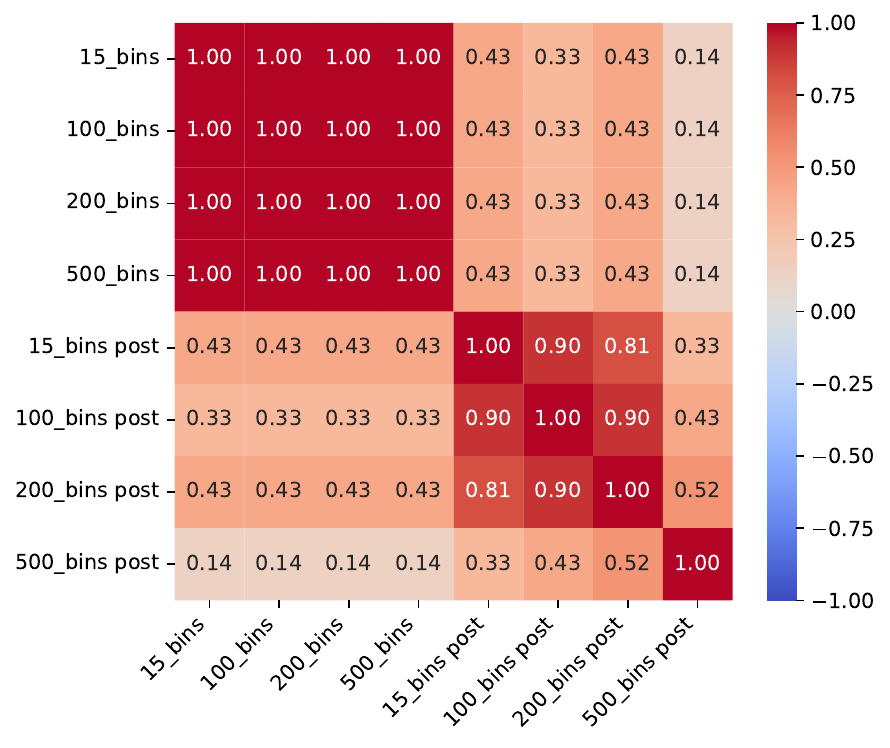}
        \label{fig:pretrained_temp_cifar100}
    \end{subfigure}
    \hfill
    \begin{subfigure}{0.32\textwidth}
        \centering
        \includegraphics[width=\textwidth]{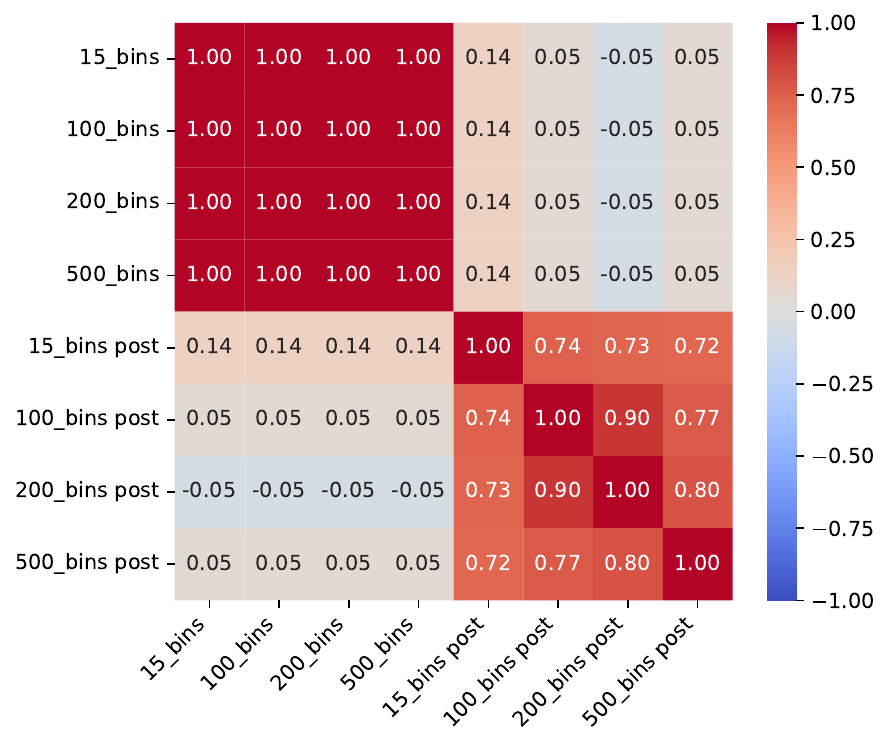}
        \label{fig:pretrained_temp_IN}
    \end{subfigure}
    \hfill
    \caption{Kendall Ranking Correlation Matrix of ECE before and after temperature scaling with different bin size on CIFAR-10 (left), CIFAR-100 (middle) and ImageNet (right).}
    \label{fig:pretrained_corr_beforeandafter_Temp}
\end{figure}

The right graph of Fig.~\ref{fig:pretrained_corr_beforeandafter_Temp} reveals a pattern consistent with Section 4.4, highlighting that the choice of bin size significantly influences post temperature scaling, while there is minimal correlation between pre temperature scaling ECE and post temperature scaling ECE. 

\begin{figure}[!ht]
\vspace{-0.1in}
    \centering
    \begin{subfigure}{0.32\textwidth}
        \centering
        \includegraphics[width=\textwidth]{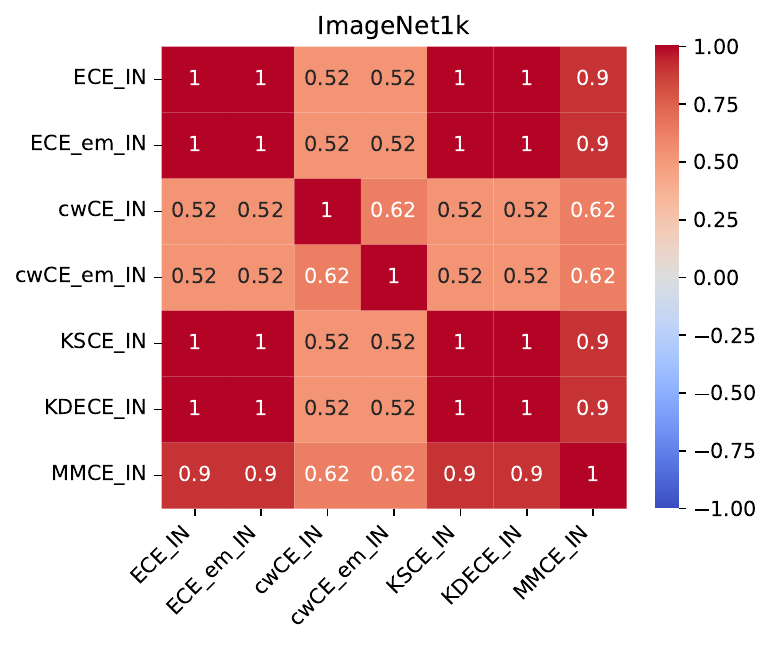}
        \label{fig:pretrained_imagenet_all_metrics_pre}
    \end{subfigure}
    \begin{subfigure}{0.32\textwidth}
        \centering
        \includegraphics[width=\textwidth]{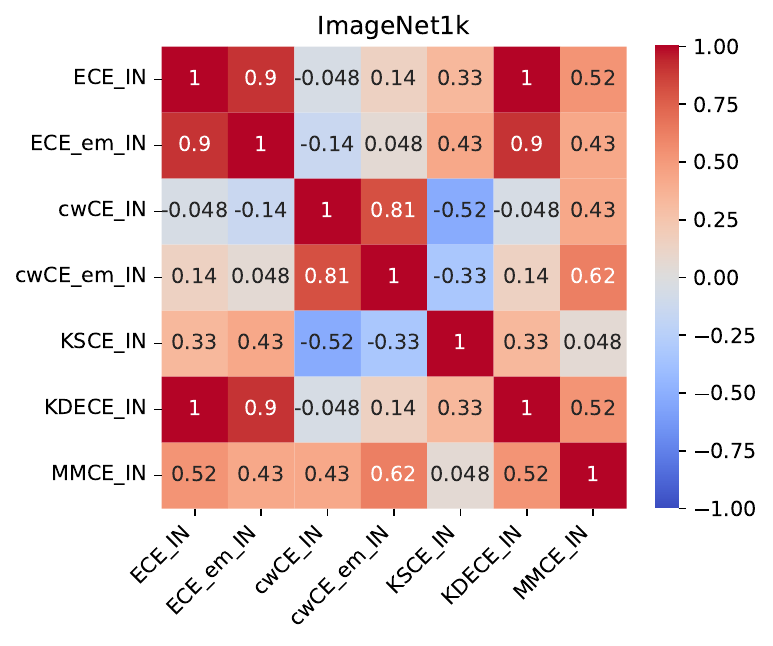}
        \label{fig:pretrained_imagenet_all_metrics_post}
    \end{subfigure}
    \caption{Kendall ranking correlation between diverse calibration metrics on ImageNet before and after temperature scaling.}
    \label{fig:pretrained_imagenet_all_metrics_pre_post}
\end{figure}

Similarly, in Fig.~\ref{fig:pretrained_imagenet_all_metrics_pre_post}, post temperature scaling calibration metrics exhibit lower correlations. Notably, equal mass classwise ECE demonstrate relatively weaker correlations with others, emphasizing the unreliability of this metric. These observations align with the findings presented in the main paper.

\subsection{Pretrained Models on Zero Shot Calibration}
From Fig.~\ref{fig:pretrained_corr_beforeandafter_Temp}, a pattern akin to that observed on ImageNet emerges, revealing that bin size exerts a more significant influence on post hoc calibration. Notably, well-calibrated models do not consistently manifest improved calibration performance after post-hoc calibration techniques, as elucidated in section 4.4.

\begin{figure}[!ht]
\begin{center}
   \includegraphics[width=0.6\linewidth]{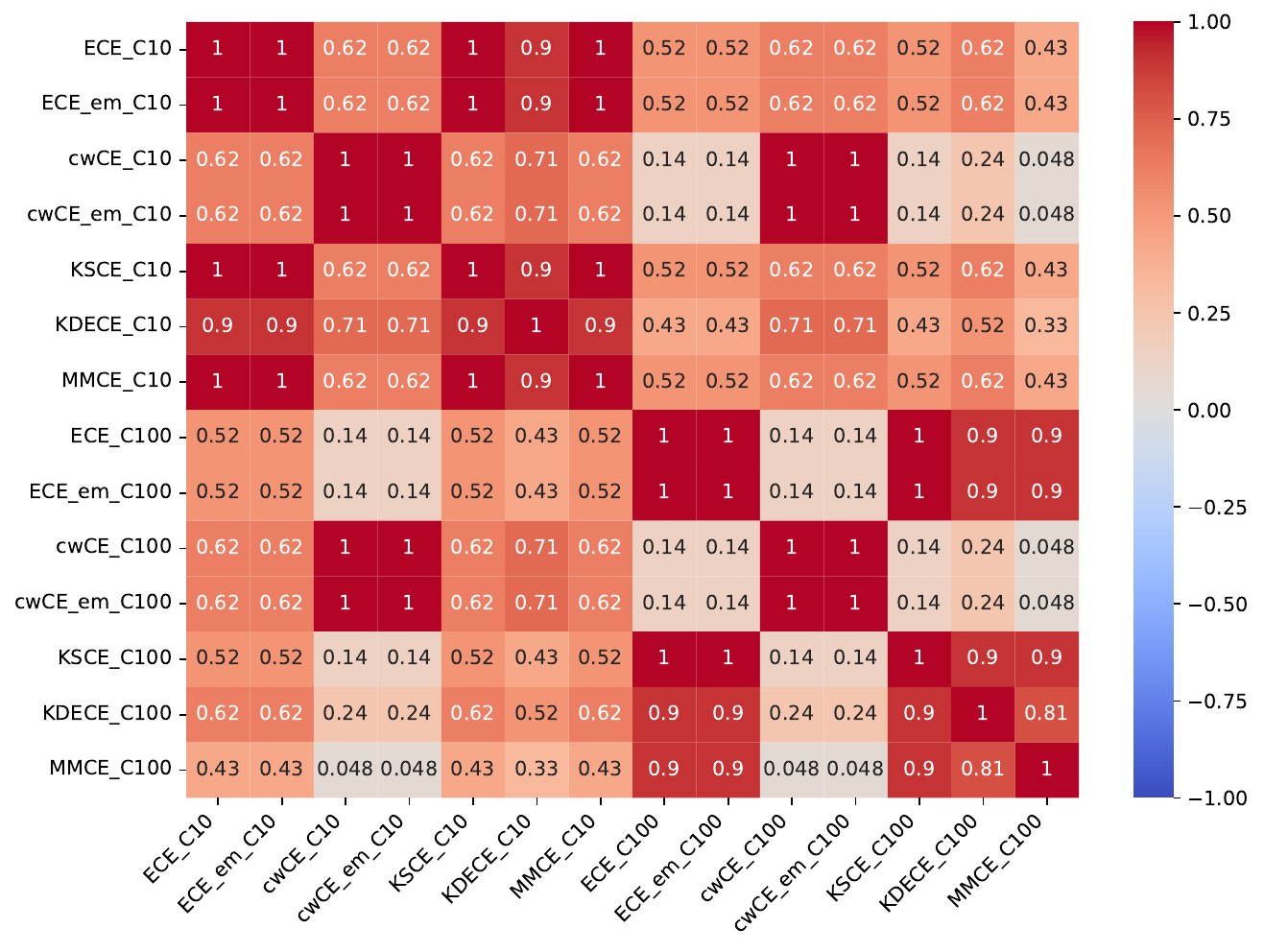}
\end{center}
    \caption{Kendall ranking correlation between all calibration metrics on CIFAR10 and CIFAR100.}
\label{fig:pretrained_c10_c100_all_metrics_pre}
\end{figure}

\begin{figure}[!ht]
\begin{center}
   \includegraphics[width=0.6\linewidth]{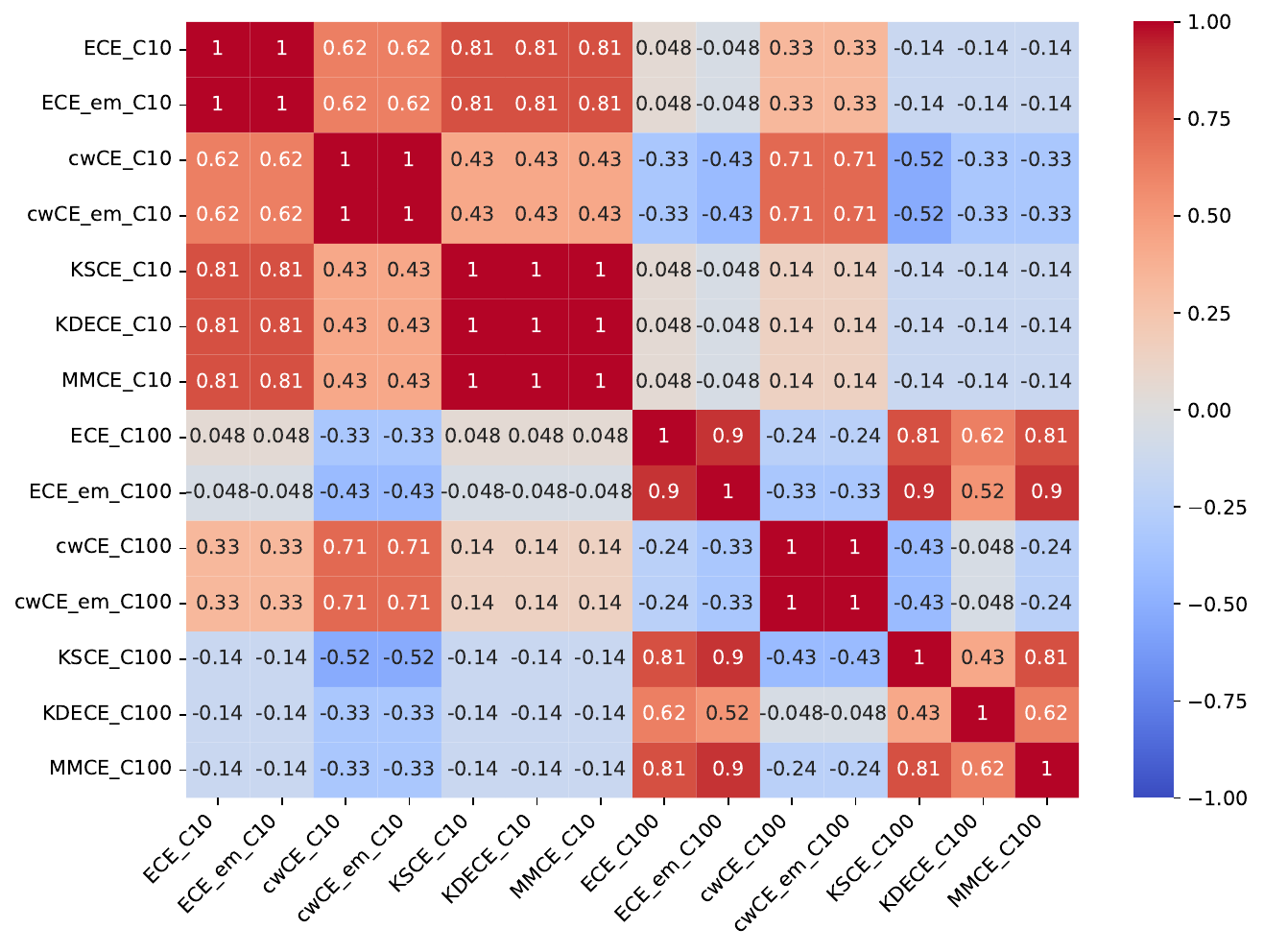}
\end{center}
    \caption{Kendall ranking correlation between all calibration metrics on CIFAR10 and CIFAR100 after temperature scaling.}
\label{fig:pretrained_c10_c100_all_metrics_post}
\end{figure}

Fig.\ref{fig:pretrained_c10_c100_all_metrics_pre} and Fig.\ref{fig:pretrained_c10_c100_all_metrics_post} further underscore the limited generalizability of calibration properties across diverse datasets, a point thoroughly discussed in section 4.1. This phenomenon is particularly evident in the context of post-temperature scaling calibration error measurements.

\clearpage
\section{Datasets}
\label{sec:Datasets}
\subsection{CIFAR-10}
This dataset serves as a standard benchmark for image classification tasks, encompassing 60,000 colored images, each of 32x32 pixels, distributed across 10 distinct classes. The original training set houses 50,000 images, allocating 5,000 images for each class, while the test set contains 10,000 images with 1,000 images per class. To facilitate validation, the test images in CIFAR-10 are split into two groups, with 20\% allocated for validation purposes.

\subsection{CIFAR-100}
Mirroring the structure of CIFAR-10, the CIFAR-100 dataset extends the classification to 100 fine-grained classes while retaining the same image collection. The original training and test sets remain the same with 50,000 and 10,000 images respectively. A random split is performed on the original test set, dividing it into two groups with an 80/20 ratio. The latter group serves as the validation set, while the former constitutes the new test set.

\subsection{ImageNet16-120}
The ImageNet-16-120 dataset is constructed from a down-sampled variant of ImageNet, dubbed ImageNet16×16, which significantly reduces computation costs for optimizing hyperparameters in certain classical models, as indicated in reference~\citep{chrabaszcz2017downsampled}. This down-sampling process shrinks the original ImageNet images to 16×16 pixels. From this variant, all images labeled between 1 and 120 are selected to form ImageNet-16-120. The resultant dataset comprises 151.7K training images, 3K validation images, and 3K test images, spread across 120 classes.

\clearpage
\section{Reproducibility}
\label{sec:Reproducibility}
Dataset is anonymously available at \url{https://anonymous.4open.science/r/CalibrationDataset5AE6}. The README.md file illustrate the use of our dataset. The detailed reproduction process is shown in Algorithm~\ref{Generation Algorithm}.
\begin{algorithm}[ht]
\caption{Calibration Metric Dataset Generation}
\begin{algorithmic}[1]
\Statex (i) Architecture space $A$ (NATS-Bench)
\Statex (ii) Test datasets $D$ (CIFAR-10, CIFAR-100, ImageNet16-120)
\Statex (iii) Set of calibration metrics $E$
\Statex (iv) Calibration Metric Dataset $C$

\For{$a \in A$}
    \State {$\triangleright$ Load pretrained weights for $a$}
    \State $a.load\_weights(d)$
    \For{$d \in D$}
        \For{$eval(\cdot, \cdot) \in E$}
            \State {$\triangleright$Evaluate architecture $a$ with $de$}
            \State Calibration errors $\leftarrow eval(a, de)$
            \State {$\triangleright$Extend calibration dataset with evaluations}
            \State $C[d][e][``CE"][a] \leftarrow$ Calibration errors
        \EndFor
    \EndFor
\EndFor
\end{algorithmic}
\label{Generation Algorithm}
\end{algorithm}

\clearpage
\section{NAS on Calibration}
\label{sec:NAS on Calibration}

By utilizing the proposed HCS, we are able to evaluate the performance of NAS algorithms in the pursuit of better calibrated models. Specifically, we have selected three representative NAS algorithms - Regularized Evolution~\citep{real2019regularized}, Local Search~\citep{white2021exploring}, and Random Search~\citep{li2020random} - and have measured their performance in searching models with better calibration performance across different metrics. As indicated in Table~\ref{table:nascalibration}, Regularized Evolution achieved better balance between prediction accuracy and calibration performance when searching through $\text{HCS}_{3}$.
\begin{table}[!ht]
	\centering
	\scriptsize
		\begin{tabular}{cccccccccccccc}
			\toprule   
			 & \textbf{Method}  &\textbf{Acc}  &  \textbf{ECE}  &  \textbf{\text{HCS}$_{1}$} &  \textbf{\text{HCS}$_{2}$} &  \textbf{\text{HCS}$_{3}$} & \textbf{MMCE}
                & \textbf{KSCE}\\
			\midrule
			\multirow{4}{*}{Accuracy} & Regularized Evolution~\citep{real2019regularized} & 93.91 &  4.20 & 94.84 & 95.16 & 95.32 & 3.71 & 4.18\\
		    & 
Local Search~\citep{white2021exploring} & 94.01 &  4.25 & 94.87 & 95.16 & 95.31 & 3.82 & 4.23\\   
			&  Random Search~\citep{li2020random} & 93.52 &  4.15 & 94.67 & 95.06 & 95.26 & 3.69 & 4.14\\
   			\midrule
			\multirow{4}{*}{$\text{HCS}_1$} & Regularized Evolution~\citep{real2019regularized} & 93.94 &  4.17 & 94.88 & 95.19 & 95.35 & 3.74 & 4.15\\	
			& Local Search~\citep{white2021exploring} & 93.98 &  4.09 & 94.94 & 95.26 & 95.42 & 3.67 & 4.06\\
			&  Random Search~\citep{li2020random} & 93.59 &  4.21 & 94.68 & 95.05 & 95.23 & 3.73 & 4.16\\
            \midrule
			\multirow{4}{*}{$\text{HCS}_2$} & Regularized Evolution~\citep{real2019regularized} & 93.80 &  4.14 & 94.82 & 95.17 & 95.34 & 3.71 & 4.11\\	
			& Local Search~\citep{white2021exploring} & 93.73 &  4.05 & 94.83 & 95.20 & 95.39 & 3.62 & 4.01\\
			&  Random Search~\citep{li2020random} & 93.06 &  4.12 & 94.45 & 94.92 & 95.16 & 3.58 & 4.08\\
      			\midrule
			\multirow{4}{*}{$\text{HCS}_3$} & Regularized Evolution~\citep{real2019regularized}& 93.62 &  3.90 & 94.84 & 95.26 & 95.47 & 3.42 & 3.86\\	
			& Local Search~\citep{white2021exploring}& 93.57 &  3.91 & 94.81 & 95.24 & 95.45 & 3.42 & 3.87\\
			& Random Search~\citep{li2020random}& 93.55 &  4.23 & 94.65 & 95.02 & 95.21 & 3.71 & 4.17\\

			\bottomrule
		\end{tabular}%
	\caption{\textbf{NAS methods search on HCS and Accuracy}}
	\label{table:nascalibration}
\end{table}

\clearpage
\section{Box-plot of ECE of different bin size in TSS}
\label{sec:Box-plot of ECE of different bin size in TSS}
In this section, we provide the box plots on ECE with different bin sizes for all models in TSS at different epochs, both before and after temperature scaling, on each dataset.


\begin{figure}[!ht]
\begin{center}
   \includegraphics[width=0.9\linewidth]{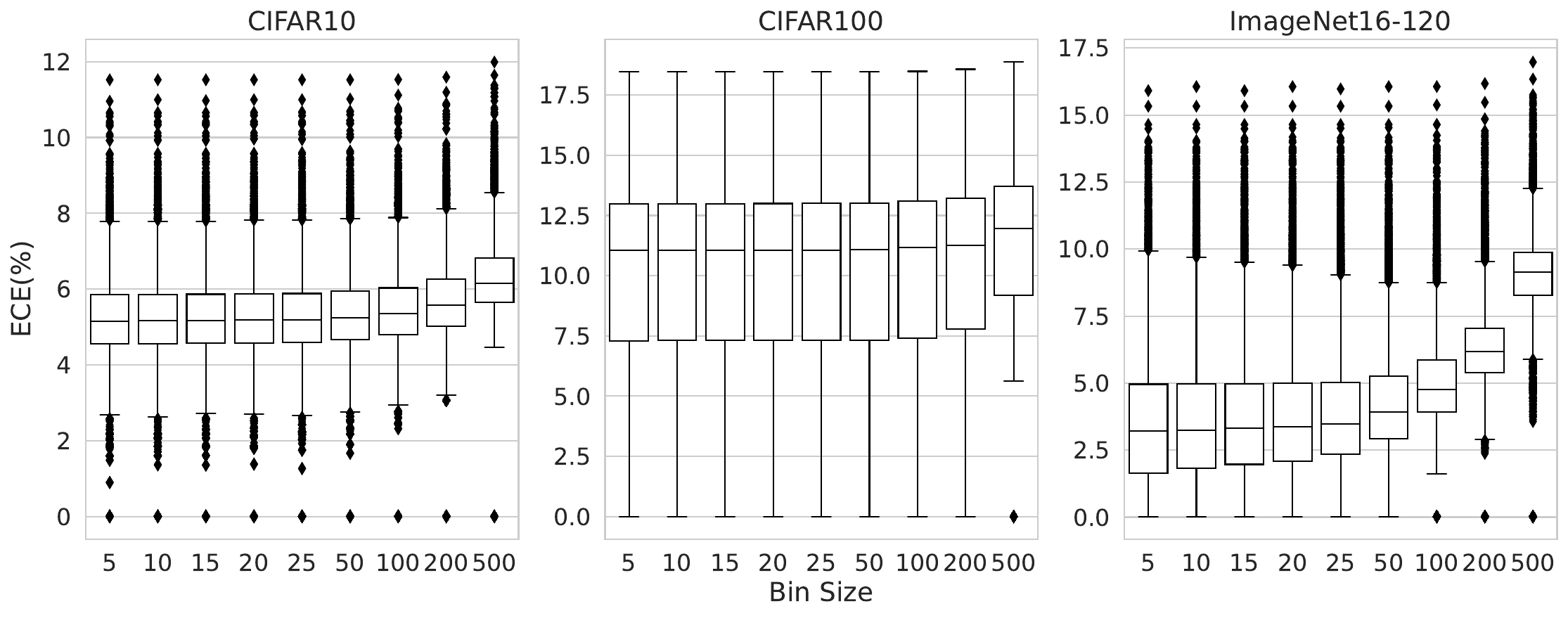}
\end{center}
    \caption{Box plots of ECE with different bin sizes are shown for models trained with 200 epochs in TSS before temperature scaling.}
\label{fig:toplevel_ew}
\end{figure}

\begin{figure}[!ht]
\begin{center}
   \includegraphics[width=0.9\linewidth]{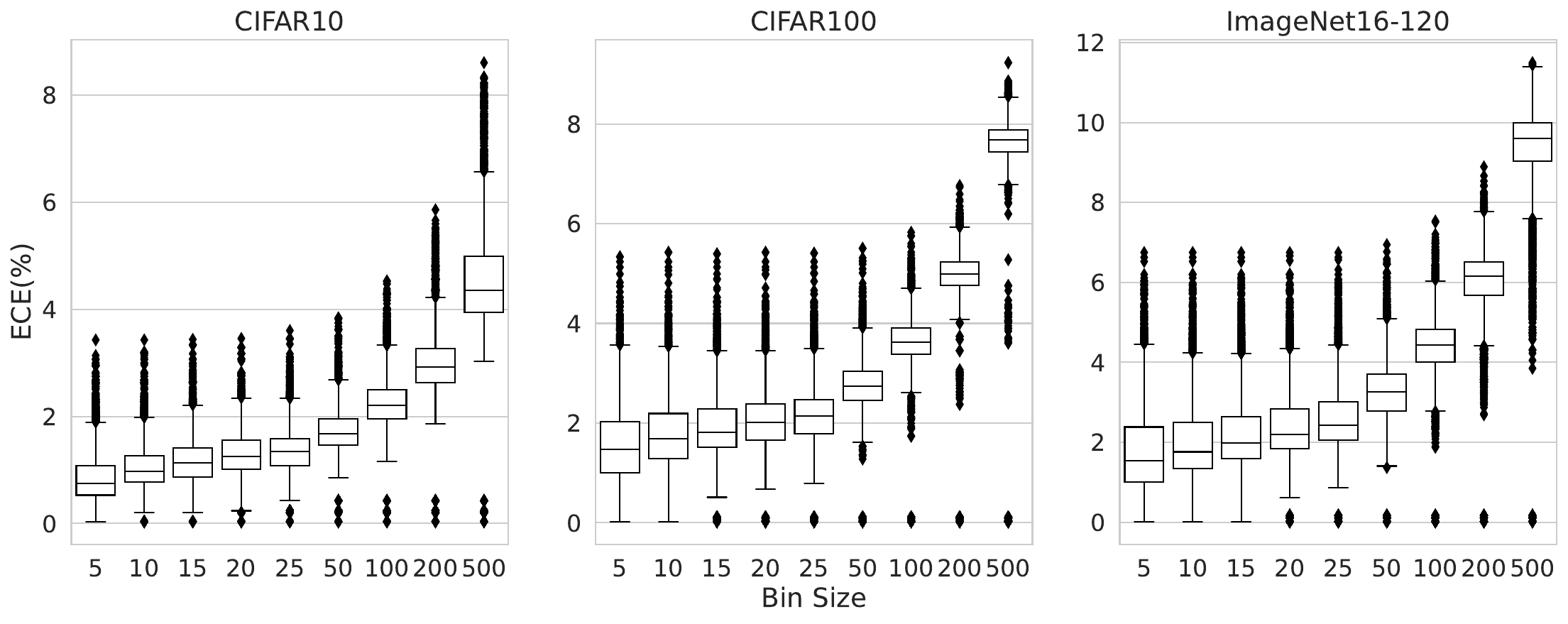}
\end{center}
    \caption{Box plots of ECE with different bin sizes are shown for models trained with 200 epochs in TSS after temperature scaling.}
\label{fig:toplevel_ew_post}
\end{figure}

\clearpage
\section{Box-plot of ECE of different bin size in SSS}
\label{sec:Box-plot of ECE of different bin size in SSS}
In this section, we provide the box plots of ECE with different bin sizes for all models in SSS at 90 epochs, both before and after temperature scaling, on each dataset.

\begin{figure}[!ht]
\begin{center}
   \includegraphics[width=\linewidth]{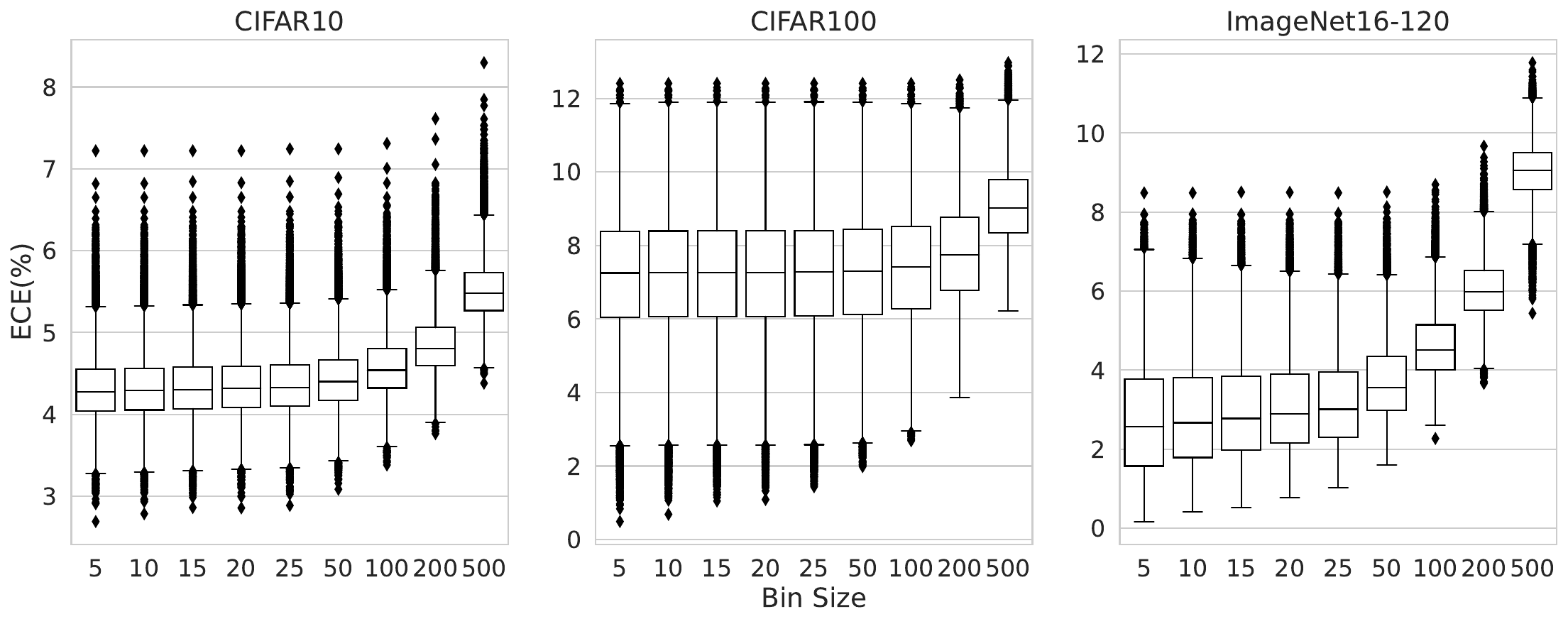}
\end{center}
    \caption{Box plots of ECE with different bin sizes are shown for models trained with 90 epochs in SSS before temperature scaling.}
\label{fig:toplevel_ew_sss}
\end{figure}

 \begin{figure}[!ht]
\begin{center}
   \includegraphics[width=\linewidth]{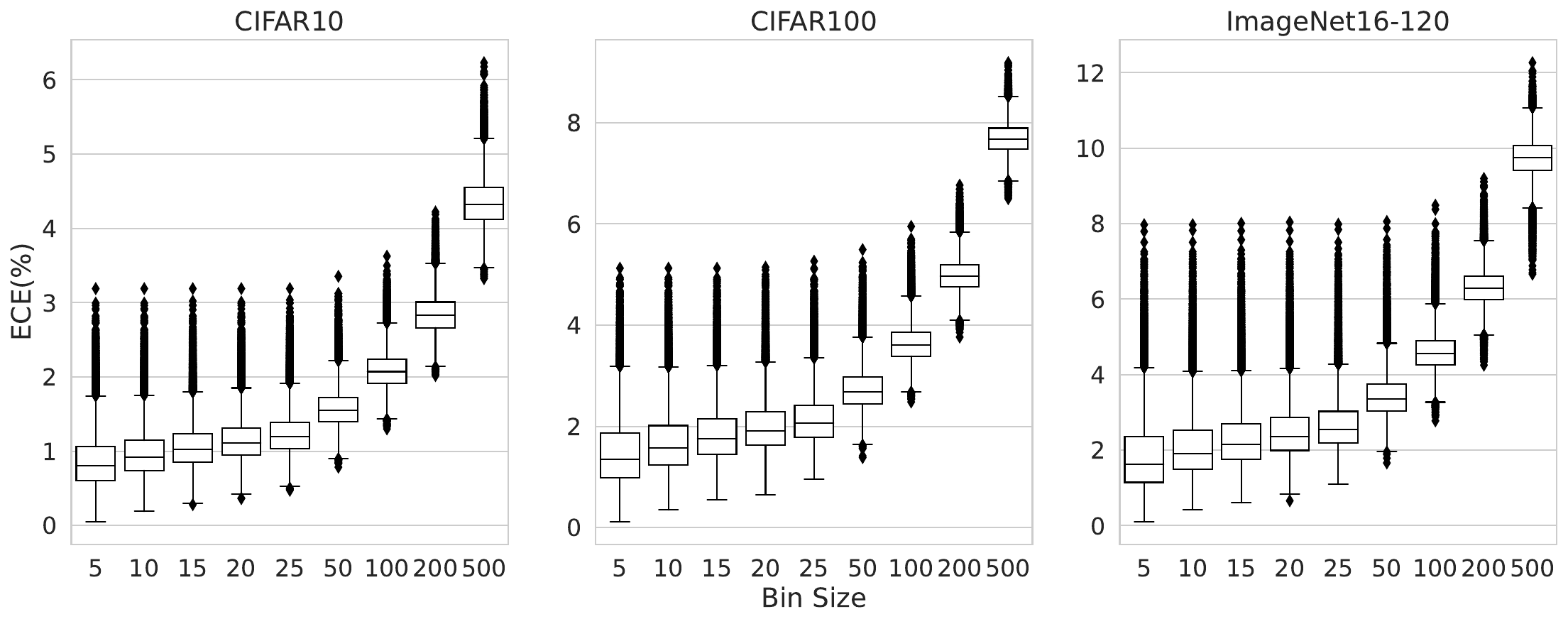}
\end{center}
    \caption{Box plots of ECE with different bin sizes are shown for models trained with 90 epochs in SSS after temperature scaling.}
\label{fig:toplevel_ew_sss_post}
\end{figure}

\clearpage
\section{Box-plot of ECE and Accuracy with Model Size}
\label{sec:Box-plot of ECE and Accuracy with Model Size}
In this section, we present the box plots of ECE and Accuracy across all architectures within the SSS subset, segregated by different size brackets. The model size is defined as the total number of kernels in each layer.
\begin{figure}[!ht]
\begin{center}
   \includegraphics[width=\linewidth]{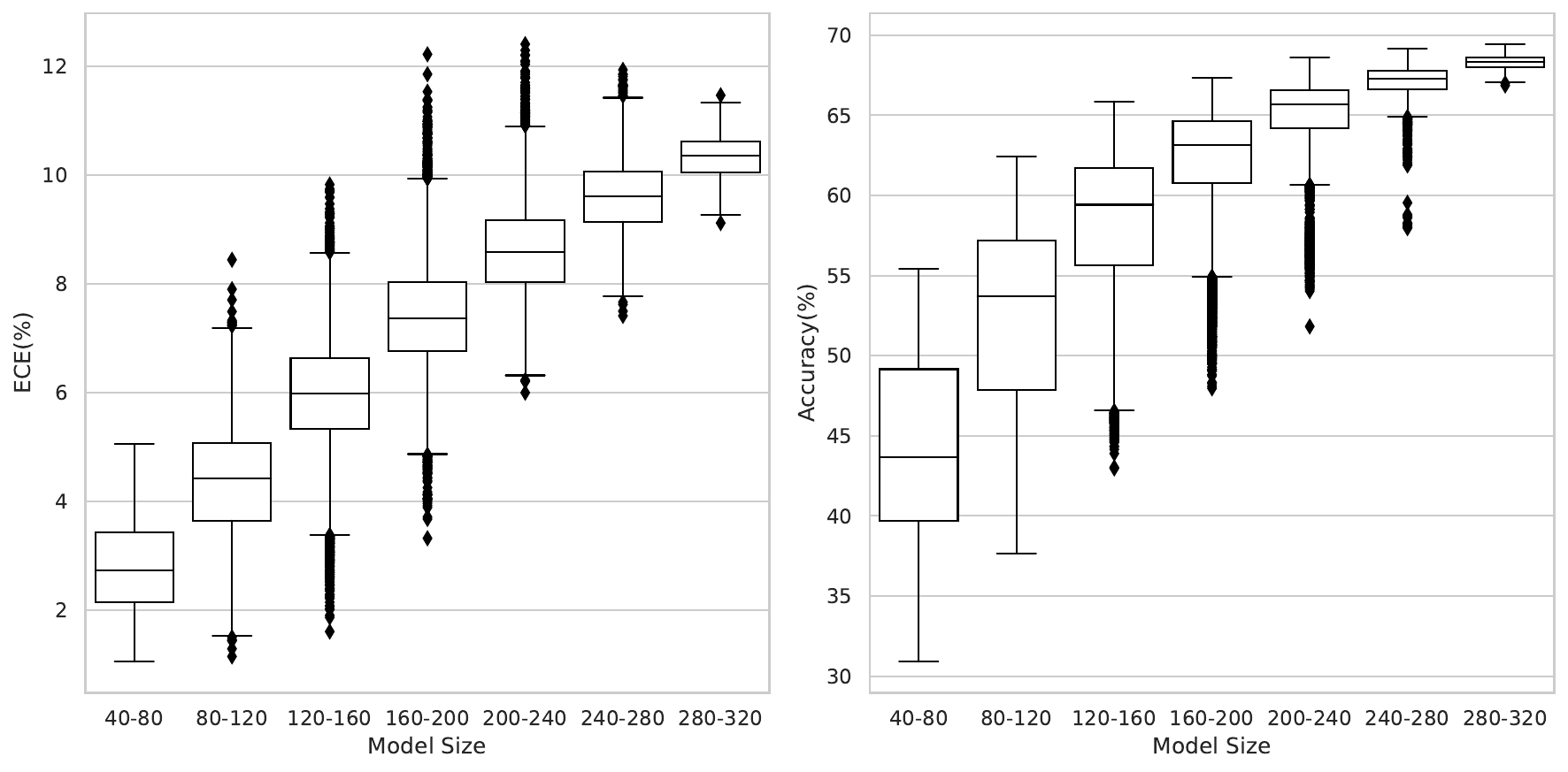}
\end{center}
    \caption{Box-plots measured on CIFAR-100 for ECE and Accuracy for all models in SSS.}
\label{fig:cifar100_sss}
\end{figure}

\begin{figure}[!ht]
\begin{center}
   \includegraphics[width=\linewidth]{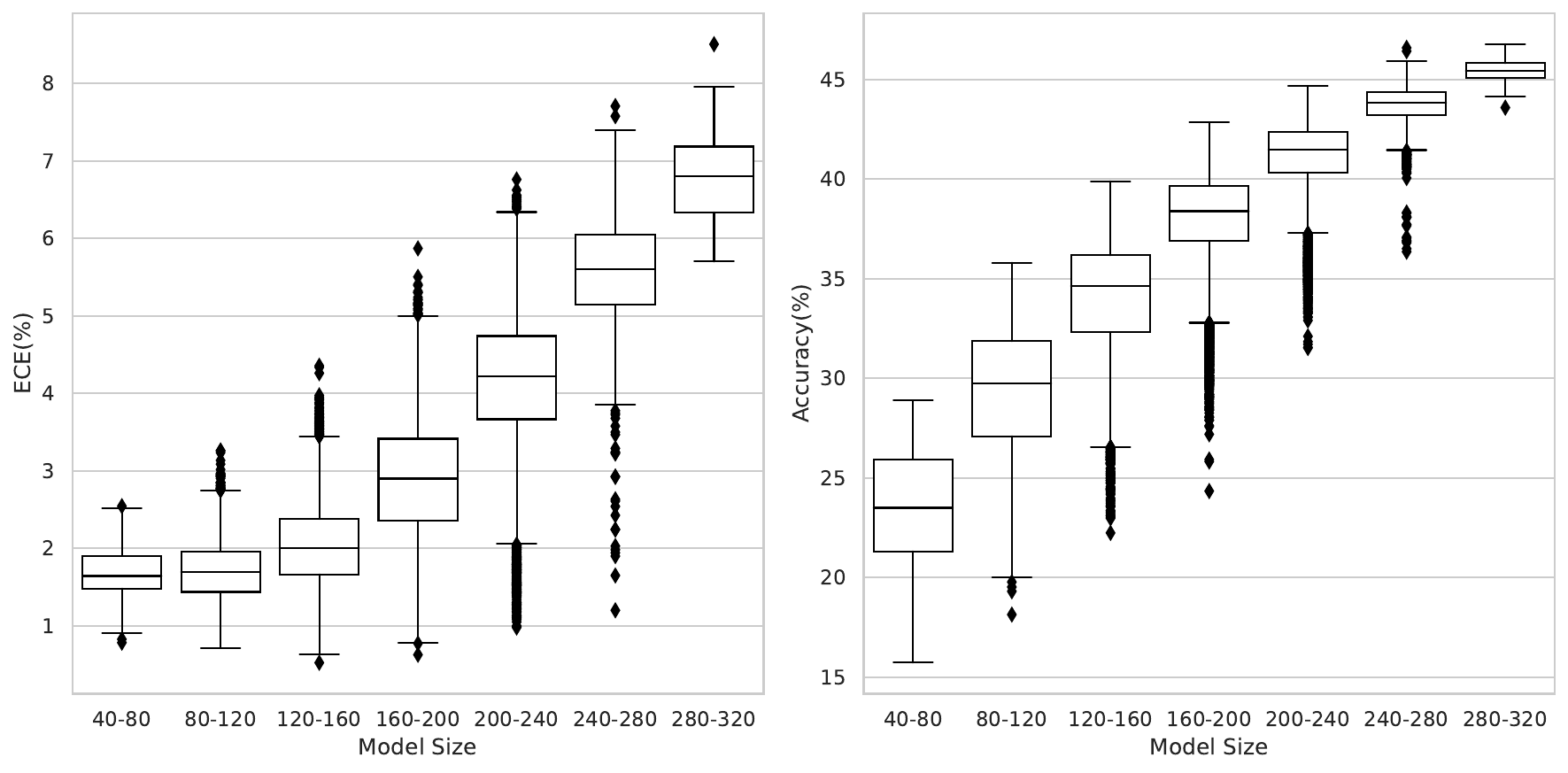}
\end{center}
    \caption{Box-plots measured on ImageNet16-120 for ECE and Accuracy for all models in SSS.}
\label{fig:imagenet_sss}
\end{figure}

\clearpage
\section{Calibration Metrics}
\label{sec:Calibration Metrics}

In our study, we consider five bin-based metrics, including Expected Calibration Error (ECE), Equal Mass Binning calibration error ($\text{ECE}{em}$), Classwise Equal-Width Binning calibration error (cwCE), Classwise Equal-Mass Binning calibration error ($\text{cwCE}{em}$), and Maximum Calibration Error (MCE). We evaluate these metrics across a wide range of bin sizes, such as 5, 10, 15, 20, 25, 50, 100, 200, and 500 bins. These metrics are assessed both before and after the application of temperature scaling. In addition to these, we examine three statistics-based metrics: Kolmogorov-Smirnov calibration error (KSCE), kernel-based Top-Label calibration error using Kernel Density Estimation (KDECE), and Maximum Mean calibration error (MMCE). We also consider two measures related to training loss: Negative Log-Likelihood (NLL) and Brier score. Furthermore, we compute the Area Under the Curve (AUC) for the True Skill Statistic (TSS) for models trained on the CIFAR-10 dataset, using two Out-of-Distribution (OoD) datasets, CIFAR-10-C (corrupted with Gaussian noise) and SVHN (Street View House Numbers). Overall, this results in the assessment of 102 different measures, providing a comprehensive evaluation of model calibration.

\noindent\textbf{Bin-based Metrics}~The Top-Label Equal-Width Binning, also known as ECE~\citep{naeini2015obtaining,guo2017calibration}, partitions the probability interval [0, 1] into equal-sized subintervals. Each data point is represented by the maximum probability output of the predicted class, dividing predictions into $m$ bins $B_1, \ldots, B_m$. Confidence and accuracy for each bin $B_i$ are calculated as $\mathrm{confidence}(B_i) = \frac{1}{|B_i|} \sum_{j \in B_i} \max f(x_j)$ and $\mathrm{accuracy}(B_i) = \frac{1}{|B_i|} \sum_{j \in B_i} \mathbb{1}{y_j \in \arg\max f(x_j)}$, with $\mathbb{1}{\cdot}$ being the Iverson bracket. The ECE is computed as $\mathrm{ECE}= \sum_{i=1}^{m} \frac{|B_i|}{n} |\mathrm{accuracy}(B_i) - \mathrm{confidence}(B_i)|$. In contrast, Equal Mass Binning calibration error ($\text{ECE}_{em}$) \citep{kumar2019verified,nixon2019measuring} uses an adaptive binning scheme, adjusting bin intervals to contain an equal number of predictions, addressing the bias-variance trade-off. Class-wise calibration errors have been proposed in \citep{kumar2019verified,nixon2019measuring,kull2019beyond}, including Classwise Equal-Width Binning calibration error (cwCE) and Classwise Equal-Mass Binning calibration error ($\text{cwCE}_{em}$). The cwCE method bins predictions per class, computes the calibration error within the bin, and averages across bins. Confidence and accuracy for each class label $k$ and bin $B_b$ are calculated as $\mathrm{confidence}(B_b, k) = \frac{1}{|B_b, k|} \sum_{j \in B_b, k} f(x_j)$ and $\mathrm{accuracy}(B_b, k) = \frac{1}{|B_b, k|} \sum_{j \in B_b, k} \mathbb{1}{y_j \in \arg\max f(x_j)}$. The cwCE is then computed as $\mathrm{cwCE}= \frac{1}{K} \sum_{k=1}^{K} \sum_{b=1}^{B} \frac{n_{b, k}}{N} |\mathrm{accuracy}(B_b, k) - \mathrm{confidence}(B_b, k)|$. $\text{cwCE}_{em}$ applies Equal-Mass binning to the multi-class setting, ensuring each bin contains an equal number of predictions for each class, offering an adaptive approach to multi-class calibration. Maximum Calibration Error (MCE) represents the worst-case error in the model's calibration and measures the maximum difference between confidence and accuracy over all bins~\citep{naeini2015obtaining}. It focuses on the most miscalibrated predictions, providing a robustness check for the calibration performance of models.

\noindent\textbf{Other Metrics}~Brier Score (BS): For a model $f: X \rightarrow P_n$, the Brier score is defined as:
\begin{equation}
BS(f) = E \left[ ||f(X) - Y'||^2_2 \right],
\end{equation}
where $Y'$ is the one-hot encoded version of $Y$. The Brier score is equivalent to the mean squared error, which suggests it captures more than just model calibration; it also measures model fit.

Lp Calibration Error (CEp): For $1 \leq p \leq 2 \in \mathbb{R}$, the Lp calibration error of model $f: X \rightarrow P_n$ is defined as:
\begin{equation}
CE_p(f) = \left( E \left[ ||f(X) - P(Y | f(X))||^p_p \right] \right)^{\frac{1}{p}}.
\end{equation}

Kernel Density Estimation (KDE) based ECE: Let $K: \mathbb{R} \rightarrow \mathbb{R}_{\geq 0}$ denote a smoothing kernel function. Given a fixed bandwidth $h > 0$, we have $K_h(a) = h^{-1}K(a/h)$. The unknown probabilities are estimated using KDE as follows:
\begin{align}
\tilde{p}(z) &= \frac{1}{h^L n_e} \sum_{i=1}^{n_e} \prod_{l=1}^{L} K_h(z_l - z^{(i)}_l), \\
\tilde{\pi}(z) &= \frac{\sum_{i=1}^{n_e} y^{(i)} \prod_{l=1}^{L} K_h(z_l - z^{(i)}_l)}{\sum_{i=1}^{n_e} \prod_{l=1}^{L} K_h(z_l - z^{(i)}_l)}, \\
ECE_{gd}(f) &= \int \left|z - \tilde{\pi}(z)\right|^d d\tilde{p}(z).
\end{align}

Kolmogorov-Smirnov calibration error (KS): This measures the maximum deviation between the predicted and actual probabilities. Given model $f: X \rightarrow P_n$, where $C = \arg \max_k f_k(X)$ and $KS(f, k)$ represents the absolute difference between the actual and predicted probabilities for class $k$, the Kolmogorov-Smirnov calibration error is defined as:
\begin{equation}
KS(f) = E \left[ KS \left(f,C\right) \right] = \sup_{z \in [0,1]} \left| \int_0^z P(Y = k | f_k(X) \leq z) dP_{f_k(X)}(z) \right|.
\end{equation}

Maximum Mean Calibration Error (MMCE): Given a reproducing kernel Hilbert space $H$ with kernel $k: [0, 1] \times [0, 1] \rightarrow \mathbb{R}$, the maximum mean calibration error of model $f: X \rightarrow P_n$ is defined as:
\begin{equation}
\text{MMCE}(f) = \sup_{h \in H, ||h|| \leq 1} \left| E \left[ \left(f_C(X) - P(Y = C | f_C(X))\right) k \left(f_C(X), .\right) \right] \right|.
\end{equation}

\newpage
\section{Calibration Kendall Ranking Confusion Matrix for TSS}
\label{sec:Calibration Kendall Ranking Confusion Matrix for TSS}
In Figure~\ref{fig:1000_all_kendall_all_datasets}, we plot a Kendall correlation matrix of ECE for different datasets. In Figure~\ref{fig:metric_correlation}, we plot a Kendall correlation matrix for different metrics. In this section, we provide comprehensive correlation matrix between calibration metrics on different datasets on TSS, filtered by different top accuracy populations.

\begin{figure}[!ht]
\begin{center}
   \includegraphics[width=\linewidth]{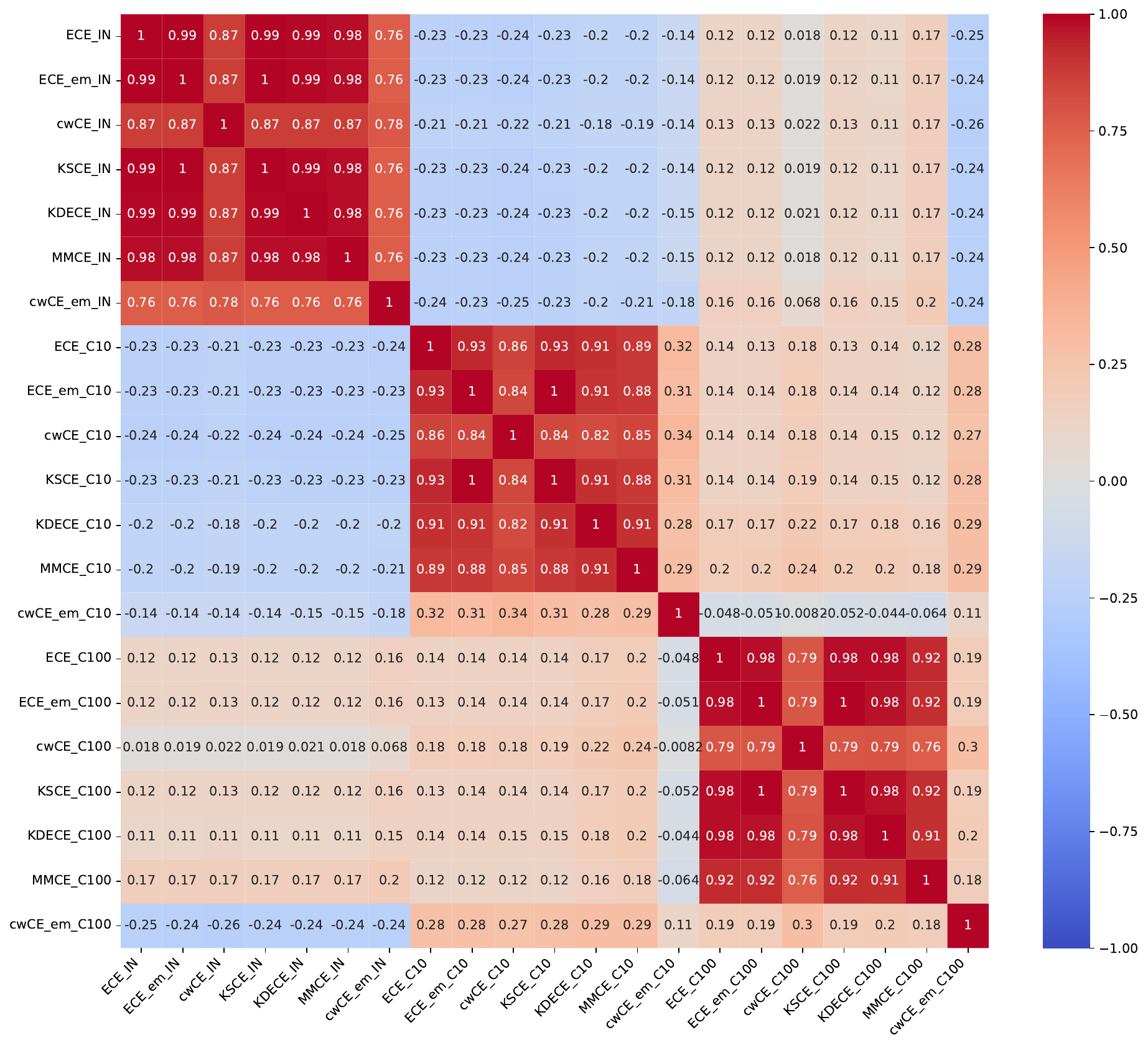}
\end{center}
    \caption{Kendall Ranking Correlation Matrix for the CIFAR-10, CIFAR-100, and the ImageNet16-120 dataset with calibration metrics measured on TSS, filtered by top 100 accuracy.}
\label{fig:Kendall_Metric_100}
\end{figure}

\begin{figure}[!ht]
\begin{center}
   \includegraphics[width=\linewidth]{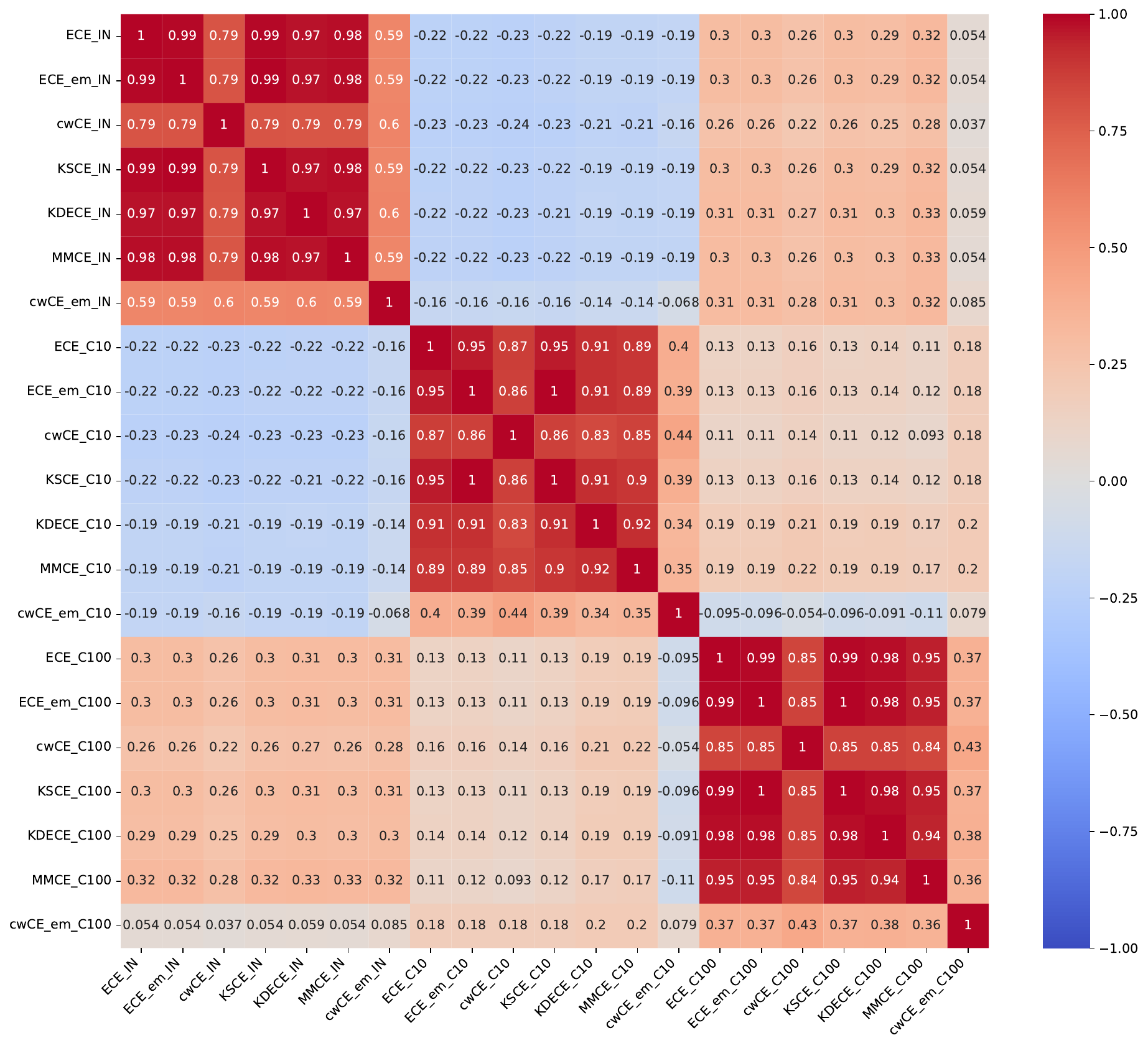}
\end{center}
    \caption{Kendall Ranking Correlation Matrix for the CIFAR-10, CIFAR-100, and the ImageNet16-120 dataset with calibration metrics measured on TSS, filtered by top 1000 accuracy.}
\label{fig:Kendall_Metric_1000}
\end{figure}

\begin{figure}[!ht]
\begin{center}
   \includegraphics[width=\linewidth]{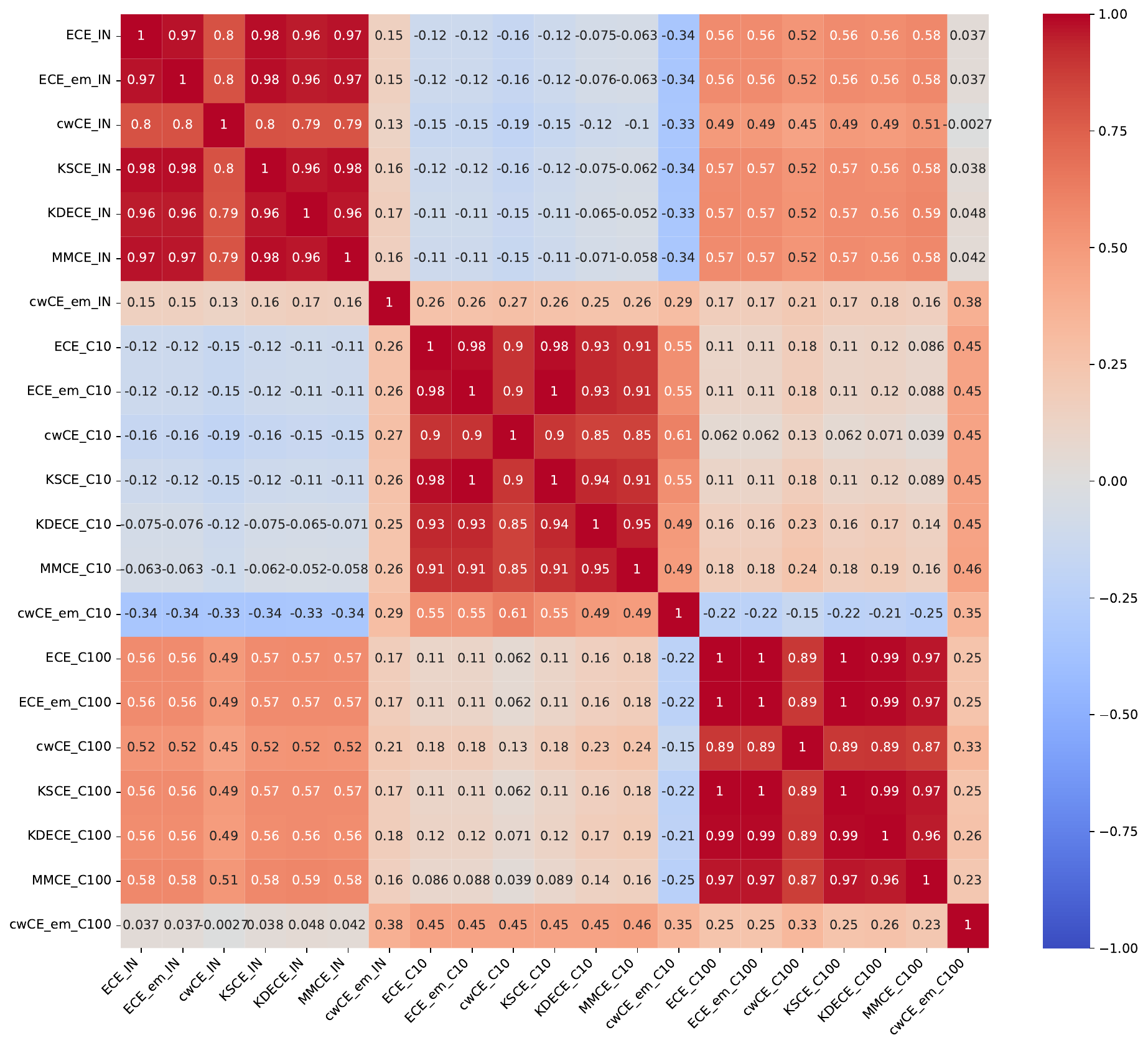}
\end{center}
    \caption{Kendall Ranking Correlation Matrix for the CIFAR-10, CIFAR-100, and the ImageNet16-120 dataset with calibration metrics measured on TSS, filtered by top 5000 accuracy.}
\label{fig:Kendall_Metric_5000}
\end{figure}

\begin{figure}[!ht]
\begin{center}
   \includegraphics[width=\linewidth]{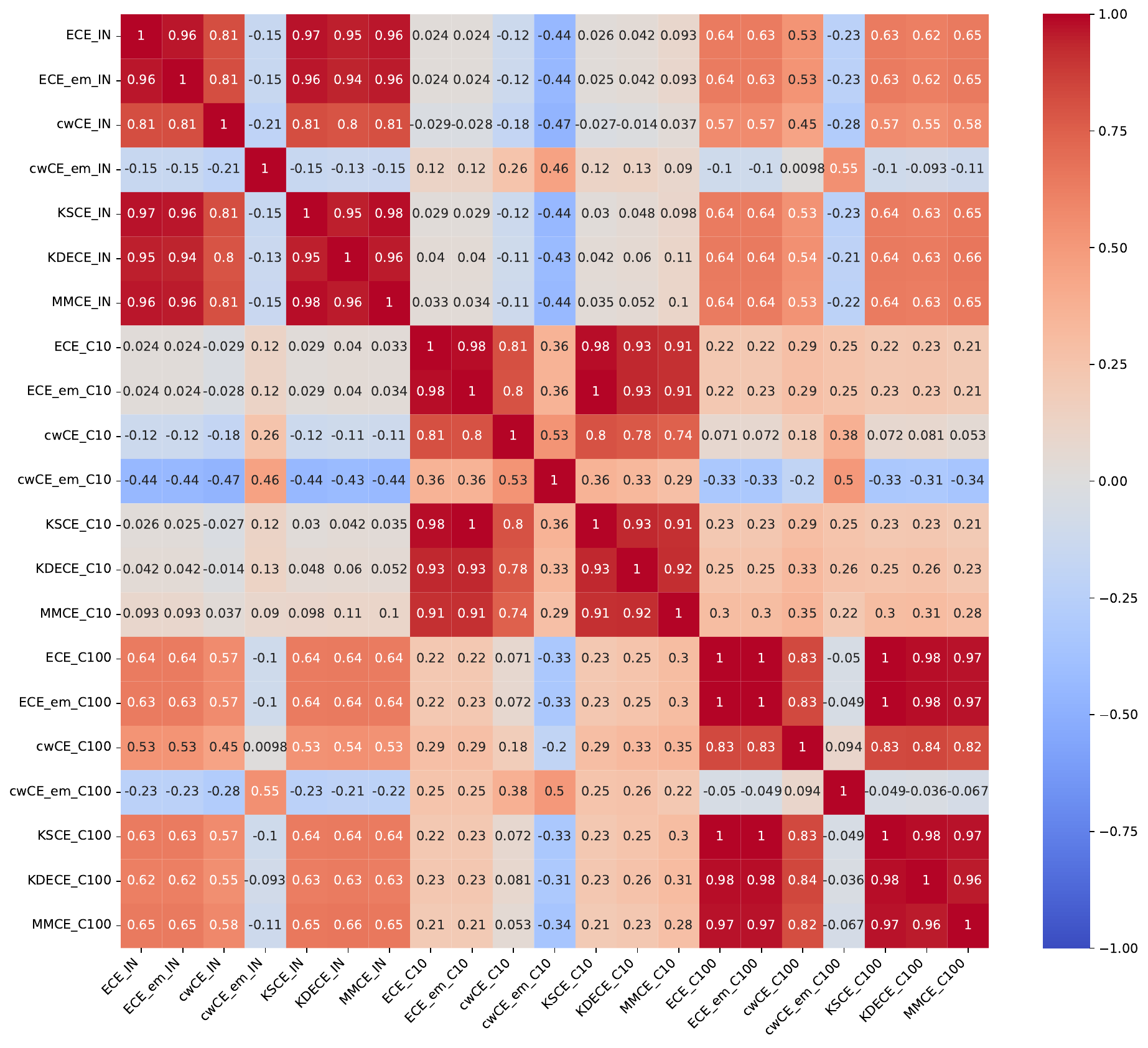}
\end{center}
    \caption{Kendall Ranking Correlation Matrix for the CIFAR-10, CIFAR-100, and the ImageNet16-120 dataset with calibration metrics measured on all models in TSS.}
\label{fig:Kendall_Metric}
\end{figure}

\clearpage
\section{Calibration Kendall Ranking Confusion Matrix for SSS}
\label{sec:Calibration Kendall Ranking Confusion Matrix for SSS}
In this section, we provide comprehensive correlation matrix between calibration metrics on different datasets on SSS, filtered by different top accuracy populations.

\begin{figure}[!ht]
\begin{center}
   \includegraphics[width=\linewidth]{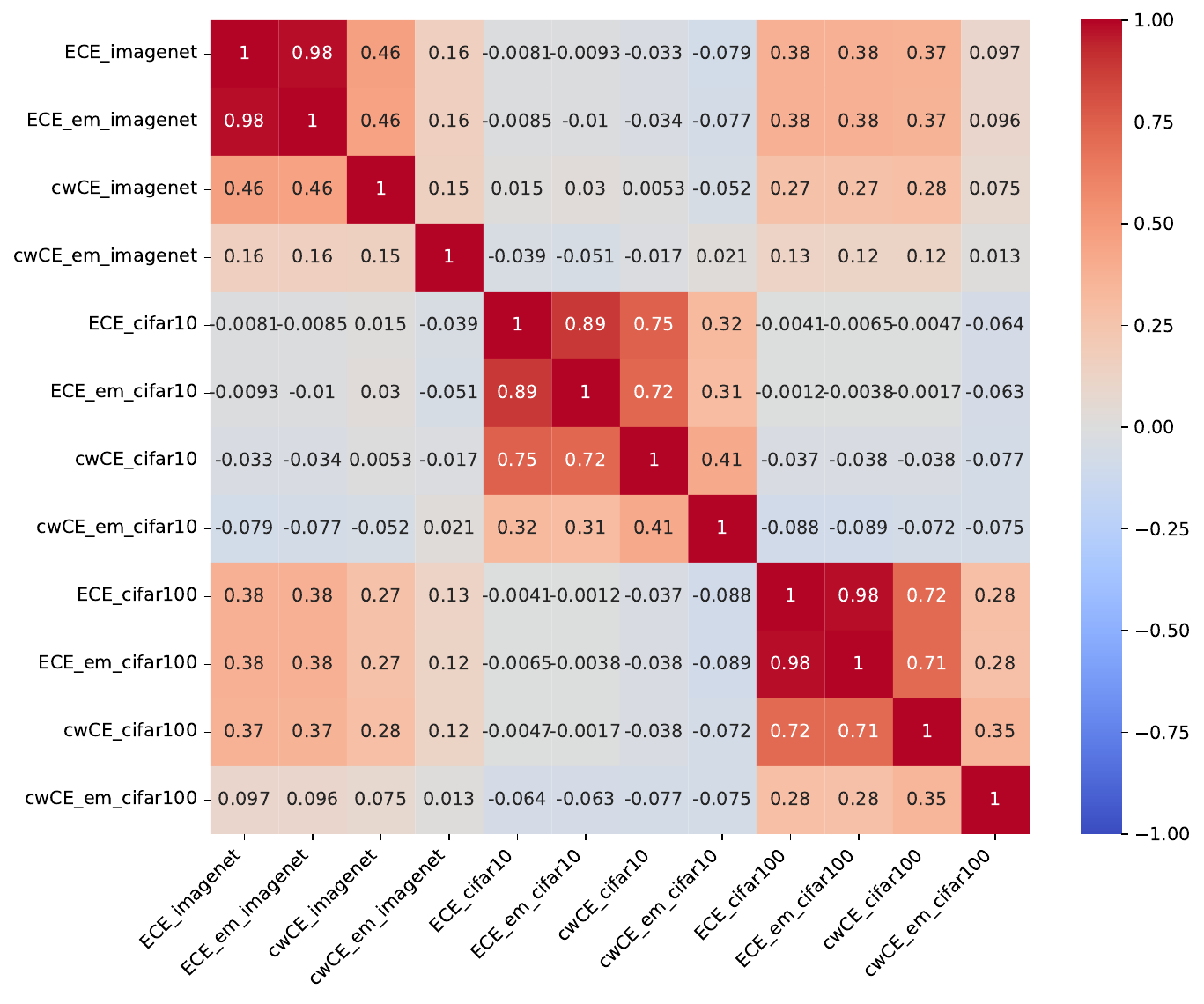}
\end{center}
    \caption{Kendall Ranking Correlation Matrix for the CIFAR-10, CIFAR-100, and the ImageNet16-120 dataset with calibration metrics measured on SSS, filtered by top 100 accuracy.}
\label{fig:Kendall_Metric_sss_100}
\end{figure}

\begin{figure}[!ht]
\begin{center}
   \includegraphics[width=\linewidth]{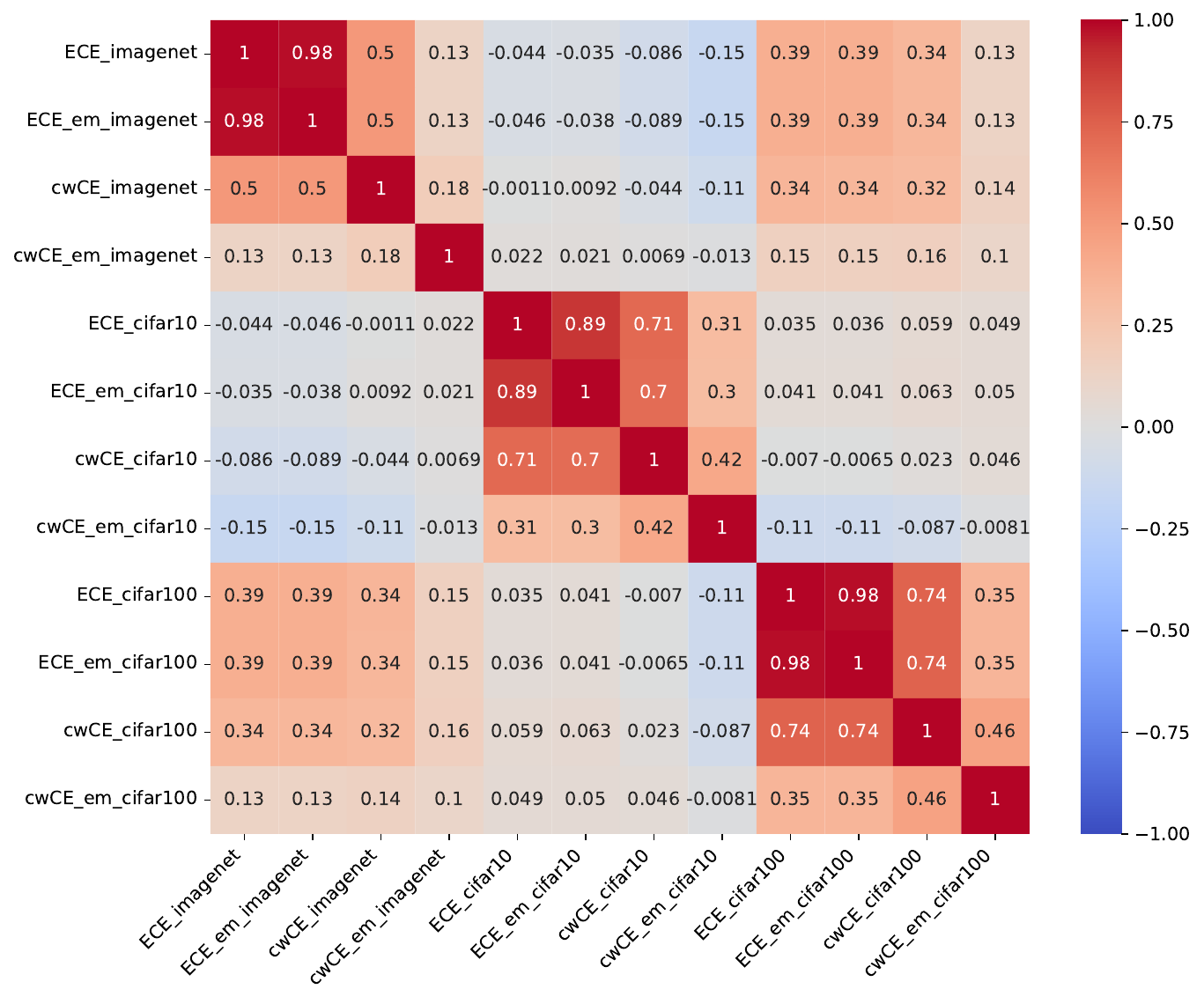}
\end{center}
    \caption{Kendall Ranking Correlation Matrix for the CIFAR-10, CIFAR-100, and the ImageNet16-120 dataset with calibration metrics measured on SSS, filtered by top 1000 accuracy.}
\label{fig:Kendall_Metric_sss_1000}
\end{figure}

\begin{figure}[!ht]
\begin{center}
   \includegraphics[width=\linewidth]{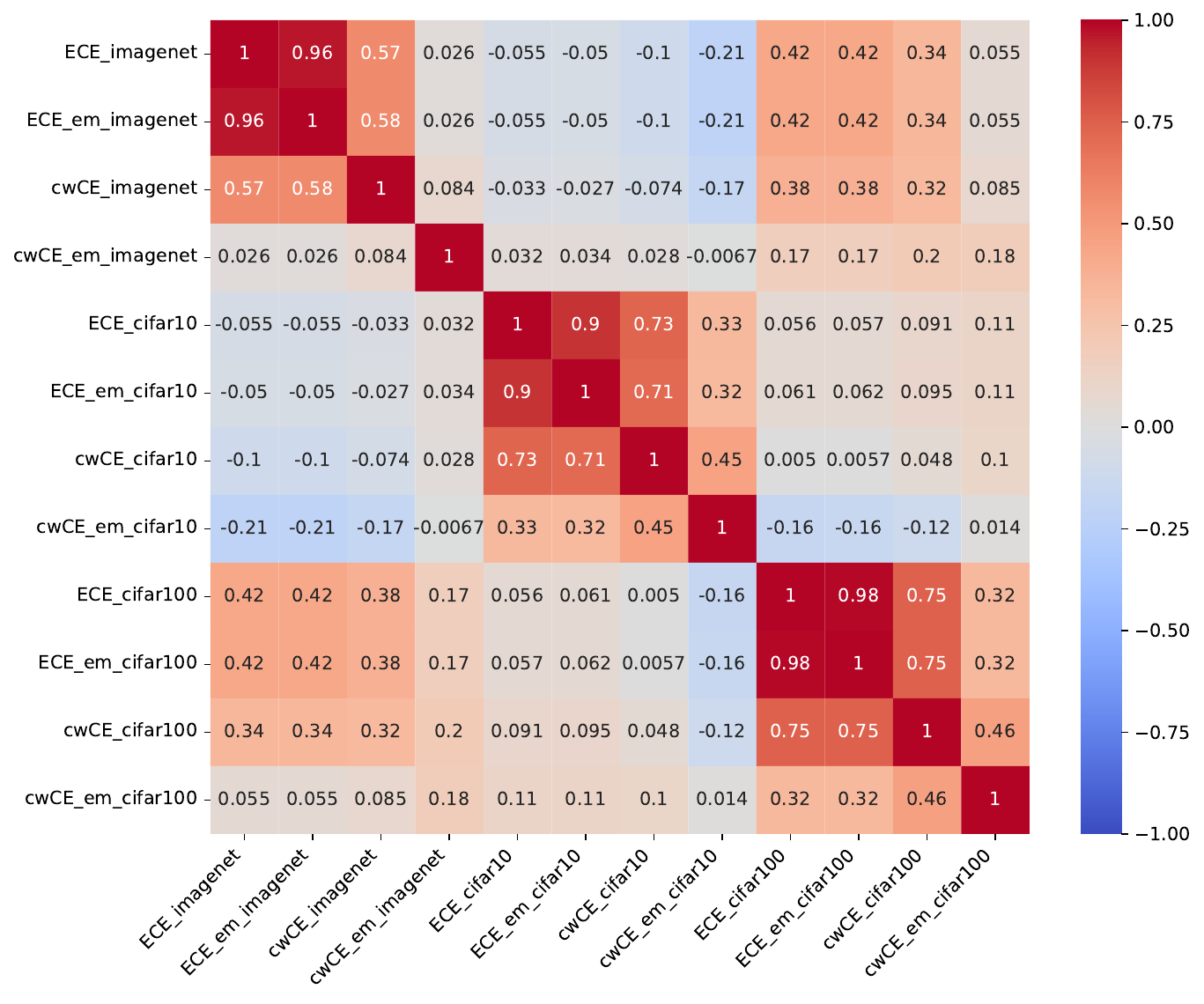}
\end{center}
    \caption{Kendall Ranking Correlation Matrix for the CIFAR-10, CIFAR-100, and the ImageNet16-120 dataset with calibration metrics measured on SSS, filtered by top 5000 accuracy.}
\label{fig:Kendall_Metric_sss_5000}
\end{figure}

\begin{figure}[!ht]
\begin{center}
   \includegraphics[width=\linewidth]{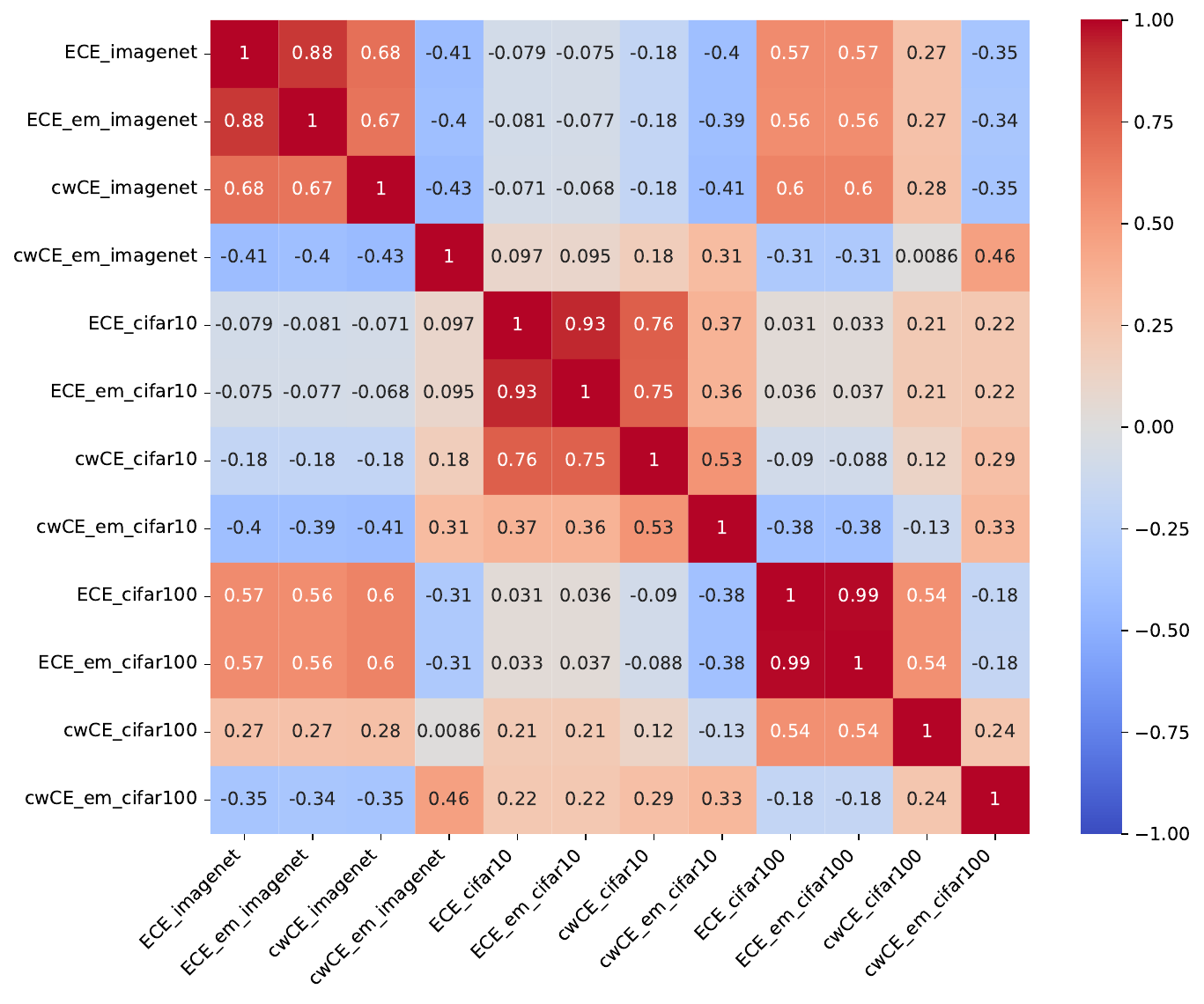}
\end{center}
    \caption{Kendall Ranking Correlation Matrix for the CIFAR-10, CIFAR-100, and the ImageNet16-120 dataset with calibration metrics measured on all models in SSS.}
\label{fig:Kendall_Metric_sss}
\end{figure}

\clearpage
\section{Kendall Ranking Correlation Matrix before and after Temperature Scaling}
\label{sec:Kendall Ranking Correlation Matrix before and after Temperature Scaling}
In this section, we provide the full correlation matrix between ECE evaluated with different bin size and pre- and post-temperature scaling.
\begin{figure}[!ht]
\begin{center}
   \includegraphics[width=\linewidth]{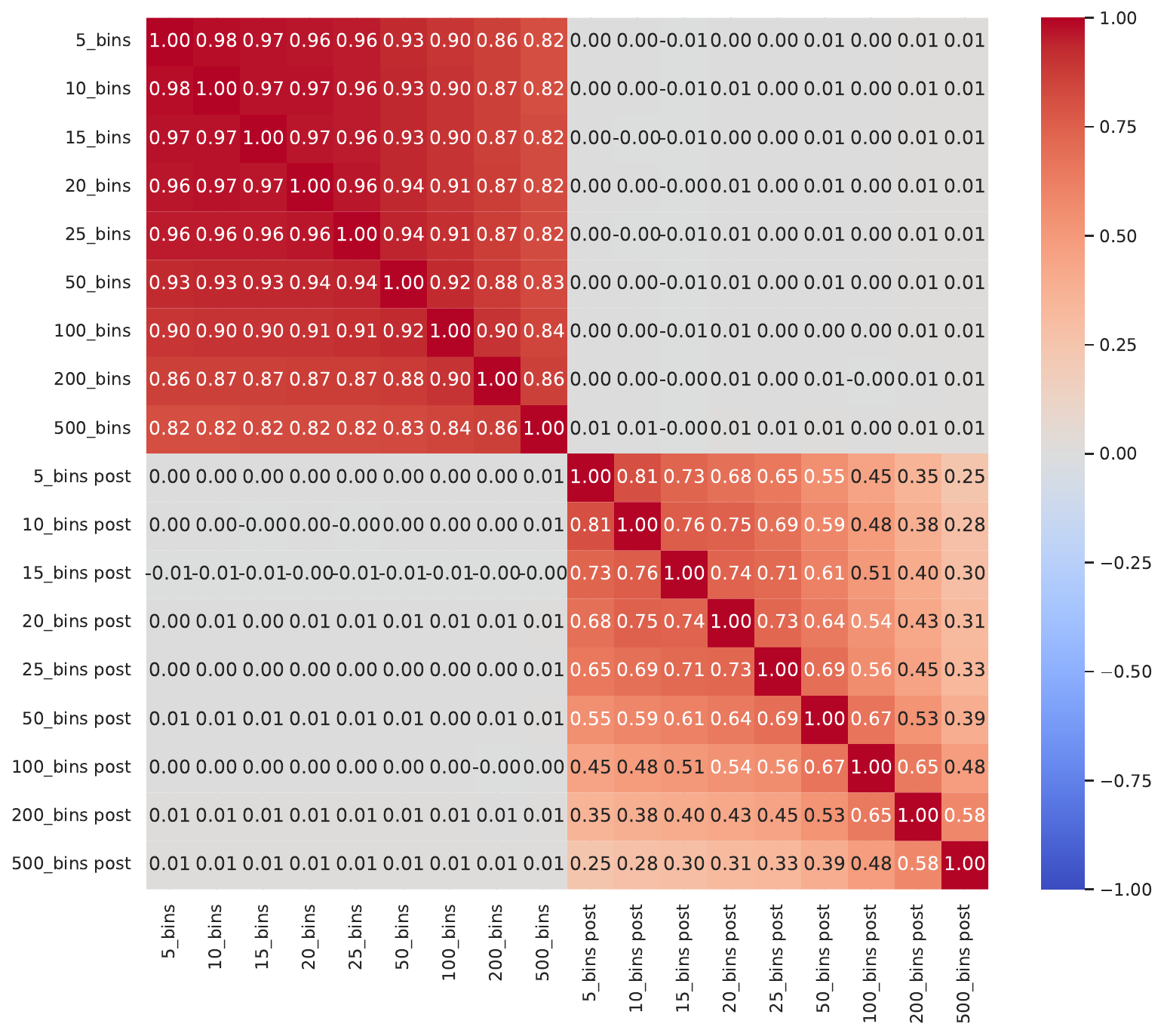}
\end{center}
    \caption{Kendall Ranking Correlation Matrix of ECE using different bin size before and after temperature scaling on CIFAR-10.}
\label{fig:Kendall_cifar10_temp}
\end{figure}

\begin{figure}[!ht]
\begin{center}
   \includegraphics[width=\linewidth]{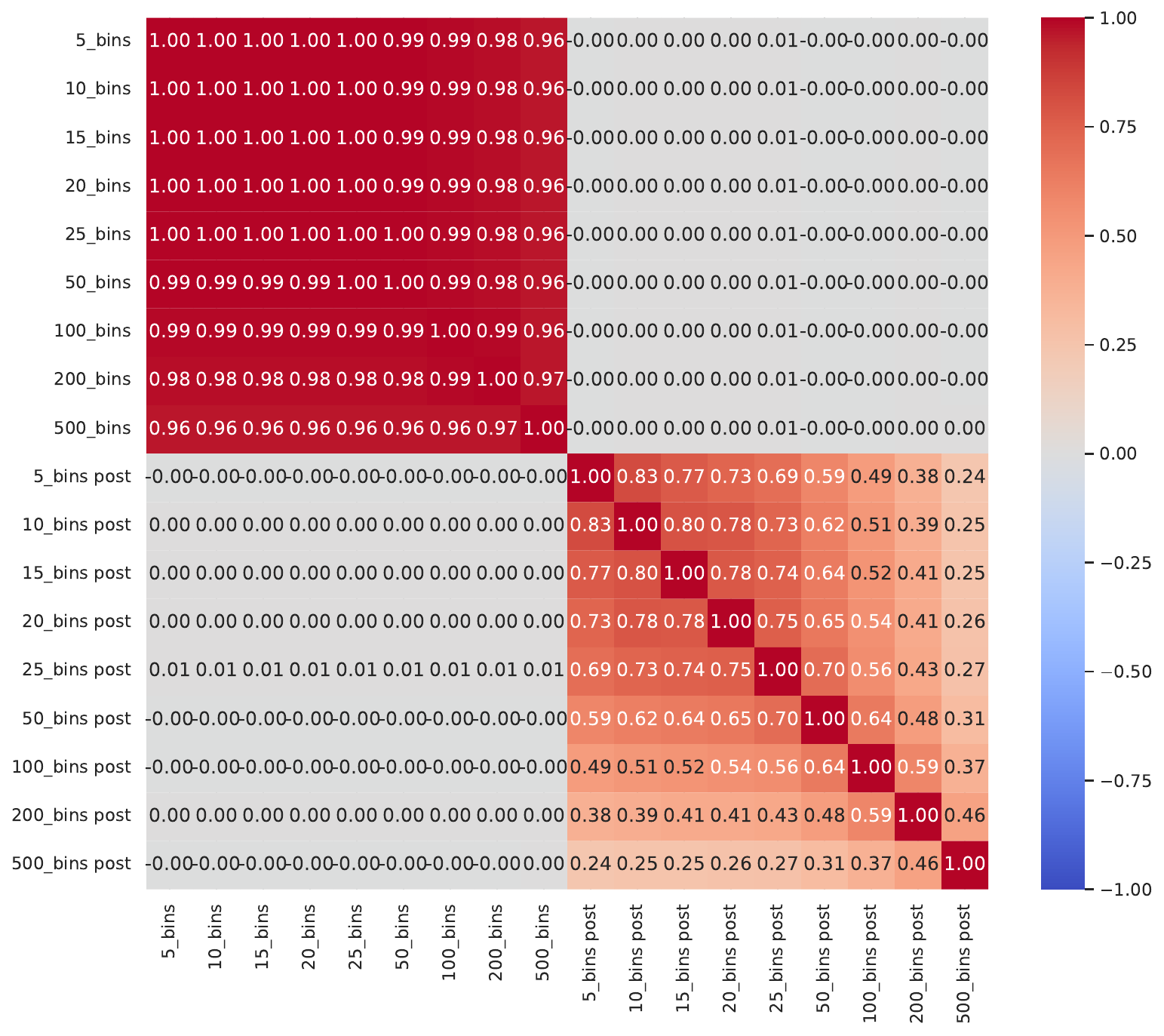}
\end{center}
    \caption{Kendall Ranking Correlation Matrix of ECE using different bin size before and after temperature scaling on CIFAR-100.}
\label{fig:Kendall_cifar100_temp}
\end{figure}

\begin{figure}[!ht]
\begin{center}
   \includegraphics[width=\linewidth]{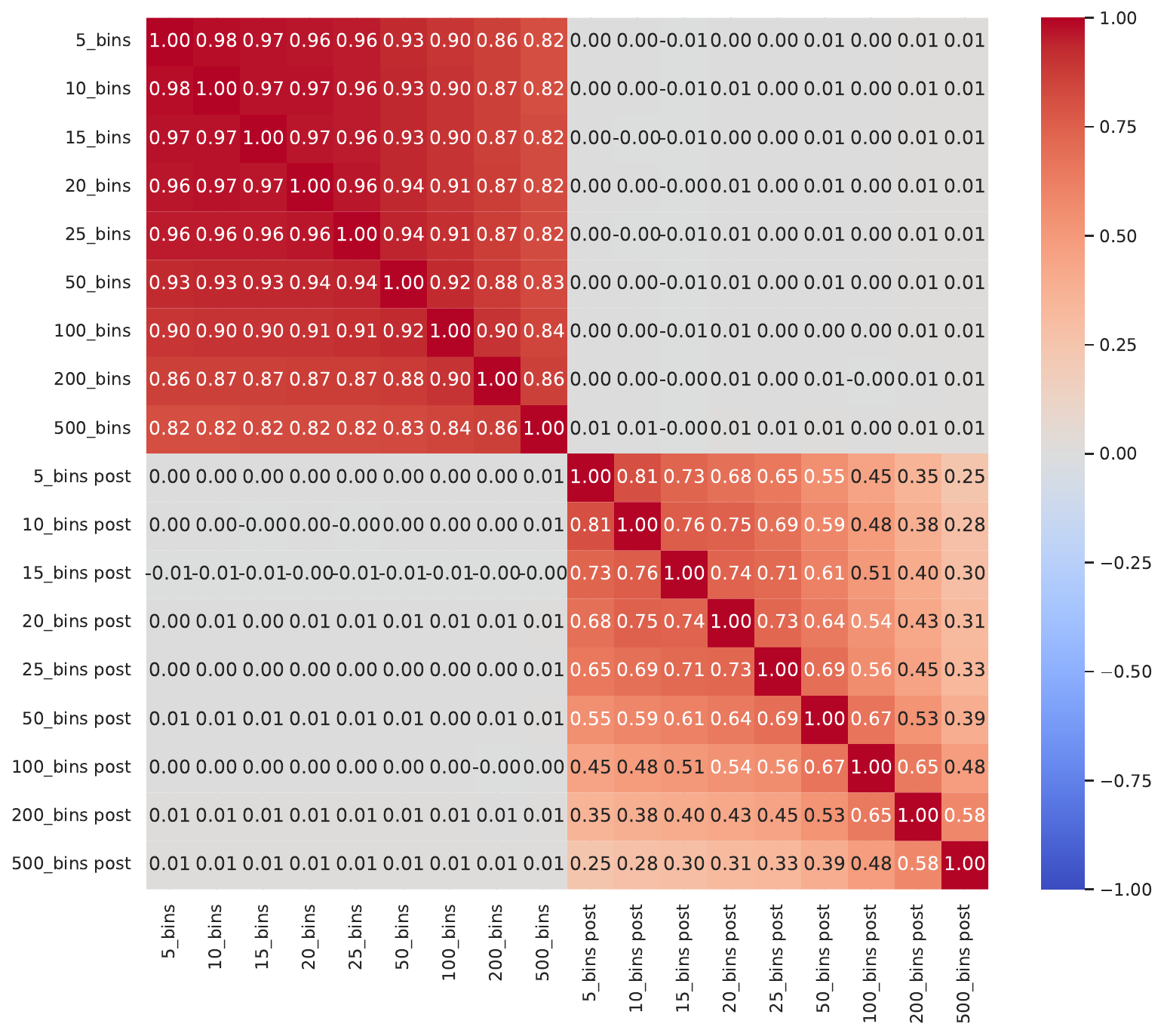}
\end{center}
    \caption{Kendall Ranking Correlation Matrix of ECE using different bin size before and after temperature scaling on ImageNet16-120.}
\label{fig:Kendall_imagenet_temp}
\end{figure}

\clearpage
\section{Relationship Accuracy, Robustness and Calibration Measurements}
\label{sec:Relationship Accuracy, Robustness and Calibration Measurements}
In Figure~\ref{fig:kendall_ece_top_ranked_models}, we plot a correlation coefficient line for selected metrics. In this section, we provide correlation matrix between key robustness metrics and calibration metrics on different datasets along multiple top ranked populations.
\begin{figure}[!ht]
\begin{center}
   \includegraphics[width=\linewidth]{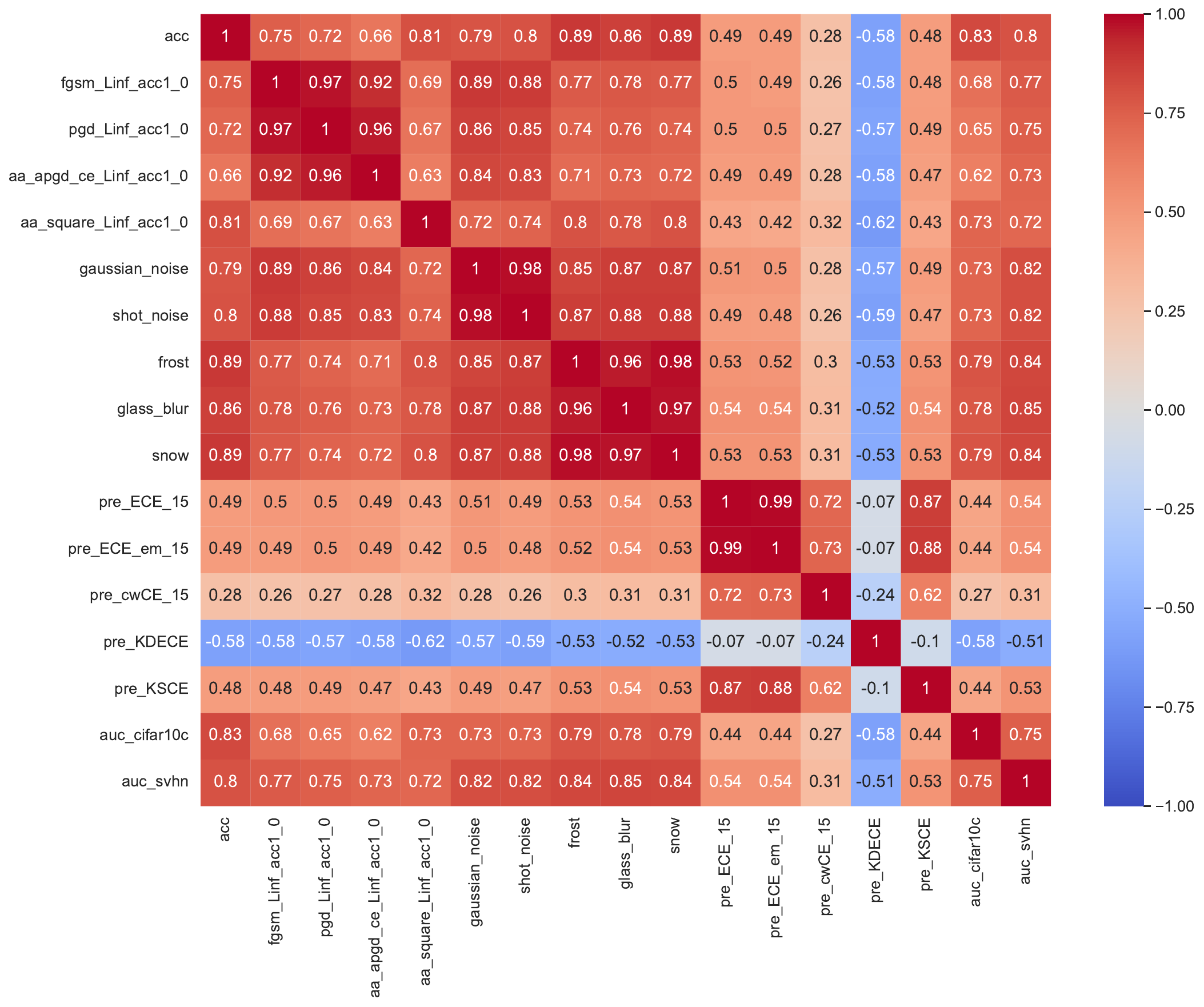}
\end{center}
    \caption{Kendall Ranking Correlation Matrix for the CIFAR-10 with calibration metrics measured on TSS, filtered by top 1000 accuracy. The Correlation Matrix presents the ranking correlations between 5 different calibration metrics and various adversarial attack and image perturbation robustness measurements.}
\label{fig:cifar10_top1000acc}
\end{figure}

\begin{figure}[!ht]
\begin{center}
   \includegraphics[width=\linewidth]{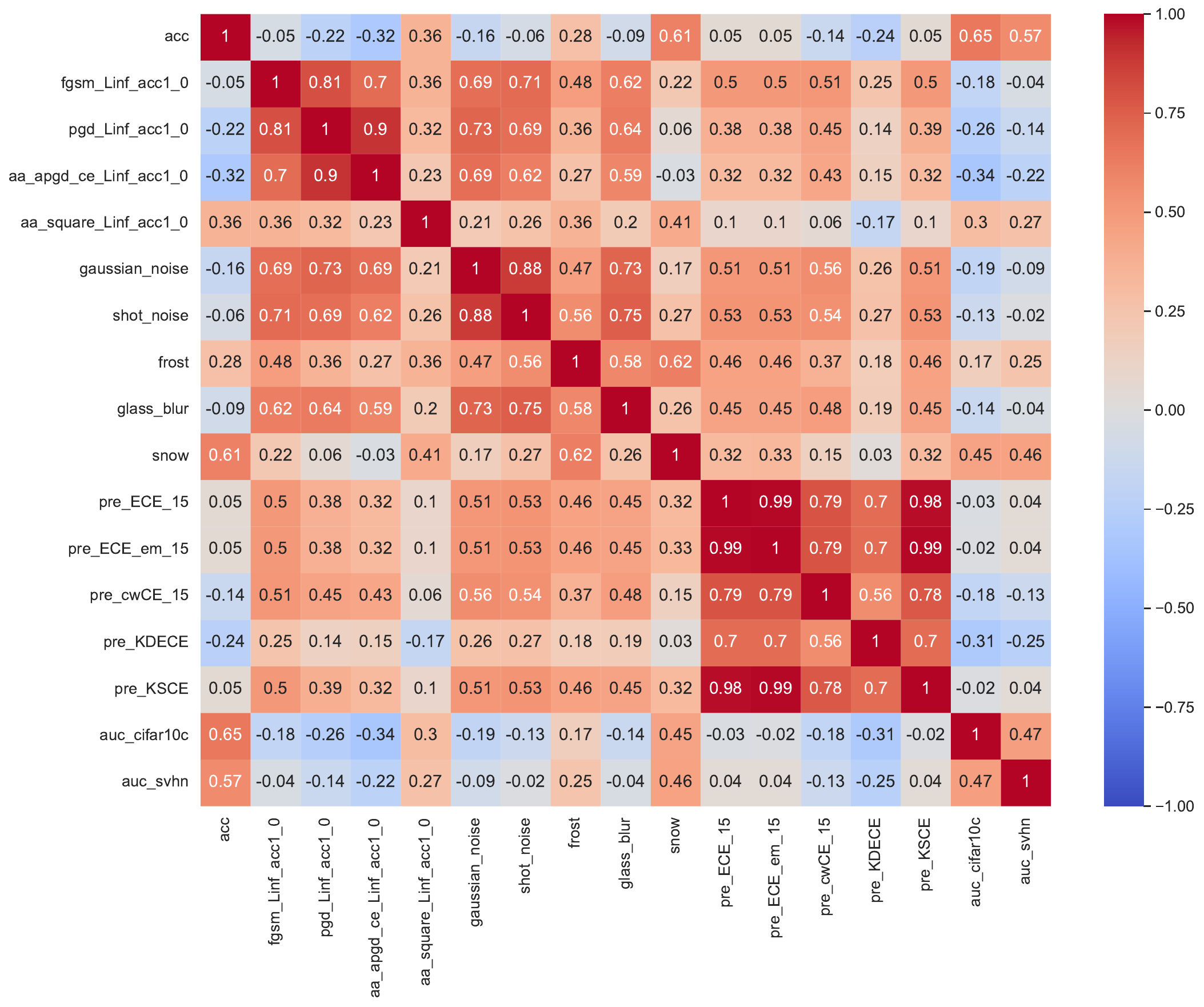}
\end{center}
    \caption{Kendall Ranking Correlation Matrix for the CIFAR-10 with calibration metrics measured on TSS, filtered by top 5000 accuracy. The Correlation Matrix presents the ranking correlations between 5 different calibration metrics and various adversarial attack and image perturbation robustness measurements.}
\label{fig:cifar10_top5000acc}
\end{figure}

\begin{figure}[!ht]
\begin{center}
   \includegraphics[width=\linewidth]{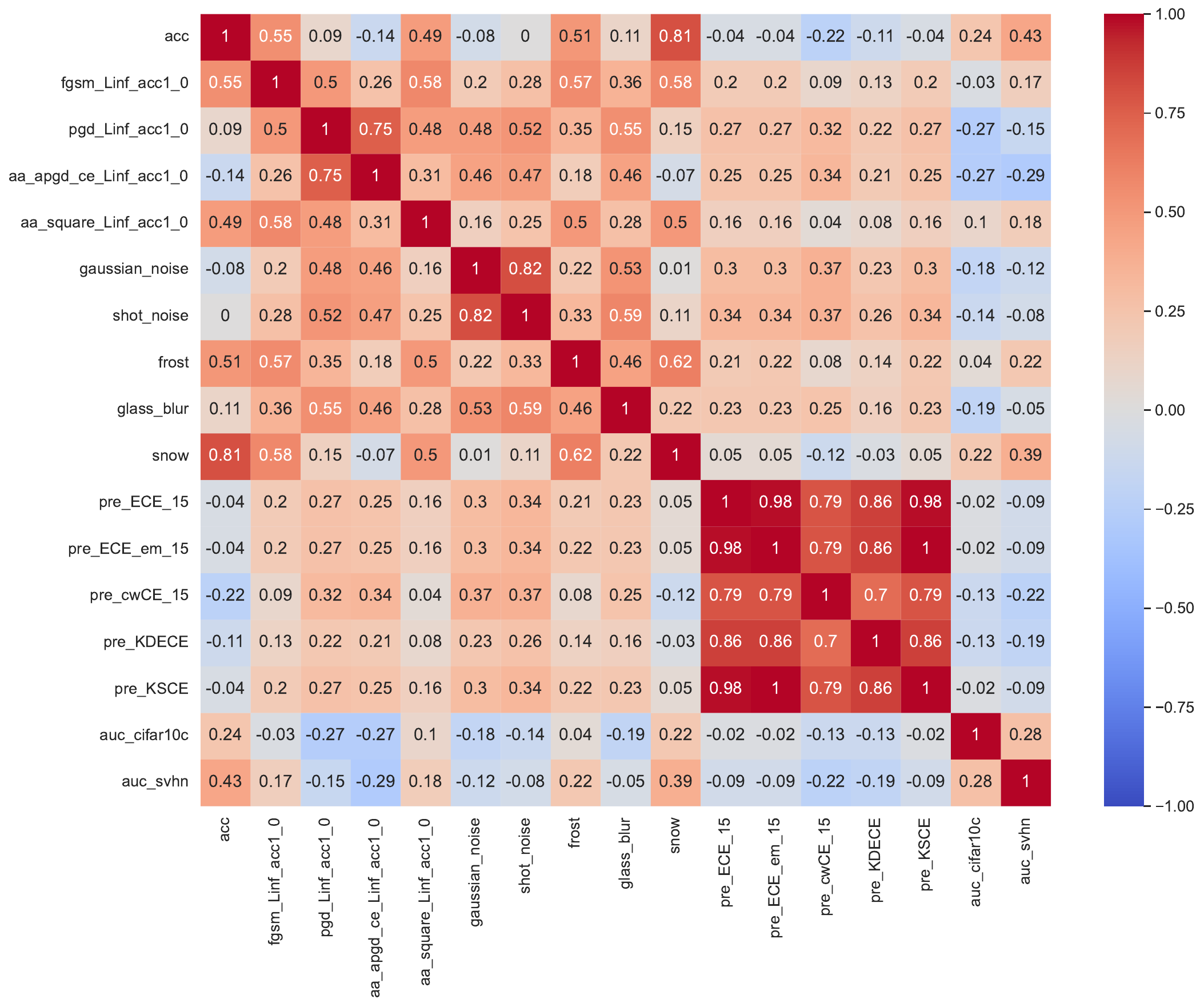}
\end{center}
    \caption{Kendall Ranking Correlation Matrix for the CIFAR-10 with calibration metrics measured on all models in TSS. The Correlation Matrix presents the ranking correlations between 5 different calibration metrics and various adversarial attack and image perturbation robustness measurements.}
\label{fig:cifar10_top15625acc}
\end{figure}

\begin{figure}[!ht]
\begin{center}
   \includegraphics[width=\linewidth]{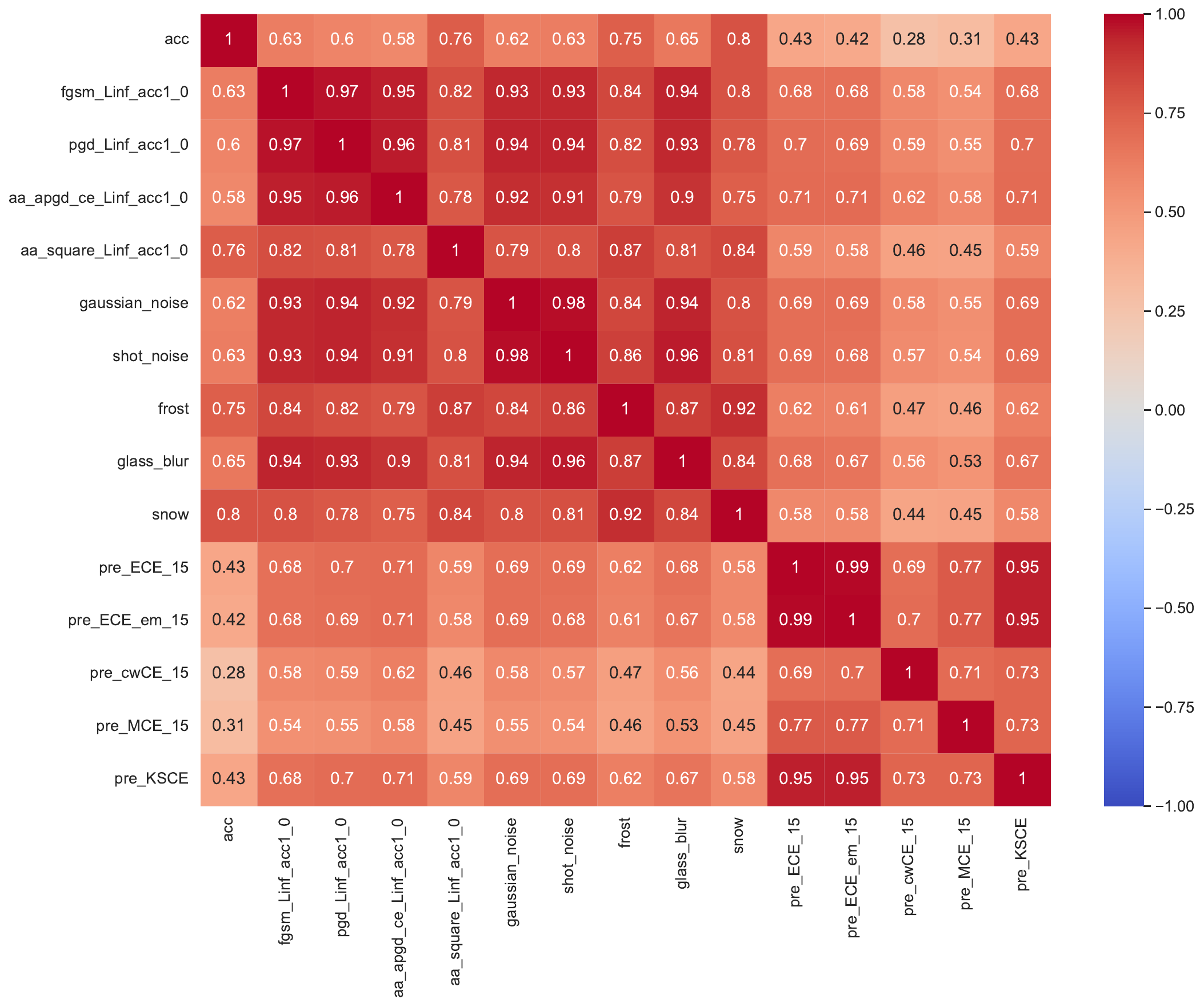}
\end{center}
    \caption{Kendall Ranking Correlation Matrix for the CIFAR-100 with calibration metrics measured on TSS, filtered by top 1000 accuracy. The Correlation Matrix presents the ranking correlations between 5 different calibration metrics and various adversarial attack and image perturbation robustness measurements.}
\label{fig:cifar100_top1000acc}
\end{figure}

\begin{figure}[!ht]
\begin{center}
   \includegraphics[width=\linewidth]{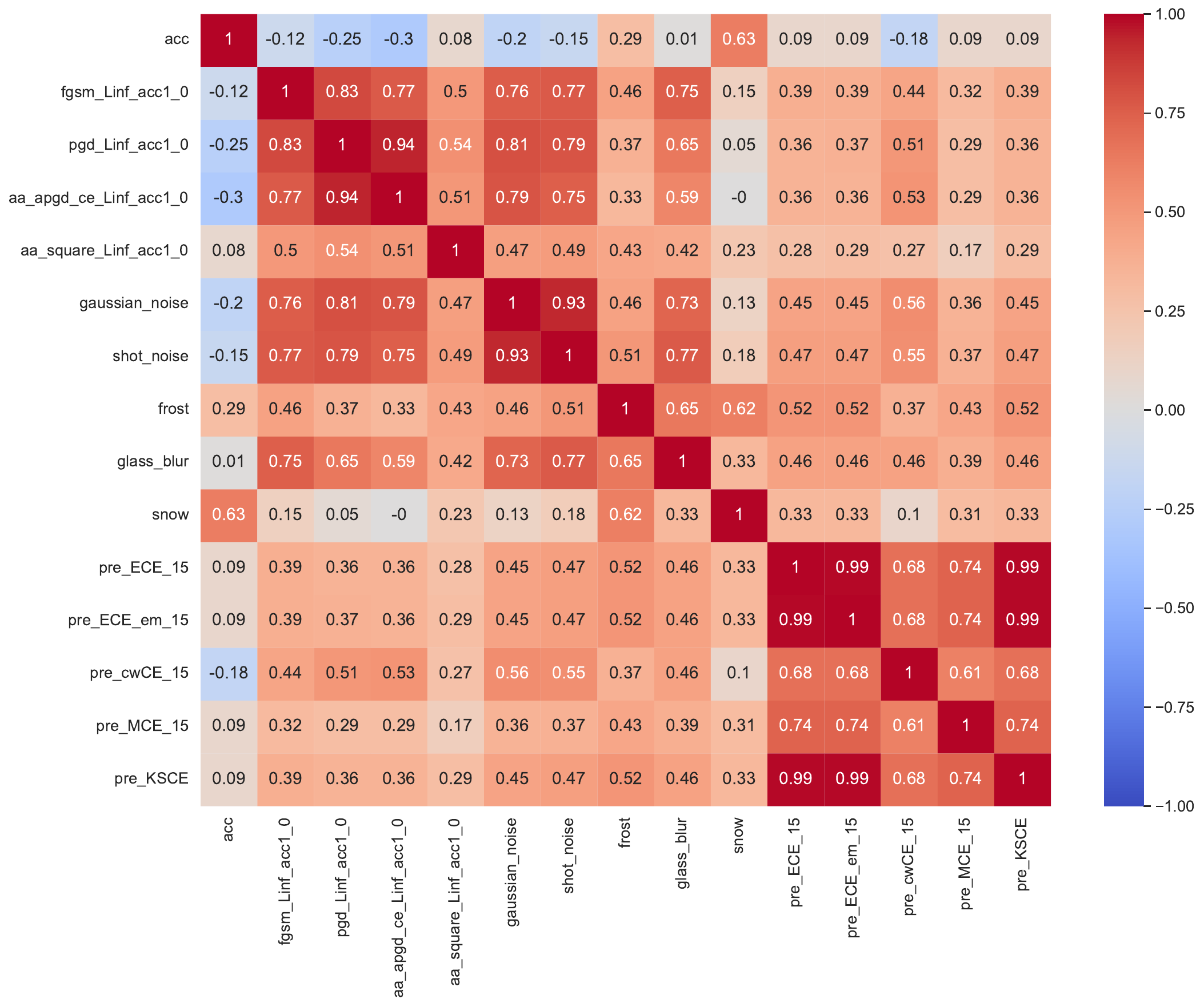}
\end{center}
    \caption{Kendall Ranking Correlation Matrix for the CIFAR-100 with calibration metrics measured on TSS, filtered by top 5000 accuracy. The Correlation Matrix presents the ranking correlations between 5 different calibration metrics and various adversarial attack and image perturbation robustness measurements.}
\label{fig:cifar100_top5000acc}
\end{figure}

\begin{figure}[!ht]
\begin{center}
   \includegraphics[width=\linewidth]{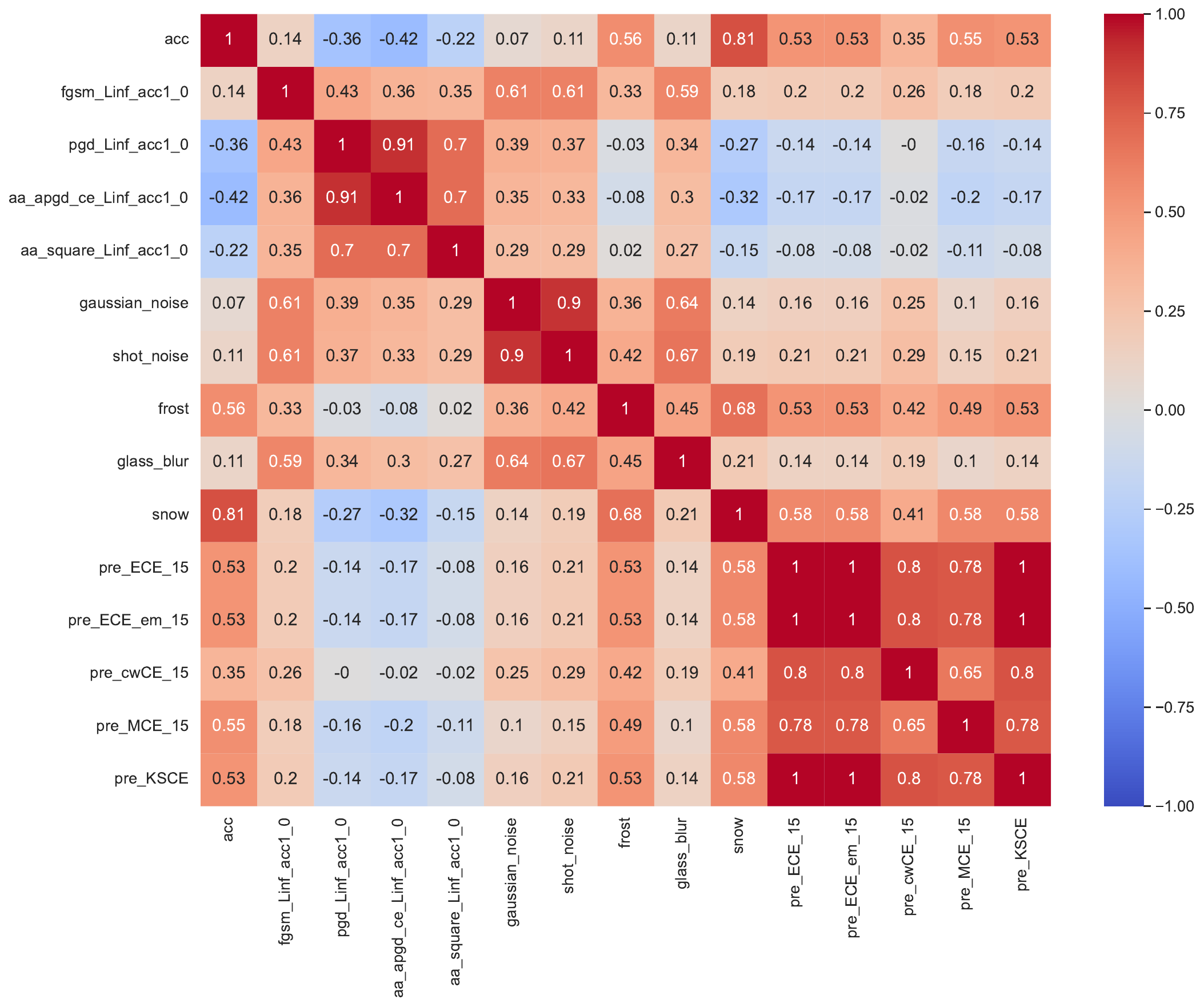}
\end{center}
    \caption{Kendall Ranking Correlation Matrix for the CIFAR-100 with calibration metrics measured on all models in TSS. The Correlation Matrix presents the ranking correlations between 5 different calibration metrics and various adversarial attack and image perturbation robustness measurements.}
\label{fig:cifar100_top15625acc}
\end{figure}

\begin{figure}[!ht]
\begin{center}
   \includegraphics[width=\linewidth]{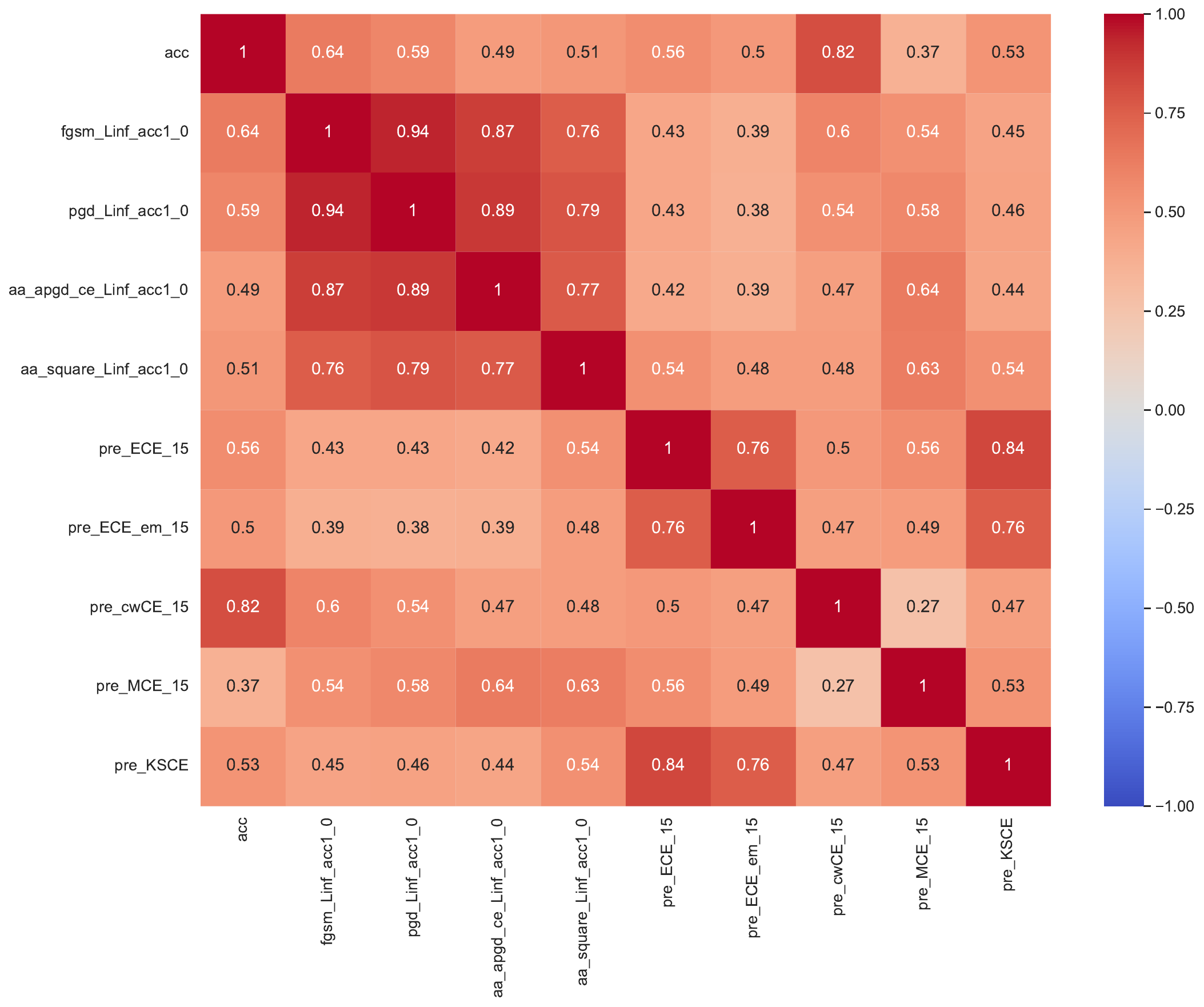}
\end{center}
    \caption{Kendall Ranking Correlation Matrix for the ImageNet16-120 with calibration metrics measured on TSS, filtered by top 1000 accuracy. The Correlation Matrix presents the ranking correlations between 5 different calibration metrics and various adversarial attack and image perturbation robustness measurements.}
\label{fig:ImageNet_top1000acc}
\end{figure}

\begin{figure}[!ht]
\begin{center}
   \includegraphics[width=\linewidth]{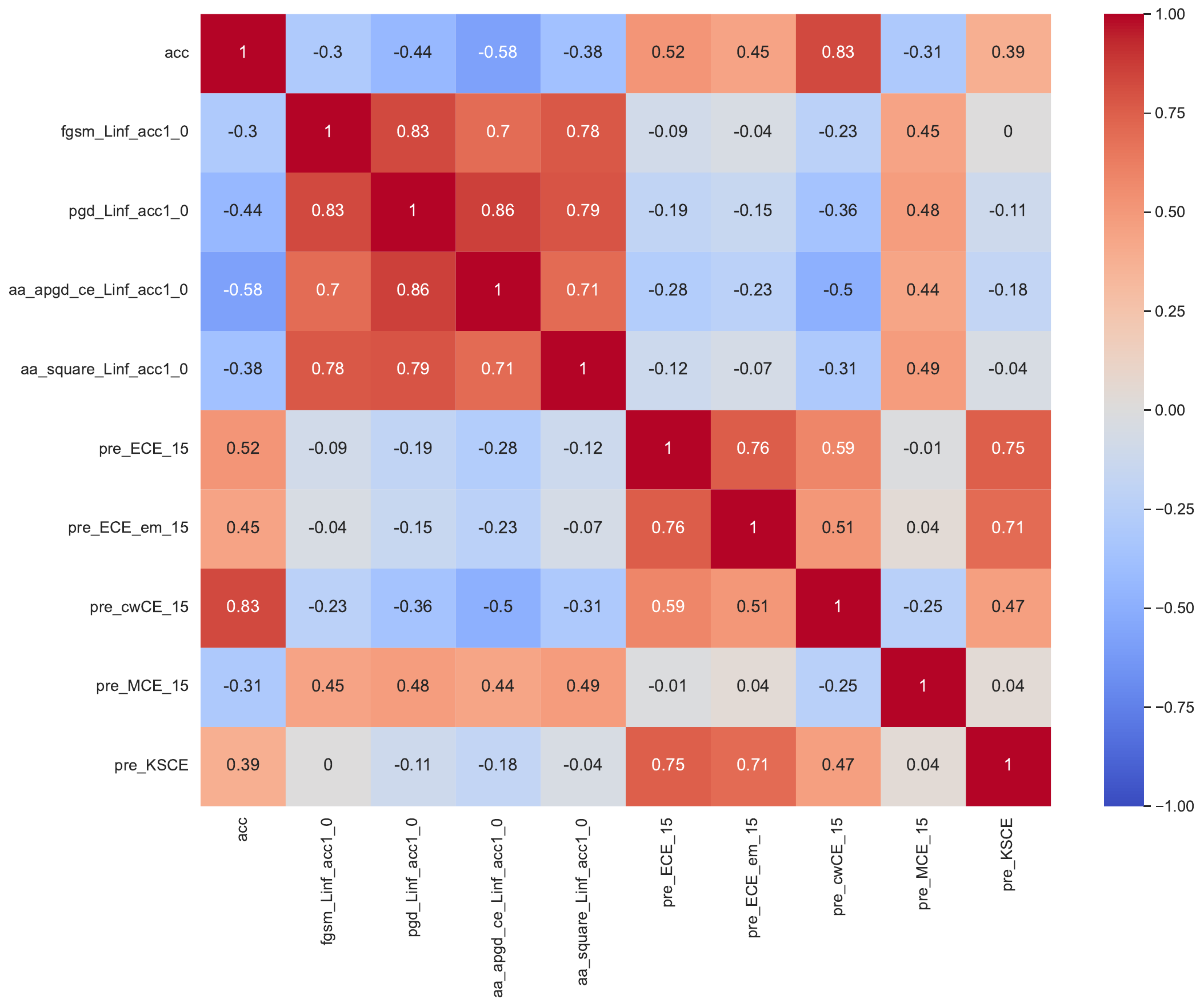}
\end{center}
    \caption{Kendall Ranking Correlation Matrix for the ImageNet16-120 with calibration metrics measured on TSS, filtered by top 5000 accuracy. The Correlation Matrix presents the ranking correlations between 5 different calibration metrics and various adversarial attack and image perturbation robustness measurements.}
\label{fig:ImageNet_top5000acc}
\end{figure}

\begin{figure}[!ht]
\begin{center}
   \includegraphics[width=\linewidth]{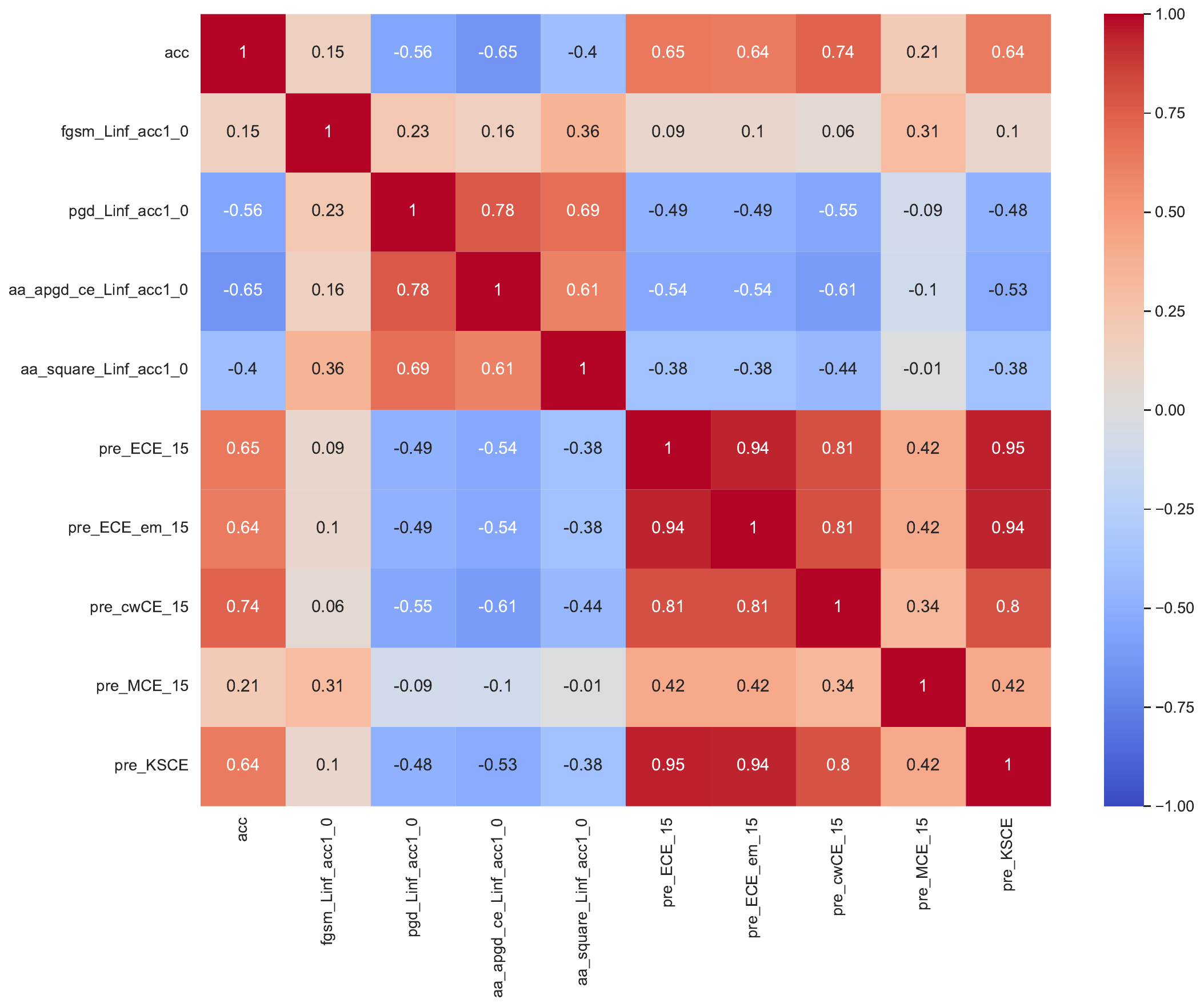}
\end{center}
    \caption{Kendall Ranking Correlation Matrix for the ImageNet16-120 with calibration metrics measured on all models in TSS. The Correlation Matrix presents the ranking correlations between 5 different calibration metrics and various adversarial attack and image perturbation robustness measurements.}
\label{fig:long}
\label{fig:onecol}
\end{figure}

\end{document}